\pdfoutput=1
\documentclass[letterpaper, 10 pt, journal, twoside]{support/IEEEtran}

\usepackage{makecell}
\usepackage{times}
\usepackage[pdftex]{graphicx}
\usepackage{subfigure}
\usepackage{amsmath,amssymb,amsopn,amstext,amsfonts}
\usepackage{cancel}
\usepackage[space]{cite}
\usepackage{siunitx}
\usepackage{soul}
\usepackage{cuted}
\usepackage{caption}
\usepackage{lipsum,multicol}
\usepackage{stfloats}
\everymath{\displaystyle}
\usepackage{booktabs}
\usepackage{threeparttable}
\usepackage{amsthm}
\usepackage{xfrac}
\usepackage{balance}
\usepackage{color}
\usepackage{mathtools}

\usepackage{tabularx}
\usepackage{bm}
\usepackage{diagbox}
\usepackage{float}
\usepackage{epstopdf}
\usepackage{url}
\usepackage{multirow}
\usepackage{multicol}
\usepackage{tikz}
\usepackage{tablefootnote}
\usepackage[linkcolor=black,citecolor=black,urlcolor=black,colorlinks=true]{hyperref}
\usepackage{enumitem}
\usepackage[ruled,vlined,linesnumbered]{algorithm2e}
\usepackage{pifont}
\DeclareMathAlphabet\mathbfcal{OMS}{cmsy}{b}{n}

\soulregister\cite7
\soulregister\citep7
\soulregister\citet7
\soulregister\ref7
\soulregister\it7
\soulregister\pageref7

\usepackage{arydshln}
\graphicspath{{figures/}}
\DeclareGraphicsExtensions{.pdf,.png,.jpg,.eps}

\IEEEoverridecommandlockouts

\title{FAST-LIVO2: Fast, Direct LiDAR-Inertial-Visual Odometry}

\author{Chunran Zheng$^{1}$, Wei Xu$^{1}$, Zuhao Zou$^{1}$, Tong Hua$^{1}$, Chongjian Yuan$^{1}$, Dongjiao He$^{1}$, Bingyang Zhou$^{1}$, Zheng Liu$^{1}$, Jiarong Lin$^{1}$, Fangcheng Zhu$^{1}$, Yunfan Ren$^{1}$, Rong Wang$^{2}$, Fanle Meng$^{2}$, Fu Zhang$^{1\dag}$ 
    \thanks{$^{\dag}$Corresponding author 
 (email: fuzhang@hku.hk)}
    \thanks{$^{1}$Mechatronics and Robotic Systems (MaRS) Laboratory, Department of Mechanical Engineering, University of Hong Kong, Hong Kong SAR, China.}
   \thanks{$^{2}$Information Science Academy of China Electronics Technology Group Corporation}
}
\makeatletter
\let\@oldmaketitle\@maketitle
\renewcommand{\@maketitle}
{
\@oldmaketitle 
\centering
\setcounter{figure}{0}
\begin{minipage}{1.0\linewidth}
	\includegraphics[width=1.0\textwidth]{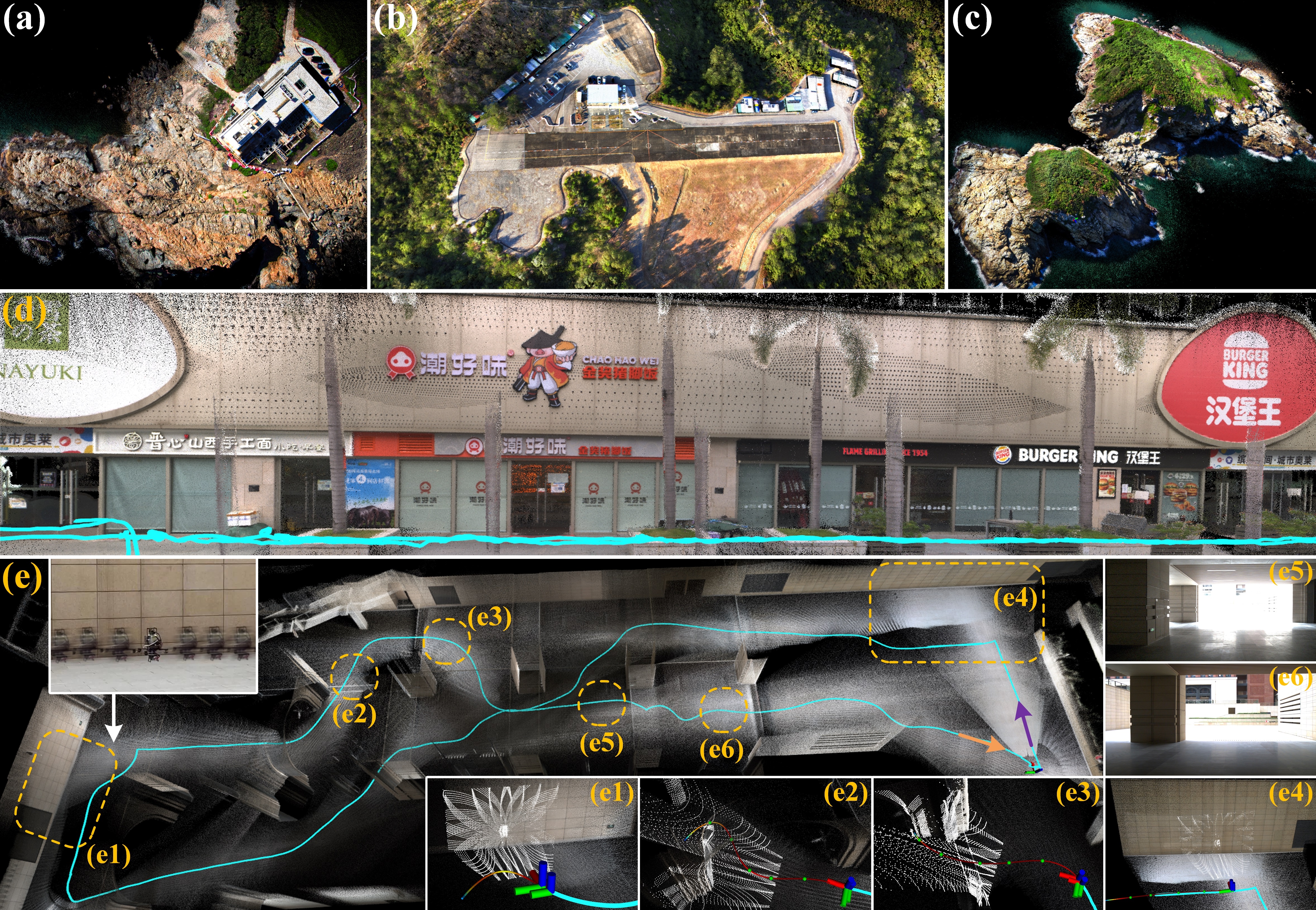}
	\vspace{-0.5cm}
	\captionof{figure}{FAST-LIVO2 mapping results generated {in real time}. (a)-(c) showcase airborne mapping, (d) represents a retail street collected with a handheld device, and (e) demonstrates an experiment where a UAV carrying a LiDAR, camera and inertial sensor perform {real-time} state estimation (i.e., FAST-LIVO2), trajectory planning, and tracking control all on its onboard computer. In (d)-(e), blue lines represent the computed trajectory. In (e1)-(e4), white points indicate the LiDAR scan at that moment, and colored lines depict the planned trajectory.
	(e1) and (e4) mark areas of LiDAR degeneration. (e2) and (e3) show obstacle avoidance. (e5) and (e6) depict the camera first-person view from indoor to outdoor, highlighting large illumination variation from sudden overexposure to normal (see our accompanying video on YouTube: \href{https://youtu.be/aSAwVqR22mo}{\tt{youtu.be/aSAwVqR22mo}}).}
	\label{fig_cover}
\end{minipage}
\vspace{-1cm}
}
\makeatother
\begin{document}
\maketitle
\begin{abstract}
	This paper proposes FAST-LIVO2: a fast, direct LiDAR-inertial-visual odometry framework to achieve accurate and robust state estimation in Simultaneous Localization and Mapping (SLAM) tasks and provide great potential in real-time, onboard robotic applications. FAST-LIVO2 fuses the IMU, LiDAR and image measurements, efficiently through an error-state iterated kalman filter (ESIKF). To address the dimension mismatch between the heterogeneous LiDAR and image measurements, we {use} a sequential update strategy in the Kalman filter. To enhance the efficiency, we {use} direct methods for both the visual and LiDAR fusion, where the LiDAR module registers raw points without extracting edge or plane features and {the visual module minimizes direct photometric errors without extracting ORB or FAST corner features. The fusion of both visual and LiDAR measurements is based on a single unified voxel map where the LiDAR module constructs the geometric structure for registering new LiDAR scans and the visual module attaches image patches to the LiDAR points (i.e., visual map points) enabling new image alignment. To enhance the accuracy of image alignment, we {use} plane priors from the LiDAR points in the voxel map (and even refine the plane prior in the alignment process) and update the reference patch dynamically after new images are aligned.}
    Furthermore, to enhance the robustness of image alignment, FAST-LIVO2 employs an on-demanding raycast operation and estimates the image exposure time {in real time}.
	We conduct extensive experiments on both benchmark and private datasets, demonstrating that our proposed system significantly outperforms other state-of-the-art odometry systems in terms of accuracy, robustness, and computation efficiency. Moreover, the effectiveness of key modules in the system is also validated. Lastly, we detail three applications of FAST-LIVO2: UAV onboard navigation demonstrating the system's computation efficiency for real-time onboard navigation, airborne mapping showcasing the system's mapping accuracy, and 3D model rendering (mesh-based and NeRF-based) underscoring the suitability of our reconstructed dense map for subsequent rendering tasks. We open source our code, dataset and application of this work on GitHub\footnote[1]{\url{https://github.com/hku-mars/FAST-LIVO2}} to benefit the robotics community.
\end{abstract}
\vspace{-0.13cm}
\begin{IEEEkeywords}
	Simultaneous Localization and Mapping (SLAM), Sensor Fusion, 3D Reconstruction, Aerial Navigation.
\end{IEEEkeywords}

\section{Introduction}
\vspace{-0.05cm}
\IEEEPARstart{I}{n} recent years, simultaneous localization and mapping (SLAM) technology has seen significant advancements, particularly in real-time 3D reconstruction and localization in unknown environments. Due to its ability to estimate poses and reconstruct maps in real time, SLAM has become indispensable for various robot navigation tasks. The localization process delivers crucial state feedback for the robot's onboard controllers, while the dense 3D map provides key environmental information, such as free spaces and obstacles, essential for effective trajectory planning. A colored map also carries substantial semantic information, enabling a vivid representation of the real world that opens up vast potential applications, such as virtual and augmented reality, 3D modeling, and robot-human interactions.

Currently, several SLAM frameworks have been successfully implemented with single-measurement sensors, primarily cameras \cite{mur2017orb, engel2017direct, engel2014lsd, forster2016svo} or LiDAR \cite{zhang2014loam, lin2020loam, shan2018lego}. Although visual and LiDAR SLAM have shown promise in their own domains, each has inherent limitations that constrain their performance in various scenarios.

Visual SLAM, leveraging cost-effective CMOS sensors and lenses, is capable of establishing accurate data associations, thereby achieving a certain level of localization accuracy. The abundance of color information further enriches the semantic perception. Further leveraging this enhanced scene comprehension, deep learning methods are employed for robust feature extraction and dynamic object filtering. However, the lack of direct depth measurement in visual SLAM necessitates concurrent optimization of map points via operations such as triangulation or depth filtering, which introduces significant computational overhead that often limits map accuracy and density. Visual SLAM also encounters numerous other limitations, such as varying measurement noise across different scales, sensitivity to illumination changes, and the impact of texture-less environments on data association.

LiDAR SLAM, utilizing LiDAR sensors, obtains precise depth measurements directly, offering superior precision and efficiency in localization and mapping tasks compared with visual SLAM. Despite these strengths, LiDAR SLAM exhibits several significant shortcomings. On one hand, the point cloud maps it reconstructs, albeit detailed, lack color information, thereby reducing their information scale. On the other hand, LiDAR SLAM performance tends to deteriorate in environments presenting insufficient geometric constraints, such as narrow tunnels, a single and extended wall, etc. 

As the demand to operate intelligent robots in the real world grows, especially in environments that often lack structure or texture, it is becoming clear that existing systems relying on a single sensor {cannot} provide the accurate and robust pose estimation as required. To address this issue, the fusion of commonly-used sensors such as LiDAR, camera, and IMU is gaining increasing attention. This strategy not only combines the strengths of these sensors to provide enhanced pose estimation, but also aids in the construction of accurate, dense, and colored point cloud maps, even in environments where the performance of {individual sensors degenerate}. 

Efficient and accurate LiDAR-inertial-visual odometry (LIVO) and mapping are still challenging problems: 1) The entire LIVO system is tasked with processing LiDAR measurements, consisting of hundreds to thousands of points per second, as well as high-rate, high-resolution images. The challenge of fully utilizing such a vast amount of data, particularly with limited onboard resources, necessitates exceptional computational efficiency; 2) Many existing systems typically incorporate a LiDAR-Inertial Odometry (LIO) subsystem and a Visual-Inertial Odometry (VIO) subsystem, each necessitating the extraction of features from visual and LiDAR data respectively to reduce computational load. In environments that lack structure or texture, this extraction process often results in limited feature points. Furthermore, to optimize feature extraction, extensive engineering adaptations are essential to accommodate the variability in LiDAR scanning patterns and point densities; 3) To reduce computational demands and achieve tighter integration between camera and LiDAR measurements, a unified map is essential to manage sparse points and the observed high-resolution image measurements simultaneously. {However, designing and maintaining such maps are particularly challenging considering the heterogeneous measurements of LiDAR and cameras;} 4) To ensure the {accuracy} of the reconstructed colored point cloud, pose estimation needs to achieve pixel-level accuracy. Meeting this standard presents considerable challenges: proper hardware synchronization, rigorous pre-calibration of extrinsic parameters between LiDAR and cameras, precise recovery of exposure time, and a fusion strategy capable of reaching pixel-level accuracy in real time.

Motivated by these issues, we propose FAST-LIVO2, a high-efficiency LIVO system that tightly integrates LiDAR, image and IMU measurements through a sequentially updated error-state iterated Kalman filter (ESIKF). With the prior from IMU propagation, the system state is updated sequentially, first by the LiDAR measurements and then by the image measurements, both utilizing direct methods based on a {single unified voxel map}. Specifically, {in the LiDAR update, the system registers raw points to the map to construct and update its geometric structure, and in the visual update, the system reuses LiDAR map points as the visual map points directly without extracting, triangulating}, or optimizing any visual features from images. The chosen visual map points in the map are attached with reference image patches previously observed and then projected to the current image to align its pose by minimizing the direct photometric errors (i.e., sparse image alignment). To improve the accuracy in the image alignment, FAST-LIVO2 dynamically updates the reference patches and {uses} the plane priors obtained from LiDAR points. For improved computation efficiency, FAST-LIVO2 {uses} LiDAR points to identify visual map points visible from the current image and conduct an on-demanding voxel raycast in case of no LiDAR points. FAST-LIVO2 also estimates the exposure time {in real time} to handle illumination variation. 

FAST-LIVO2 is developed based on FAST-LIVO first proposed in our previous work \cite{zheng2022fast}. 
The new contributions compared to FAST-LIVO are listed below: 
\begin{enumerate}	
        \item  {We propose an efficient ESIKF framework with sequential update to address the dimension mismatch between LiDAR and visual measurements, improving the robustness of FAST-LIVO that uses asynchronous updates.}
        \item  {We use (and even refine) plane priors from LiDAR points for improved accuracy. In contrast, FAST-LIVO assumes all pixels in a patch share the same depth, a wild assumption significantly reducing the accuracy of affine warping in image alignment.}
        \item  {We propose a reference patch update strategy to improve the accuracy of image alignment, by selecting high-quality, inlier reference patches that have large parallax and sufficient texture details. FAST-LIVO selects the reference patch based on proximity to the current view, often resulting in low-quality reference patches degrading the accuracy.}
        \item  {We conduct online exposure time estimation for handling environment illumination variation. FAST-LIVO did not address this issue, leading to poor convergence in image alignment under significant lighting changes.}
        \item  {We propose on-demand voxel raycasting to enhance the system robustness in the absence of LiDAR point measurements caused by LiDAR close proximity blind zones, an issue not considered in FAST-LIVO.}
\end{enumerate}

 {Each of the above contributions are evaluated in comprehensive ablation studies to verify their effectiveness.} We implement the proposed system as practical open software, meticulously optimized for real-time operation on both Intel and ARM processors. The system is versatile, supporting multi-line spinning LiDARs, emerging solid-state LiDARs with unconventional scanning patterns, as well as both pinhole cameras and various fisheye cameras.

Besides, we conduct extensive experiments on 25 sequences of public datasets (i.e., Hilti and NTU-VIRAL datasets), alongside various representative private datasets, enabling a comparison with other state-of-the-art SLAM systems (e.g., R3LIVE, LVI-SAM, FAST-LIO2, etc). Both qualitative and quantitative results demonstrate that our proposed system significantly outpaces other counterparts in terms of accuracy and robustness at a reduced computation cost. 

Taking a step further to underline the real-world applicability and versatility of our system, we deploy three distinctive applications. Firstly, fully onboard autonomous UAV navigation, demonstrating the system's real-time capabilities, marks a pioneering instance of employing a LiDAR-inertial-visual system for real-world autonomous UAV flights. Secondly, airborne mapping showcases the system's pixel-level precision under structure-less environments in practical use. Lastly, the high-quality generation of mesh, texturing, and NeRF models underscores the system's suitability for rendering tasks. We make our code and dataset available on GitHub.

\section{Related Works}\label{sec:related_work}

\subsection{Direct Methods}
Direct methods stand out as a prominent approach for fast pose estimation in both visual and LiDAR SLAM. Unlike feature-based methods \cite{qin2018vins, mur2015orb, zhang2014loam, lin2020loam} which necessitate the extraction of salient feature points (e.g., corners and edge pixels in images; plane and edge points in LiDAR scans) and the generation of robust descriptors for matching, direct methods directly leverage raw measurements to optimize the sensor pose \cite{driectmethod-1} by minimizing an error function based on photometric error or point-to-plane residuals, e.g., \cite{engel2014lsd, forster2014svo, 9697912, yuan2022efficient}. By eliminating the time-consuming feature extraction and matching, direct methods offer fast pose estimation. Nonetheless, the absence of feature matching requires fairly accurate state prior estimation to avoid local {minima}. 

Direct methods in visual SLAM can be broadly categorized into dense direct, semi-dense direct, and sparse direct methods. Dense direct methods, predominantly adopted for RGB-D cameras with full depth measurements as exemplified by \cite{meilland2011real,tykkala2011direct,kerl2013robust}, apply image-to-model alignment for pose estimation. In contrast, semi-dense direct methods \cite{engel2013semi, engel2014lsd} implement direct image alignment by capitalizing on pixels with significant {gray-level} gradients for estimation. Sparse direct methods \cite{forster2014svo, engel2017direct} focus on delivering accurate state estimation through only a few well-selected raw patches, thus further diminishing the computational burden in comparison to both dense and semi-dense direct methods. 

Unlike direct visual SLAM methods, direct LiDAR SLAM systems \cite{9697912, yuan2022efficient, chen2023direct, wang2022d} do not distinguish between dense and sparse approaches and commonly {use} spatially-downsampled or temporally-downsampled raw points in each scan to construct constraints for pose optimization.

In our work, we harness the principles of the direct method for both LiDAR and visual modules. The LiDAR module of our system is adapted from VoxelMap \cite{yuan2022efficient}, and the visual model is based on a variant of sparse direct method \cite{forster2014svo}. While drawing inspiration from sparse direct image alignment in \cite{forster2014svo}, our visual module differs by re-utilizing the LiDAR points as visual map points, thus mitigating the intensive backend computations (i.e., feature alignment, sliding window optimization and/or depth filtering). 

\subsection{LiDAR-visual(-inertial) SLAM}
The incorporation of multiple sensors in LiDAR-visual-inertial SLAM equips the system with the capability to handle a wide range of challenging environments, particularly when one sensor experiences failure or partial degeneration. Motivated by this, the research community has seen the emergence of various LiDAR-visual-inertial SLAM systems. 
{Existing methods can generally be divided into two categories: loosely coupled and tightly coupled. The classification can be determined from two perspectives: the state estimation level and the raw measurement level. At the state estimation level, the key is whether the estimate from one sensor serves as an optimization objective in another sensor's model. At the raw measurement level, it involves whether raw data from different sensors are combined.}

Zhang \textit{et al.} propose a LiDAR-Visual-Inertial SLAM system \cite{zhang2018laser} that is loosely coupled at the state estimation level. In this system, VIO subsystem only provides the initial pose for the scan registration in LIO subsystem, instead of being optimized jointly with the scan registration. VIL-SLAM \cite{shao2019stereo} employs a similar loosely-coupled method, not utilizing joint optimization of LiDAR, camera, and IMU measurements. 

Some systems (e.g., DEMO \cite{zhang2017real}, LIMO \cite{graeter2018limo}, CamVox \cite{zhu2021camvox}, \cite{huang2020lidar}) {use} 3D LiDAR points to provide depth measurements for the visual module \cite{mur2017orb, campos2021orb, forster2016svo}. While these systems exhibit measurement-level tight coupling, they remain loosely-coupled in state estimation, primarily due to the absence of constraints directly derived from LiDAR measurements at state estimation. Another issue arises as 3D LiDAR points do not have a one-to-one correspondence with 2D image feature points and/or lines due to mismatched resolutions. This mismatch requires interpolation in depth association, introducing potential errors. To address this, DVL-SLAM \cite{shin2020dvl} employs a direct method for visual tracking, wherein the LiDAR points are directly projected into the image to ascertain the depth of corresponding pixel positions.

The works mentioned above have not achieved tight coupling at the state estimation level. In pursuit of higher accuracy and robustness, many recent studies have emerged that jointly optimize sensor data in a tightly-coupled manner. To name a few, LIC-Fusion \cite{zuo2019lic} tightly fuses IMU measurements, sparse visual features, and LiDAR plane and edge features based on the MSCKF \cite{sun2018robust} framework. The subsequent LIC-Fusion2.0 \cite{zuo2020lic} enhances LiDAR pose estimation by implementing plane-feature tracking within a sliding window. VILENS \cite{wisth2021unified} offers a joint optimization of visual, LiDAR, and inertial data through a unified factor graph, relying on fixed lag smoothing. R2LIVE \cite{lin2021r2live} tightly fuses the LiDAR, camera and IMU measurements in an on-manifold iterated Kalman filter \cite{bell1993iekf}. For the VIO subsystem in R2LIVE, a sliding window optimization is used to triangulate the locations of visual features in the map. 

Several systems achieve complete tight coupling at both the measurement and state estimation levels. LVI-SAM \cite{shan2021lvi} fuses the LiDAR, visual and inertial sensors in a tightly-coupled smoothing and mapping framework, which is built atop a factor graph. The VIO subsystem performs visual feature tracking and extracts feature depth using LiDAR scans. R3LIVE \cite{lin2022r} constructs
the geometric structure of the global map by LIO and renders map texture by VIO. These two subsystems estimate the system state jointly by fusing their respective LiDAR or visual data with IMUs. The advanced version, R3LIVE++ \cite{lin2023r}, estimates exposure time {in real time} and conducts photometric calibration {in advance}\cite{engel2016photometrically}, which enables the system to recover the radiance of map points. Unlike most previously mentioned LiDAR-inertial-visual systems that rely on feature-based methods for both LIO and VIO subsystems, R3LIVE series \cite{lin2022r, lin2023r} adopt direct methods for both without feature extraction, enabling them to capture subtle environmental features even in texture-less or structure-less scenarios. 

Our system also jointly estimates the state using LiDAR, image and IMU data, and maintains a tightly-coupled voxel map at the measurement level. Furtheremore, our system uses direct methods, harnessing raw LiDAR points for LiDAR scan registration and employing raw image patches for visual tracking. The key difference between our system and R3LIVE (or R3LIVE++) is that R3LIVE (and R3LIVE++) operate at an individual pixel level in the VIO, while our system operates at image patch levels. This difference bestows our system with marked advantages. Firstly, in terms of robustness, our methodology {uses} a simplified, one-step frame-to-map sparse image alignment for pose estimation, mitigating the heavy reliance on an accurate initial state that is has to be obtained by a frame-to-frame optical flow in R3LIVEs. Consequently, our system simplifies and improves upon the two-stage frame-to-frame and frame-to-map operations in R3LIVE. Secondly, from a computational standpoint, the VIO in R3LIVE predominantly employs a dense direct method which is computationally expensive, necessitating extensive points for residual construction and rendering. In contrast, our sparse direct method provides enhanced computational efficiency. Lastly, our system exploits information at the resolution of raw image patches, whereas R3LIVE is capped at the resolution of its point map.



The visual module of our system is most similar to DV-LOAM \cite{wang2021dv}, SDV-LOAM \cite{yuan2023sdv} {and LVIO-Fusion \cite{10452777}}, which projects LiDAR points attached with patches into a new image and tracks the image by minimizing the direct photometric error. However, they have several key differences, such as the use of separate maps for vision and LiDAR, reliance on the assumption of constant depth in patch warping in the visual module, loosely-coupling at the state estimation level, and two stages of frame-to-frame and frame-to-keyframes for image alignment. In contrast, our system tightly integrates frame-to-map image alignment, LiDAR scan registration, and IMU measurements in an iterated Kalman filter. Besides, thanks to the single unified map for both LiDAR and visual modules, our system can directly employ the plane priors provided by the LiDAR points to accelerate the image alignment. 
\begin{table}[htp]
	\centering
	\caption{Some Important Notations}
	\label{tab:symbols}
	\begin{tabular}{lll}
		\toprule
		Notations  & Explanation \\
		\midrule
		$\boxplus$ / $\boxminus$ & The encapsulated “boxplus” and \\ &  \hspace{2px} “boxminus” operations on the state manifold \\
		${^G(\cdot)}$ & A vector ${(\cdot)}$ in global world frame \\
		${^C(\cdot)}$ & A vector ${(\cdot)}$ in camera frame\\
		$^I\mathbf{T}_L$ & The extrinsic of LiDAR frame w.r.t. IMU frame \\
		$^C\mathbf{T}_I$ &  The extrinsic of IMU frame w.r.t. camera frame \\
		$^G\mathbf T_{I}$ & The pose of IMU frame at time $k$ w.r.t. the global frame \\
		$\mathbf{x}, \widehat{\mathbf{x}}$, $\bar{\mathbf{x}}$ & The ground-truth, predicted and updated estimation of $\mathbf{x}$\\
		$\widehat{\mathbf{x}}^{\kappa}$ & The $\kappa$-th update of $\mathbf{x}$ \\
		$\bm{\delta} \mathbf{x}$ & The error state between ground-truth $\mathbf{x}$ and its estimation \\
		\bottomrule
	\end{tabular}
    \vspace{-0.5cm}
\end{table}
\section{System Overview}\label{sec:overview}
\begin{figure*}[t]
	\begin{center}
		{\includegraphics[width=2.04\columnwidth]{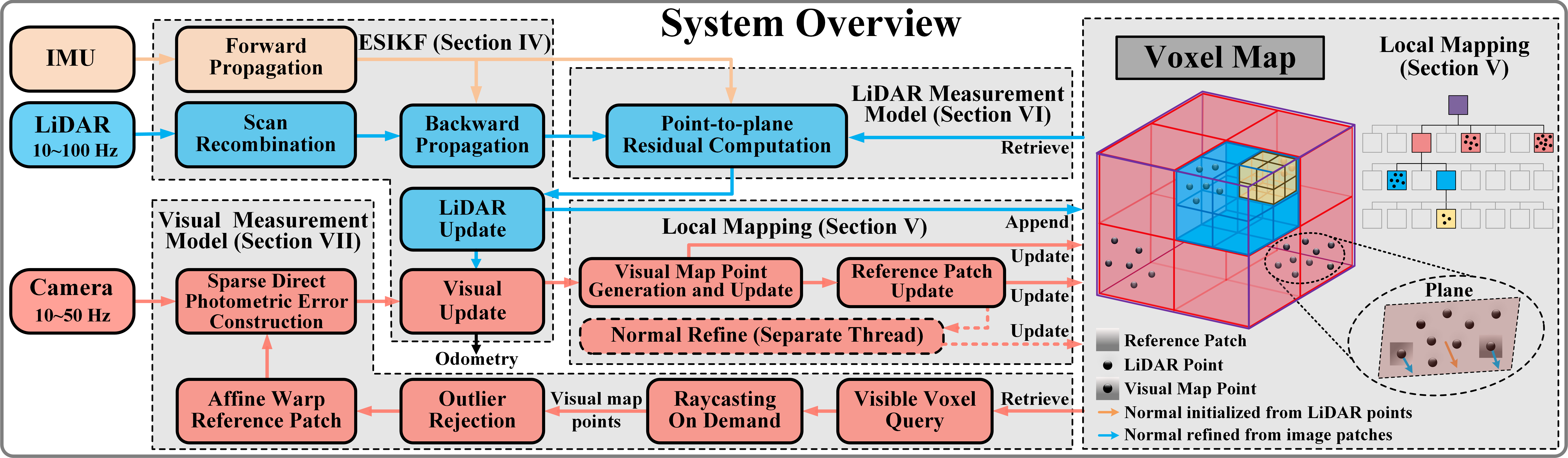}}
	\end{center}
	\vspace{-0.3cm}
	\caption{\label{fig:framework}System overview of FAST-LIVO2. }
	\vspace{-0.4cm}
\end{figure*}

{The overview of our system is shown in Fig. \ref{fig:framework}, which contains four sections: ESIKF (Section \ref{sec:KF}), Local Mapping (Section \ref{sec:local_mapping}), LiDAR Measurement Model (Section \ref{sec:lidar_section}), and Visual Measurement Model (Section \ref{sec:visual_section}).} 

The asynchronously-sampled LiDAR points are first recombined into scans at the camera's sampling time through scan recombination. Then, we tightly couple the LiDAR, image and inertial measurements via an ESIKF with sequential state update, where the system state is updated sequentially, first by the LiDAR measurements and then by the image measurements, both utilizing direct methods based on a single unified voxel map (Section \ref{sec:KF}). To construct {the} LiDAR measurement model in the ESIKF update (Section \ref{sec:lidar_section}), we compute the frame-to-map point-to-plane residual. To establish visual measurement model (Section \ref{sec:visual_section}), we extract the visual map points within the current FoV from the map, making use of visible voxel query and on-demand raycasting; after the extraction, {we identify and discard outlier visual map points} (e.g., points that are occluded or exhibit depth discontinuity); we then compute frame-to-map image photometric errors for visual update.

The local map for both visual and LiDAR updates is a voxel-map structure (Section \ref{sec:local_mapping}):
the LiDAR points construct and update the map's geometric structure, while the visual images append image patches to selected map points (i.e., visual map points) and update reference patches dynamically. The updated reference patches have their normal vectors further refined in a separate thread. 
\begin{figure}[htp]
	\begin{center}
		{\includegraphics[width=0.9\columnwidth]{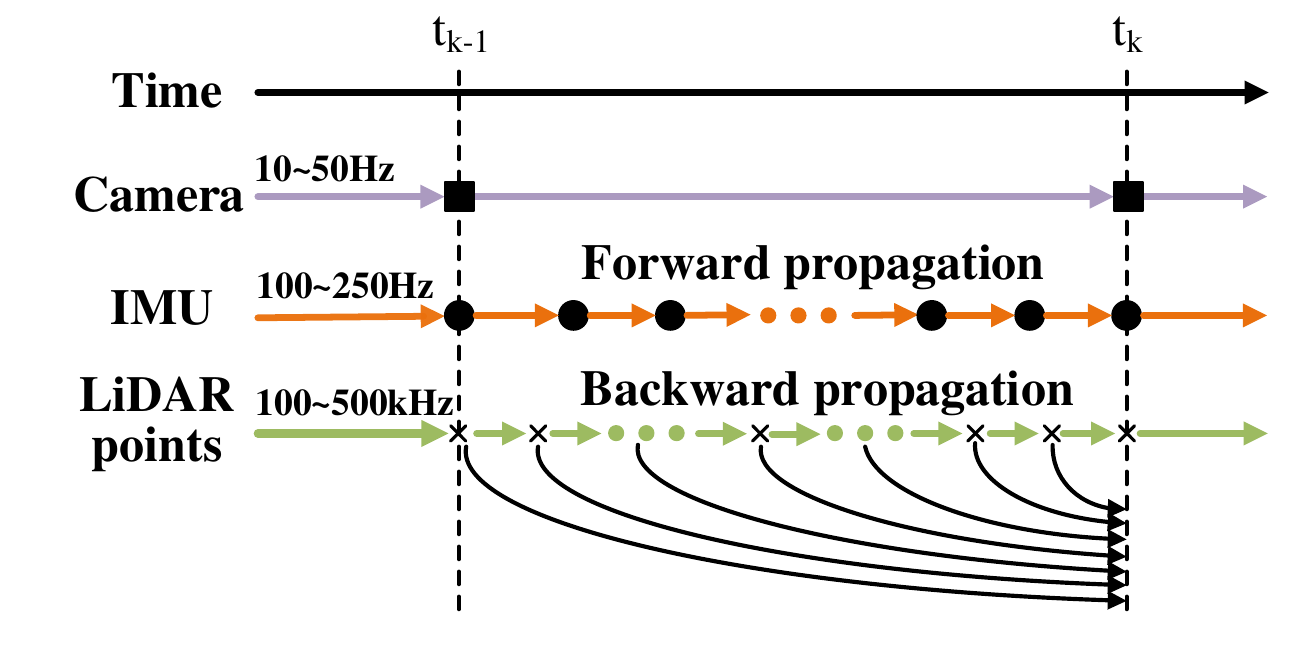}}
	\end{center}
	\vspace{-0.3cm}
	\caption{\label{fig:sequence}{Illustration of scan recombination, forward propagation and backward propagation applied to input data.}}
	\vspace{-0.5cm}
\end{figure}
\section{Error-State Iterated Kalman Filter with Sequential State Update}\label{sec:KF}
{This section outlines the system's architecture, based on the sequentially-updated Error-State Iterated Kalman Filter (ESIKF) framework. }
\vspace{-0.4cm}
\subsection{Notations and State Transition Model}\label{sec:notation}
In our system, we assume the time offsets among the three sensors (LiDAR, IMU and camera) are known, which can be calibrated or synchronized in advance. We take IMU frame (denoted as $I$) as the body frame and the first body frame as the global frame (denoted as $G$). Besides, we assume that the three sensors are rigidly attached and the extrinsic, defined in Table \ref{tab:symbols}, are pre-calibrated. Then, the discrete state transition model at the $i$-th IMU measurement is:
\begin{equation}
    \mathbf{x}_{i+1} = \mathbf{x}_{i} \boxplus \left(\Delta t\mathbf{f}\left(\mathbf{x}_i, \mathbf{u}_i, \mathbf{w}_i \right)\right)
    \label{eq_state_transition}
\end{equation}
where $\Delta t$ is the IMU sample period, the state $\mathbf{x}$, input $\mathbf{u}$, process noise $\mathbf w$, and function $\mathbf{f}$ are defined as follows:
\begin{equation}
\begin{aligned}
    \mathcal{M} &\triangleq SO(3) \times \mathbb{R}^{16},\ \text{dim}(\mathcal{M}) = 19  \\
    \mathbf{x} &\triangleq
    \begin{bmatrix}
        ^G\mathbf{R}_{I}^T & ^G\mathbf{p}_{I}^T & ^G\mathbf{v}_{I}^T & \mathbf{b}_{{g}}^T & \mathbf{b}_{{a}}^T & ^G\mathbf{g}^T & \tau
    \end{bmatrix}^T  \in \mathcal{M}  \\
    \mathbf{u} &\triangleq
    \begin{bmatrix}
        \boldsymbol{\omega}_{m}^T  & \mathbf{a}_{m}^T
    \end{bmatrix}^T, \hspace{0.2cm}
    \mathbf{w} \triangleq
    \begin{bmatrix}
        \mathbf{n}_{{g}}^T  & \mathbf{n}_{{a}}^T &
        \mathbf{n}_{\mathbf{b}_{g}}^T  & \mathbf{n}_{\mathbf{b}_{a}}^T  &
        {n_{\tau}}
    \end{bmatrix}^T \\
    \mathbf{f}&(\mathbf{x}, \mathbf{u}, \mathbf{w} ) = 
        \begin{bmatrix}
            \boldsymbol{\omega}_{m} - \mathbf{b}_{{g}} - \mathbf{n}_{{g}}  \\
            ^G\mathbf{v}_I+\frac{1}{2}(^G\mathbf{R}_{I}\left( \mathbf{a}_{m} - \mathbf{b}_{{a}} - \mathbf{n}_{{a}}\right)+{^{G}\mathbf{g}} )\Delta t \\
            ^G\mathbf{R}_{I}\left( \mathbf{a}_{m} - \mathbf{b}_{{a}} - \mathbf{n}_{{a}}\right) + {^{G}\mathbf{g}}  \\
            \mathbf{n}_{\mathbf{b}_{{g}}}\\
            \mathbf{n}_{\mathbf{b}_{{a}}} \\
            \mathbf{0}_{3\times 1} \\
            n_{\tau}
        \end{bmatrix}
\end{aligned}
\end{equation}
where {$^G\mathbf{R}_I$, $^G\mathbf{p}_I$, and $^G\mathbf{v}_I$ respectively denote the IMU attitude, position, and velocity in the global frame}, ${^G\mathbf{g}}$ is the gravity vector in the global frame, {$\tau$ is the inverse camera exposure time relative to the first frame}, $n_{\tau}$ is the Gaussian noise that models $\tau$ as a random walk, $\boldsymbol{\omega}_m$ and $\mathbf{a}_m$ are the raw IMU measurements, $\mathbf{n}_{g}$ and $\mathbf{n}_{a}$ are measurement noises in $\boldsymbol{\omega}_m$ and $\mathbf{a}_m$, $\mathbf{b}_{a}$ and $\mathbf{b}_{g}$ are IMU bias, which are modeled as random walk driven by Gaussian noise $\mathbf{n}_{\mathbf{b}_g}$ and $\mathbf{n}_{\mathbf{b}_a}$, respectively. 
\subsection{Scan Recombination} \label{subsec:scan_recombination}
We employ the scan recombination to segment the high-frequency, sequentially-sampled LiDAR raw points into distinct LiDAR scans at the camera sampling moments, as depicted in Fig. \ref{fig:sequence}. This ensures both camera and LiDAR data are synchronized at the same frequency (e.g., 10 Hz), allowing for update of state at the same time. 
\vspace{-0.3cm}
\subsection{Propagation} \label{subsec:propagation}
In the ESIKF framework, the state and covariance are propagated from time $t_{k-1}$, when the last LiDAR scan and image frame are received, to time $t_k$, when the current LiDAR scan and image frame are received. {This forward propagation predicts the state at each IMU input $\mathbf{u}_i$ during $t_{k-1}$ and $t_k$, by setting the process noise $\mathbf{w}_i$ in (\ref{eq_state_transition}) to zero.} {Denote the propagated state as $\widehat{\mathbf{x}}$ and covariance as $\widehat{\mathbf{P}}$, which will serve as a prior distribution for the subsequent update in } Section \ref{sec:KFupdate}.
Moreover, to compensate for motion distortion, we conduct a backward propagation as in \cite{xu2020fastlio}, ensuring points in a LiDAR scan are ``measured" at the scan-end time $t_k$. Note that for notation simplification, we omit the subscript $k$ in all state vectors.
\begin{algorithm}[t]
	\caption{Sequential State Update}
	\label{alg:state_estimation}
	Scan recombination to synchronize LiDAR and image data at the camera rate;\\
	Forward propagation to obtain state prediction $\widehat{\mathbf x}$ and its covariance $\widehat{\mathbf P}$;\\
	Backward propagation for LiDAR points motion compensation; \\
	\textcolor{gray}{\tcp{Point-to-plane LiDAR update}}
	$\kappa = -1$, $\widehat{\mathbf x}^{\kappa = 0} = \widehat{\mathbf x}$; \\
	\Repeat{$\| \widehat{\mathbf x}^{\kappa+1} \boxminus \widehat{\mathbf x}^\kappa\|< \epsilon$}{
		$\kappa =\kappa+1$;\\
		Compute residual $\mathbf z_{l}^{\kappa}$ and Jacobin $\mathbf H_{l}^{\kappa}$;\\
		Compute the state update $\widehat{\mathbf x}^{\kappa+1}$;\\}
	{$\widehat{\mathbf{x}}
	= \widehat{\mathbf x}^{\kappa+1}$, $\widehat{\mathbf P} = \left( \mathbf I - \mathbf K \mathbf H \right) {\widehat{\mathbf P}}$;} \\
	\textcolor{gray}{\tcp{Sparse direct visual update}}
	$level = -1$;\\
	\Repeat{$level >= 2$}{
		$\kappa = -1$, $\widehat{\mathbf x}^{\kappa = 0} = \widehat{\mathbf x}$;\\
		$level = level + 1$; \\
		\Repeat{$\| \widehat{\mathbf x}^{\kappa+1} \boxminus \widehat{\mathbf x}^\kappa\|< \epsilon$}{
			$\kappa =\kappa+1$;\\
			Compute residual $\mathbf z_{c}^{\kappa}$ and Jacobin $\mathbf H_{c}^\kappa$;\\
			Compute the state update $\widehat{\mathbf x}^{\kappa+1}$;\\}
		$\widehat{\mathbf x} = \widehat{\mathbf x}^{\kappa+1}$;\\
	}
	$\bar{\mathbf x} = \widehat{\mathbf x}^{\kappa+1}$; $\bar{\mathbf P} = \left( \mathbf I - \mathbf K \mathbf H \right) {\widehat{\mathbf P}}$. \\
\end{algorithm}
\vspace{-0.3cm}
\subsection{Sequential Update}\label{sec:KFupdate}
The IMU propagated state $\widehat{\mathbf{x}}$ and covariance $\widehat{\mathbf{P}}$ impose a prior distribution for $\mathbf{x}$, the system state at time $t_k$, as follows:
\begin{equation}
\label{eq:prior}
\mathbf{x} \boxminus \widehat{\mathbf{x}} 
\sim \mathcal{N}(\mathbf{0}, \widehat{\mathbf{P}})
\end{equation}

We denote the above prior distribution as $p(\mathbf x)$ and the measurement models for the LiDAR and camera as:
\begin{equation}
	\begin{split}~
	\label{eq:general_measure}
	\begin{bmatrix}
	\mathbf{y}_l \\
	\mathbf{y}_c
	\end{bmatrix} = 
	\begin{bmatrix}
		\mathbf{h}_l(\mathbf{x}, \mathbf{v}_l)\\
		\mathbf{h}_c(\mathbf{x}, \mathbf{v}_c)
	\end{bmatrix}
	\end{split}
\end{equation}
where $\mathbf{v}_l\sim\mathcal{N}(\mathbf{0}, \boldsymbol{\Sigma}_{\mathbf{v}_l})$ and $\mathbf{v}_c\sim\mathcal{N}(\mathbf{0}, \boldsymbol{\Sigma}_{\mathbf{v}_c})$ respectively denote the measurement noises for the LiDAR and camera.

A standard ESIKF \cite{he2023symbolic} would update the state $\mathbf x$ using all the current measurements, including both LiDAR measurement $\mathbf y_l$ and image measurements $\mathbf y_c$. However, LiDAR and image
 measurements are two different sensing modalities, whose data {dimensions do} not match. Furthermore, the fusion of image measurement may be performed at various levels of the image pyramid. 
  To address the dimension mismatch and give more flexibility for each module, we propose a sequential update strategy. {This strategy is theoretically equivalent to the standard update using all measurements, assuming statistical independence of LiDAR measurements $\mathbf y_l$ and image measurements $\mathbf y_c$ given the state vector $\mathbf{x}$ (i.e., measurements corrupted by statistically independent noise).}

To introduce the sequential update, we rewrite the total conditional distribution for the current state $\mathbf{x}$ as:
\begin{align}
	\begin{aligned}
		\label{eq:sequential_valid}
		p(\mathbf{x}\!\mid\!\mathbf{y}_l, \mathbf{y}_c) &\propto p(\mathbf{x}, \mathbf{y}_l, \mathbf{y}_c) = p(\mathbf{y}_c\!\mid\!\mathbf{x}, \mathbf{y}_l)p(\mathbf{x},\mathbf{y}_l) \\
		&= p(\mathbf{y}_c\!\mid\!\mathbf{x}) \underbrace{p(\mathbf{y}_l\!\mid\!\mathbf{x})p(\mathbf{x})}_{\propto p( \mathbf{x} \mid \mathbf{y}_l)}	
	\end{aligned}
\end{align}
Equation (\ref{eq:sequential_valid}) implies that the total conditional distribution $p(\mathbf{x}\hspace{-0.6em}\mid\hspace{-0.6em}\mathbf{y}_l, \mathbf{y}_c)$ can be obtained by two sequential Bayesian updates. The first step fuses only the LiDAR measurement $\mathbf{y}_l$ with the IMU-propagated prior distribution $p(\mathbf x)$ to obtain the distribution $p(\mathbf{x}\!\mid\!\mathbf{y}_l)$: 
\begin{equation}
		\label{eq:first}
			p(\mathbf{x}\!\mid\!\mathbf{y}_l) \propto p(\mathbf{y}_l\!\mid\!\mathbf{x})p(\mathbf{x})	
\end{equation}
The second step then fuses the camera measurement $\mathbf{y}_c$ with $p(\mathbf{x}\!\mid\!\mathbf{y}_l)$ to obtain the final posterior distribution of $\mathbf{x}$:
\begin{equation}
		\label{eq:second}
		p(\mathbf{x}\!\mid\!\mathbf{y}_l, \mathbf{y}_c) \propto p(\mathbf{y}_c\!\mid\!\mathbf{x})p(\mathbf{x}\!\mid\!\mathbf{y}_l)
\end{equation}

Interestingly, the two fusion in (\ref{eq:first}) and (\ref{eq:second}) follow the same form: 
\begin{equation}
		\label{eq:both}
		q(\mathbf{x}\!\mid\!\mathbf{y}) \propto q(\mathbf{y}\!\mid\!\mathbf{x})q(\mathbf{x})
\end{equation}
To conduct the fusion in (\ref{eq:both}) for either LiDAR or image measurements, we detail the prior distribution $q(\mathbf{x})$ and measurement model $q(\mathbf{y}\!\mid\!\mathbf{x})$ as follows. For the prior distribution $q(\mathbf{x})$, denote it as $\mathbf x = \widehat{\mathbf x} \boxplus \bm{\delta} \mathbf x$ with $\bm{\delta} \mathbf x \sim \mathcal{N}(\mathbf 0, \widehat{\mathbf P})$. In case of the LiDAR update (i.e., the first step), {$(\widehat{\mathbf x}, \widehat{\mathbf P})$} is the state and covariance obtained from the propagation step. In case of the visual update (i.e., the second step), {$(\widehat{\mathbf x}, \widehat{\mathbf P})$} is the converged state and covariance obtained from the LiDAR update.

To obtain the measurement model distribution $q(\mathbf{y}\!\mid\!\mathbf{x})$, denote
state estimated at the $\kappa$-th iteration as $\widehat{\mathbf {x}}^\kappa$, where $\widehat{\mathbf x}^0 = \widehat{\mathbf x}$. Approximating the measurement model (\ref{eq:general_measure}) (either the LiDAR or camera measurement) through its first-order Taylor expansion made at $\widehat{\mathbf {x}}^\kappa$ leads to:
\begin{align}
\label{e:derive_ekf1}
&\mathbf y \!\mid\!\mathbf{x} \simeq  \underbrace{\mathbf{h}(\widehat{\mathbf{x}}^{\kappa}, \mathbf{0})}_{{\mathbf z^\kappa}} + \mathbf{H}^{\kappa}\bm{\delta} \mathbf{x}^{\kappa} + \mathbf {L}^{\kappa} \, \mathbf{v} \\
\label{e:derive_ekf3}
&q(\mathbf{y}\!\mid\!\mathbf{x}) \simeq \mathcal{N}(\mathbf{h}(\widehat{\mathbf{x}}^{\kappa}, \mathbf{0}) + \mathbf{H}^{\kappa}\bm{\delta} \mathbf{x}^{\kappa}, \mathbf{R})
\end{align}
where $\bm{\delta} \mathbf{x}^\kappa = \mathbf x \boxminus \widehat{\mathbf {x}}^\kappa$, {$\mathbf z^\kappa$ is the residual,} $\mathbf{L}^{\kappa}\, \mathbf{v} \sim\mathcal{N}(\mathbf{0}, \mathbf{R})$ is the lumped measurement noise, $\mathbf{H}^{\kappa}$ and $\mathbf {L}^{\kappa}$ are the Jacobian matrixes of $\mathbf{h}(\widehat{\mathbf{x}}^{\kappa} \boxplus \bm{\delta} \mathbf{x}^{\kappa}, \mathbf{v})$ with respect to $\bm{\delta} \mathbf{x}^{\kappa}$ and $\mathbf v$, evaluated at zero, respectively. 

Then, substituting the prior distribution $q(\mathbf x)$ and the measurement distribution $q(\mathbf{y}\!\mid\!\mathbf{x})$ in (\ref{e:derive_ekf3}) into the posterior distribution (\ref{eq:both}) and {performing maximum likelihood estimation (MLE), we can obtain the maximum a-posterior estimation (MAP)} of $\bm{\delta} \mathbf x^{\kappa}$ (and hence $\mathbf x^{\kappa}$) from the standard update step in the ESIKF framework \cite{he2023symbolic}: 
\begin{equation}
	\begin{split}~\label{e:kalman_gain}
		\mathbf K &= \left((\mathbf H^{\kappa})^T {\mathbf R}^{-1} \mathbf H^{\kappa} + {\widehat{\mathbf P}}^{-1} \right)^{-1}(\mathbf H^{\kappa})^T \mathbf R^{-1}, \\
		\widehat{\mathbf x}^{\kappa+1} & = \widehat{\mathbf x}^{\kappa} \! \boxplus \!  \left( -\mathbf K  {\mathbf z}^\kappa  - (\mathbf I - \mathbf K \mathbf H^{\kappa} )\left( \widehat{\mathbf x}^{\kappa} \! \boxminus \widehat{\mathbf x} \right)  \right)
	\end{split}
\end{equation}
The converged state and covariance matrix then makes the mean and covariance of the posterior distribution $q(\mathbf x| \mathbf y)$. 

Kalman filter with sequential update has been investigated in the literature, such as in  \cite{willner1976kalman,ma2023globally}. This paper adopts this approach for ESIKF for LiDAR and camera systems. The implementation of the ESIKF with sequential update is detailed in {\bf Algorithm \ref{alg:state_estimation}}. In the first step (Line 6-10), the error state is updated from the LiDAR measurements (Section \ref{subsec:lidar_meas}) iteratively until convergence. The converged state and covariance estimates, {denoted again as $\widehat{\mathbf x}$ and $\widehat{\mathbf P}$}, are {used} to update the geometry of the map (Section \ref{subsubsec:mapconstruct}), and subsequently refined in the second step visual update (Line 13 - 23) on each level of the image pyramid (Section \ref{sec:sparse_direct_visual_model}) until convergence. The optimal state and covariance, denoted as $\bar{\mathbf x}$ and $\bar{\mathbf P}$, are employed for propagating incoming IMU measurements (Section \ref{subsec:propagation}) and update the visual structures of the map (Section \ref{subsubsec:Reference Patch} and \ref{subsubsec: normalrefine}). 
\section{Local Mapping}\label{sec:local_mapping}
\subsection{Map Structure}\label{subsec:data_structure}
Our map employs an adaptive voxel structure presented in \cite{yuan2022efficient}, which is organized by a Hash table and an octree for each Hash entry (Fig. \ref{fig:framework}). The hash table manages root voxels, each with a fixed dimension of $0.5 \times 0.5 \times 0.5$ meters. Each root voxel encapsulates an octree structure to further organize leaf voxels of varying sizes. A leaf voxel represents a local plane and stores a plane feature (i.e., plane center, normal vector, and uncertainty) along with a set of LiDAR raw points situated on this plane. Some of these points are attached with three-level image patches ($8\times8$ patch size), which we refer to as visual map points. Converged visual map points are only attached with reference patches, while non-converged ones are attached with reference patches and other visible patches (see Section \ref{subsubsec: normalrefine}). The varying size of the leaf voxel allows it to represent local planes of different scales, thus being adaptable to environments {with different structure} \cite{yuan2022efficient}. 

\begin{figure}[htp]
    \centering
    \vspace{-0.3cm}
    \includegraphics[width=1.0\columnwidth]{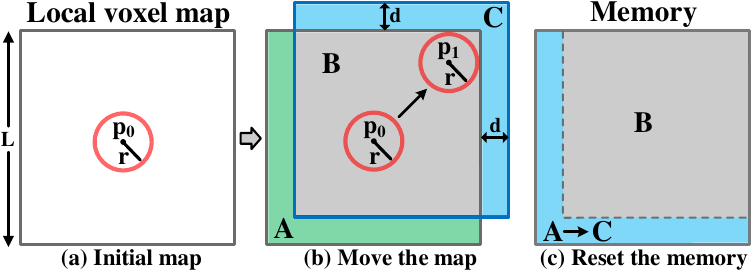}
    \caption{{2D demonstration of local map slide. In (a), the grey rectangle is the initial map region with length $L$. The red circle is the initial
    detection area centered at $\mathbf{p}_0$. In (b), the detection area moves to a new position $\mathbf{p}_1$ where the map boundaries are touched. The map region is moved to a new position (blue rectangle) by distance $d$. 
    In (c), the memory space B remains unchanged. The memory space A storing the green area is reset for the blue area C in (b).}}
    \label{fig:local_map}
    \vspace{-0.6cm}
\end{figure}

To prevent the size of the map from going unbound, we keep only a local map within a large local region of length $L$ around the LiDAR's current position, as illustrated in a 2D example in Fig. \ref{fig:local_map}. Initially, the map is a cube centered at the LiDAR's starting position $\mathbf{p}_0$. The LiDAR's detection area is visualized as a sphere centered at its current position, with its radius defined by the LiDAR's detection range.
When the LiDAR moves to a new position $\mathbf{p}_1$ where the detection area touches the boundaries of the map, we move the map away from the boundaries by a distance $d$. 
As the map moves, the memory containing the area moved out of the local map will be reset to store new areas moved in the local map. This ring-buffer approach ensures that our local map is maintained within a fixed size of memory. The implementation of the ring-buffer Hash map is detailed in \cite{ren2023rog}. The map move check is performed after each ESIKF update step.
\subsection{Geometry Construction and Update}\label{subsubsec:mapconstruct}
{The geometry of the map is constructed and updated from LiDAR point measurements. Specifically, after the LiDAR update in ESIKF (Section \ref{sec:KF}), we register all points from the LiDAR scan to the global frame. For each registered LiDAR point, we determine its located root voxel in the Hash map. If the voxel does not exist, we initialize the voxel with the new point and index it into the Hash map. If the determined voxel already exists in the map, we append the point to the existing voxel. After all points in a scan are distributed, we conduct geometry construction and update as follows.}

For newly-created voxels,
we determine if all the contained points lie on a plane based on the singular value decomposition. If so, we calculate {the center point $\mathbf q = \bar{\mathbf{p}}$}, plane normal $\mathbf n$, and the covariance matrix of $(\mathbf{q}, \mathbf n)$, denoted as $\boldsymbol{\Sigma}_\mathbf{n,\mathbf{q}}$, of the plane. $\boldsymbol{\Sigma}_\mathbf{n,\mathbf{q}}$ is used to characterize the plane uncertainty, which arises from both pose estimation uncertainty and the point measurement noise. The detailed plane criteria and calculation of the plane parameters and uncertainties can be referred to our previous work \cite{yuan2022efficient}. If the contained points do not lie on a plane, the voxel is continuously subdivided into eight smaller octants until either the points in the sub-voxel are determined to form a plane or the maximum layer (eg., 3) is reached. In the latter case, the points in the leaf voxel will be discarded. As a result, the map only contains voxels (either root or sub) identified as planes. 

For existing voxels that have new points appended, we assess if the new points still form a plane with the existing points in the root voxel or sub-voxel. If not, we conduct voxel sub-division as above. If yes, we update the plane's parameters ($\mathbf{q}$,$\mathbf n$) and covariance $\boldsymbol{\Sigma}_\mathbf{n,\mathbf{q}}$ as above too. Once the plane parameters converge (see \cite{yuan2022efficient}), the plane will be considered as mature and new points on this plane will be discarded. Moreover, mature planes will have their estimated plane parameters ($\mathbf{q}$,$\mathbf n$) and covariance $\boldsymbol{\Sigma}_\mathbf{n,\mathbf{q}}$ fixed. 

LiDAR points on planes (either in the root voxel or sub-voxel) will be used for generating visual map points in the subsequent section. For mature planes, 50 most recent LiDAR points are candidates for visual map point generation, while for unmature planes, all LiDAR points are the candidates. The visual map point generation process will identify some of these candidate points as visual map points and attach them with image patches for image alignment.  
\subsection{Visual Map Point Generation and Update}\label{subsubsection:generation}
To generate and update visual map points, we select the candidate LiDAR points in the map that (1) are visible from the current frame {(detailed in Section \ref{subsec:visualmapselect})}, and (2) exhibit significant {gray-level} gradients in the current image. 
We project these candidate points, after the visual update (Section \ref{sec:KFupdate}), onto the current image and retain the candidate point with the smallest depth for {the local plane in each voxel}. Then, we divide the current image into uniform {grid cells} each with $30\times30$ pixels. If a {grid cell does not contain any visual map point} projected here, we generate a new visual map point using the candidate point with the highest {gray-level} gradient and associate it with the current image patch, estimated current state (i.e., frame pose and exposure time), and the plane normal calculated from the LiDAR points as in the previous section. The patch attached to the visual map points has three layers of the same size (e.g., $11 \times 11$ pixels), each layer is half sampled from the previous layer, forming a patch pyramid.
{If a {grid cell} contains visual map points projected here, we add new patch (all three layers of pyramid) to the existing visual map point if (1) more than 20 frames have passed since its last patch addition, or (2) its pixel position in the current frame deviates by more than 40 pixels from its position at the last patch addition.  As a result, the map points will likely have effective patches with uniformly distributed viewing angles. Along with the patch pyramid, we also attach the estimated current state (i.e.,pose and exposure time} to the map point.

\subsection{Reference Patch Update}\label{subsubsec:Reference Patch}
A visual map point could have more than one patch due to the addition of new patches. We need to choose one reference patch for image alignment in the visual update. In detail, we score each {patch} $\mathbf f$ based on photometric similarity and viewing angle as follows: 
\begin{align}
    \label{eq:score}
    \text{NCC}(\mathbf{f}, \mathbf{g}) &= \frac{\sum_{x, y} [\mathbf{f}(x, y) - \bar{\mathbf{f}}] [\mathbf{g}(x, y) - \bar{\mathbf{g}}]}{\sqrt{\sum_{x, y} [\mathbf{f}(x, y) - \bar{\mathbf{f}}]^2 \sum_{x, y} [\mathbf{g}(x, y) - \bar{\mathbf{g}}]^2}} \notag \\ 
    c &= \frac{\mathbf{n}\cdot{\mathbf{p}} }{ \left \| \mathbf{p} \right \| }, \,\,\,\, \omega_1 = \frac{1}{1 + e^{\text{tr}(\boldsymbol{\Sigma}_\mathbf{n})}}  \\
    S &=  (1- \omega_1) \cdot{\frac{1}{n} \sum_{i=1}^n\text{NCC}(\mathbf f, \mathbf {g}_i)} + \omega_1 \cdot{c} \notag
\end{align}
where $\text{NCC}(\mathbf{f}, \mathbf{g})$ represents the Normalized Cross-Correlation (NCC) used to measure the similarity between patch $\mathbf f$ and $\mathbf g$ {at the $0$-th pyramid level (the level with the highest resolution) of both patch}, with mean subtraction applied to both patches, $c$ denotes the cosine similarity between the normal vector $\mathbf n$ and view direction $\mathbf p/\| \mathbf p \|$ of patch $\mathbf f$ under evaluation. When the patch is directly facing the plane where the map point is located, the value of $c$ is 1. The overall score $S$ is calculated by summing the weighted $\text{NCC}$ and $c$, where the former represents the average similarity between the patch $\mathbf f$ under evaluation and all other patches $\mathbf {g}_i$ and $\text{tr}(\boldsymbol{\Sigma}_\mathbf{n})$ represents the trace of the covariance matrix of the normal vector.

Among all the patches attached to a visual map point, the one with the highest score is updated as the reference patch. The above scoring mechanism tends to choose reference patches whose (1) appearance is similar (in terms of NCC) to most of the rest {of the} patches, a technique used by MVS \cite{stereopsis2010accurate} to avoid patches on dynamic objects; (2) view direction is orthogonal to the plane, thereby maintaining texture details at a high resolution.
In contrast, the reference patches update strategy in our previous work FAST-LIVO \cite{zheng2022fast} and prior arts \cite{forster2016svo} directly select the patch with the smallest view direction difference from the current frame, causing the selected reference patch to be very close to the current frame, hence imposing weak constraints on the current pose update. 
\begin{figure}[htp]
    \centering
    \vspace{-0.4cm}
    \includegraphics[width=1.0\columnwidth]{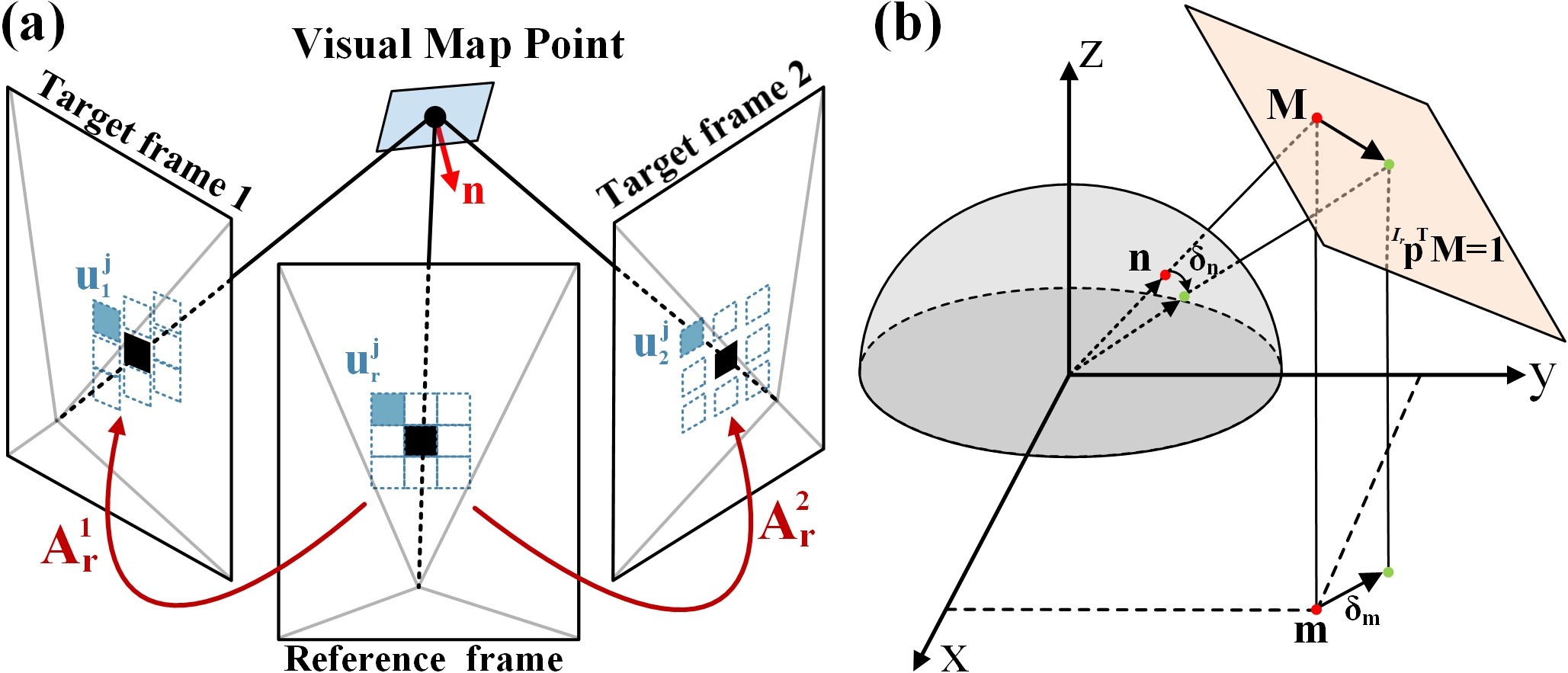}
    \vspace{-0.6cm}
    \caption{{(a) Affine warping between reference patches and target patches. (b) Any normal $^{I_r}\mathbf n \in \mathbb{S}^2$ lying on the normalized sphere is first projected into a point $\mathbf M \in \mathbb{R}^{3}$ on the plane ${^{I_r}\mathbf p}^T\mathbf M = 1 $, and then projected to a point $\mathbf m \in \mathbb{R}^2$ on the x-y plane.
    This transformation thereby converts a perturbation $\bm \delta \mathbf n$ on the sphere into a perturbation $\bm \delta \mathbf m$ in x-y plane. }}
    \label{fig:normal_opt}
\end{figure}
\vspace{-0.6cm}
\subsection{Normal Refine}\label{subsubsec: normalrefine}
Each visual map point is assumed to lie on a small local plane. Existing works \cite{forster2016svo, engel2017direct, zheng2022fast} assumed that {all pixels in a patch have the same depth, a wild assumption that does not hold in general.} We {use} plane parameters computed from the LiDAR points as detailed in Section \ref{subsubsec:mapconstruct} to achieve greater accuracy. This plane normal is crucial for performing affine warping for image alignment in the visual update process. 
To further enhance the accuracy of the affine warping, the plane normal could be further refined from the patches attached to the visual map point. Specifically, we refine the plane normal in the reference patch by minimizing the photometric error with respect to the {other} patches attached to the visual map point. 
\subsubsection{Affine warping}\label{subsec:affinewarp}
Affine warping is used to transform patch pixels from the reference frame (i.e., the source patch) to patch pixels in the rest {of the} frames (i.e., the target patch), illustrated in Fig. \ref{fig:normal_opt}(a). {Let $\mathbf{u}^j_r$ be the $j$-th pixel coordinates in the source patch and $\mathbf{u}^j_i$ be the $j$-th pixel coordinates in the $i$-th target patch.} Assuming all pixels in the patch lie in a local plane with normal ${^{I_r}\mathbf n}$ and visual map point position $^{I_r}{\mathbf p}$ {(which corresponds to the center pixel for both source and target patches)}, both represented in the source patch frame, we have: 
\begin{equation} \label{equ:warpaffine}
    \begin{aligned} 
        \mathbf{u}^j_i &= \mathbf {A}^{i}_r\mathbf{u}^j_r \\
        \mathbf {A}^{i}_r &= \mathbf{P}({^{I_i}\mathbf {R}}_{I_r}+{^{I_i}\mathbf {t}}_{I_r}\frac{1}{ {^{I_r}\mathbf n^T}\cdot{^{I_r}{\mathbf p}}} {^{I_r}\mathbf n^T})\mathbf{P}^{-1} 
    \end{aligned}
\end{equation}
where {$\mathbf{A}_r^i$ represents the affine warping matrix that transforms the pixel coordinates from the source (or reference) patch to the $i$-th target patch,} ${^{I_i}\mathbf {R}}_{I_r}$ and ${^{I_i}\mathbf {t}}_{I_r}$ denote the relative pose of the reference frame $I_r$ w.r.t. the target frame $I_i$. 
To use fisheye images directly without rectifying them to pinhole images, we implement projection matrix $\mathbf{P}$ and back projection matrix $\mathbf{P}^{-1}$ based on different camera models (e.g., $\mathbf P$ is the camera intrinsic matrix for the pinhole camera model).
\subsubsection{Normal Optimization}\label{subsec:normal_opt}
To refine the plane normal $^{I_r}\mathbf n$, we minimize photometric errors between the reference patch and other image patches {at the $0$-th pyramid level} (i.e., the highest-resolution level):
\begin{align}
        \label{eq:normal_opt}
        ^{I_r}\mathbf n^{*} &= \arg\min_{^{I_r}\mathbf n \in \mathbb{S}^2} \sum_{i \in S} \sum_{j=1}^{N^2} \left \| \tau_i\mathbf {I}_{i}(\mathbf A^{i}_r \mathbf u^j_r)-\tau_r\mathbf{I}_r(\mathbf u^j_r) \right \|_2 
\end{align}
where $N$ is the path size, {$\tau_r$ and $\tau_i$ are the inverse exposure times of the reference frame and the $i$-th target frame,} respectively.
$\mathbf{I}_r(\mathbf u^j_r)$ denote $j$-th patch pixel in the reference frame, $\mathbf{I}_i(\mathbf A^{i}_r \mathbf u^j_r)$ denotes the $j$-th path pixel in the $i$-th target frame, and $S$ is the set of all target frames. 

\subsubsection{Optimization Variable Transformation}\label{subsubsec:var_change}

To enhance the computational efficiency, we reparameterize the least squares problem in (\ref{eq:normal_opt}). 
Note that the optimization variable $^{I_r}\mathbf n$ only appeared in $\mathbf M \triangleq \frac{1}{ {^{I_r}\mathbf n^T}\cdot{^{I_r}{\mathbf p}}} {^{I_r}\mathbf n} \in \mathbb{R}^3$ in (\ref{equ:warpaffine}), the optimization over $^{I_r}\mathbf n$ can be conducted over $\mathbf M$. Moreover, the vector $\mathbf M$ is subject to constraint ${^{I_r}\mathbf p} \cdot \mathbf M = 1$, meaning that $\mathbf M$ can be parameterized as follows: 
\begin{equation}
\begin{aligned} \label{eq:transform}
        \mathbf M &= \begin{bmatrix}
            \mathbf M_x \\ 
            \mathbf M_y \\ 
            \frac{1}{{}^{I_r}\mathbf p_z}-\frac{{}^{I_r}\mathbf p_x}{{}^{I_r}\mathbf p_z} \mathbf M_x - \frac{{}^{I_r}\mathbf p_y}{{}^{I_r}\mathbf p_z} \mathbf M_y
        \end{bmatrix} 
        = \mathbf B \mathbf m + \mathbf b \\
        \mathbf B &= \begin{bmatrix}
            1 & 0 \\
            0 & 1 \\
            -\frac{{^{I_r}\mathbf p}_x}{{^{I_r}\mathbf p}_z} & -\frac{{^{I_r}\mathbf p}_y}{{^{I_r}\mathbf p}_z} 
        \end{bmatrix}, \mathbf b = \begin{bmatrix}
            0 \\
            0 \\
            \frac{1}{{^{I_r}\mathbf p}_z}
        \end{bmatrix},  \mathbf m=\begin{bmatrix}
        \mathbf{M}_x \\ \mathbf{M}_y
        \end{bmatrix} \in \mathbb{R}^2
    \end{aligned}
\end{equation}
where ${^{I_r}\mathbf p}_z \neq 0$ since no such reference patch could be chosen for the visual map point. The relation among $^{I_r}\mathbf n$, $\mathbf M$, and $\mathbf m$ are shown in Fig. \ref{fig:normal_opt} (b). 

Finally, the optimization in (\ref{eq:normal_opt}) is conducted over the vector $\mathbf m \in \mathbb{R}^2$ without any constraints. This optimization can be performed in a separate thread to avoid blocking the main odometry thread. The optimized parameter $\mathbf m^*$ can then be used to recover the optimal normal vector $^{I_r}\mathbf n^*$:

\begin{equation}
    ^{I_r}\mathbf n^* = \frac{\mathbf M^*}{\| \mathbf M^* \|}, \mathbf M^* = \mathbf B \mathbf m^* + \mathbf b
\end{equation}

Once the plane normal converges, the reference patch and normal vector for this visual map point are fixed without further refinement, and all other patches are deleted.

\section{LiDAR Measurement Model}\label{sec:lidar_section}
This section details the LiDAR measurement model $\mathbf{y}_l = \mathbf{h}_l (\mathbf x, \mathbf{v}_l)$ used in the LiDAR update of ESIKF in Section \ref{sec:KFupdate}.
\subsection{Point-to-plane LiDAR Measurement Model}\label{subsec:lidar_meas}
{After obtaining the undistorted points $\{{}^L\mathbf{p}_j\}$ in a scan}, we project them to the global frame using the estimated state $\widehat{\mathbf{x}}^{\kappa}$ at the $\kappa$-th iteration of LiDAR update:
	\begin{equation}\label{lidar_transformation}
		^G{\widehat{\mathbf{p}}^{\kappa}_j}={}^G{\widehat{\mathbf T}}^{\kappa}_{I} {}^{I}{\mathbf T}_{L} {}^L{\mathbf p}_j
	\end{equation}
We then identify the root or sub voxel where $^G{\widehat{\mathbf{p}}^{\kappa}_j}$ lies in the Hash map. If no voxel is found or the voxel does not contain a plane, the point is discarded. Otherwise, we use the plane in the voxel to establish a measurement equation for the LiDAR point. {Specifically, we assume the true LiDAR point $^L{\mathbf{p}^{gt}_j}$, given the accurate LiDAR pose $^G\mathbf T_I$, should lie on the plane with normal $\mathbf{n}^{gt}_j$ and center point ${{\mathbf{q}}^{gt}_j }$ in the voxel. i.e.,}
\begin{align} 
		\mathbf 0 
		&= ({\mathbf{n}_j^{gt}})^T ({}^G{\mathbf T}_{I} {}^{I}{\mathbf T}_{L} \, ^L{\mathbf p}_j^{gt} - \mathbf{q}_j^{gt}) 
\end{align}
Since the ground-true point $^L{\mathbf{p}^{gt}_j}$ is measured as $^L{\mathbf{p}_j}$ with ranging and bearing noises $\bm{\delta}{^{L}{\mathbf{p}_j}}$, we have $^L{\mathbf{p}^{gt}_j}={}^L\mathbf{p}_j-\bm{\delta}{^{L}{\mathbf{p }_j}}$. Likewise, the plane parameters $(\mathbf{n}^{gt}_j, {\mathbf{q}}^{gt}_j)$ are estimated as $(\mathbf{n}_j, {\mathbf{q}}_j)$ with covariance $\boldsymbol{\Sigma}_\mathbf{n,\mathbf q}$ (Section \ref{subsubsec:mapconstruct}), so we have: $\mathbf{n}_j^{gt} = \mathbf{n}_j \boxminus \bm{\delta} \mathbf{n}_j, \,\,\, {\mathbf{q}}_j^{gt} = {\mathbf{q}}_j - \bm{\delta}{\mathbf{q}_j}$.
Therefore,
\begin{align} \label{lidar_meas_model}
		\underbrace{\mathbf 0}_{\mathbf y_l } 
		&= \underbrace{{(\mathbf{n}_j  \boxminus  \bm{\delta} \mathbf{n}_j)}^T ({}^G{\mathbf T}_{I} {}^{I}{\mathbf T}_{L} ({}^L\mathbf {p}_j - \bm{\delta}{^{L}{\mathbf{p }_j}}) - ({\mathbf{q}}_j - \bm{\delta}{\mathbf{q}_j}))}_{\mathbf{h}_l(\mathbf{x}, \mathbf{v}_l) }
\end{align}
where the measurement noise $\mathbf{v}_l = (\bm{\delta}{^{L}{\mathbf{p }_j}}, \bm{\delta} \mathbf{n}_j, \bm{\delta}{\mathbf{q}_j})$ consisting of the noise associated with the LiDAR point, the normal vector, and the plane center respectively. 
\subsection{LiDAR Measurement Noise with Beam Divergence}\label{subsec:meas_noise}
\begin{figure}[htp]
	\begin{center}
		{\includegraphics[width=1.0\columnwidth]{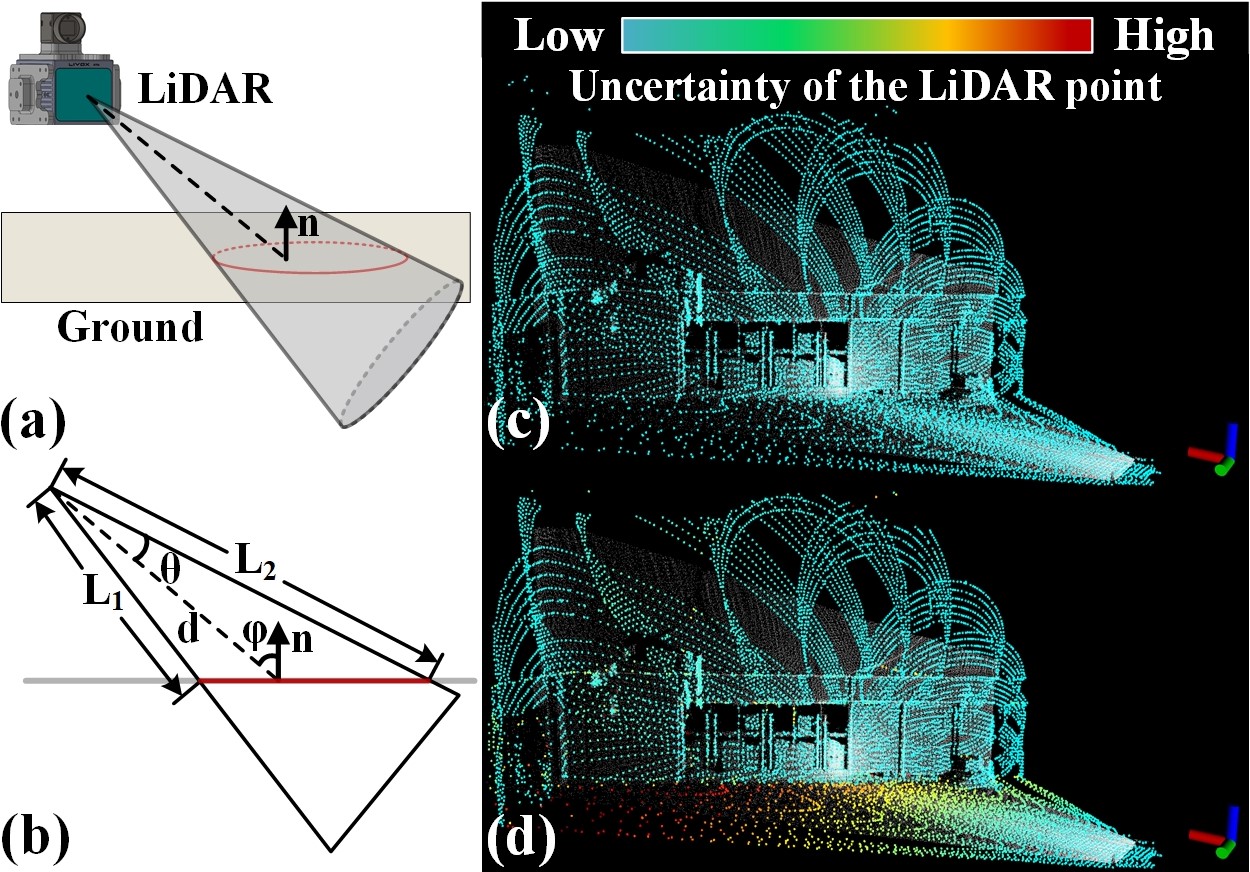}}
	\end{center}
	\vspace{-0.2cm}
	\caption{\label{fig:variance} (a) and (b) respectively illustrate the 3D and side cross-sectional views of the LiDAR point uncertainty model considering the laser beam divergence angle $\theta$. The red contour outlines the area that a laser beam spreads. (c) and (d) color the points in a scan by the point location uncertainty. Compared to (c), (d) further takes into account the ranging uncertainty $\delta d$ resulting from the beam divergence angle. This leads to a higher uncertainty for ground points due to the large spread area of laser beams.}
    \vspace{-0.3cm}
\end{figure}

The uncertainty of a LiDAR point $\bm{\delta}{^{L}{\mathbf{p}_j}}$ in the local LiDAR frame is decomposed into two components in \cite{yuan2022efficient}, the ranging uncertainty $\delta d$  caused by laser time of flight (TOF), and the bearing direction uncertainty $\bm{\delta} \bm{\omega}$ originated from encoders.
Besides these uncertainties, we also consider uncertainties caused by the laser beam divergence angle $\theta$, as illustrated in Fig. \ref{fig:variance}. As the angle $\varphi$ between the bearing direction and normal vector  increases, the ranging uncertainty of the LiDAR point increases significantly, while the bearing direction uncertainty remains unaffected. The $\delta d$ due to the laser beam divergence angle can be modeled as:
\begin{equation}
    \delta d = L_2 - L_1 = d\!\left( \frac{\cos \varphi}{\cos(\theta + \varphi)} - \frac{\cos \varphi}{\cos(\theta - \varphi)} \right)
\end{equation}
 Considering $\delta d$ influenced by TOF and laser beam divergence, {when our system selects more points from the ground or walls (see Fig. \ref{fig:variance} (c, d)),} it achieves a more precise pose estimation than that not considering such effect.
        \begin{figure}[htp]
            \begin{center}
                {\includegraphics[width=1.0\columnwidth]{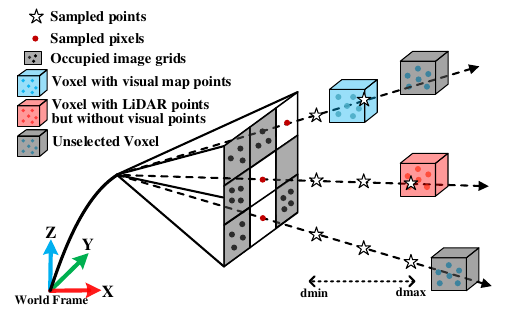}}
            \end{center}
            \vspace{-0.2cm}
            \caption{\label{fig:raycasting}{The illustration of on-demand voxel raycasting.}}
            \vspace{-0.2cm}
        \end{figure}
		\section{Visual Measurement Model}\label{sec:visual_section}
		{This section details the visual measurement model $\mathbf{y}_c = \mathbf{h}_c (\mathbf x, \mathbf{v}_c)$ used in the visual update of ESIKF in Section \ref{sec:KFupdate}.}
        \subsection{Visual Map Point Selection}\label{subsec:visualmapselect}
		To perform sparse image alignment in the visual update, we begin by selecting appropriate visual map points. We first extract the set of map points (termed as visual submap) that is visible in the current camera FoV, using voxel and raycasting queries. Then, the visual map points from this submap are selected and outliers are rejected. This process yields a refined set of visual map points ready for constructing visual photometric errors in the visual measurement model.
		\subsubsection{Visible Voxel Query}\label{subsubsec:voxelquery}
 Identifying map voxels within the current frame FoV is challenging due to the large number of voxels in the map. To address this issue, we poll the voxels hit by LiDAR points in the current scan. This can be done efficiently by inquiring the voxel Hash table using the measured point position. If the camera FoV is largely overlapped with the LiDAR FoV, map points in the camera FoV {likely} lie in these voxels as well. We also poll voxels hit by map points identified visible (through the same voxel query and raycasting) in the previous image frame, assuming that two consecutive image frames have large FoV overlaps. Finally, the current visual submap can be obtained as map points contained in these two types of voxels followed by a FoV check. 
		\subsubsection{Raycasting on Demand}\label{subsubsec:raycasting}
		In most cases, the visual submap can be obtained through the voxel queries above. However, a LiDAR sensor could return no points when it is too close to an object (known as the close proximity blind zones). Also, the camera FoV may not be completely covered by the LiDAR FoV. To recall more visual map points in these cases, we employ a raycasting strategy as illustrated in Fig. \ref{fig:raycasting}. We divide the image into uniform {grid cells}, each with $30 \times 30$ pixels, {and project the visual map points obtained from the voxel query onto the {grid cells}. For each image {grid cell} that is not occupied by these visual map points}, a ray is cast backward along the central pixel, where sample points are uniformly distributed along the ray in the depth direction from $d_{\min}$ to $d_{\max}$. In order to reduce computation load, the positions of sample points on each ray in the camera body frame are pre-computed. For each sampled point, we evaluate the corresponding voxel's status: {if the voxel contains map points that lie in this {grid cell} after projection, we incorporate these map points into the visual submap and cease for this ray. Otherwise,} we continue to the next sample point on the ray until reaching the maximum depth $d_{\max}$. After processing all the unoccupied image {grid cells} through raycasting, we obtain a set of visual map points that distribute among the whole image. 
		\begin{figure}[t]
			\centering
			\includegraphics[width=1.0\columnwidth]{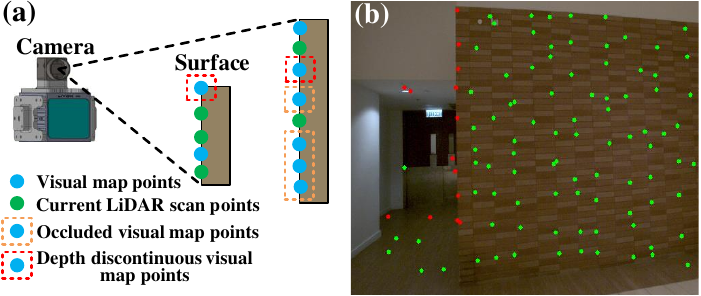}
			\caption{Outlier rejection. (a) shows the diagrammatic drawing of occluded and depth-discontinuous visual map points. (b) shows the effect of outlier rejection in real scenes. The red dots are the rejected visual map points, and the green dots are the accepted visual map points.}
			\label{fig:outlier}
               \vspace{-0.45cm}
		\end{figure}
		\subsubsection{Outlier Rejection}\label{subsubsec:outlier}
		{After voxel query and raycasting, we obtain all visual map points in the current frame FoV.}
      However, these visual map points could be occluded in the current frame, have discontinuous depth, have their reference patch taken at large view angles, {or have large view angles in the current frame}, all of which can severely degrade image alignment accuracy. To address the first issue, we project {all the visual map points in the submap} into the current frame using the pose after the LiDAR update and keep the lowest-depth points in each {grid cell} of $30 \times 30$ pixels. To address the second issue, we project the LiDAR points in the current LiDAR scan to the current frame producing a depth map. By comparing the depth of visual map points with their $9\times 9$ neighbor in the depth map, we determine their occlusion and depth variation. Occluded and depth-discontinuous map points are rejected (see Fig. \ref{fig:outlier}). {To address the third and fourth issue, we remove points where the view angle (i.e., the angle between normal vector and direction from the visual map point to the patch optical center) of the reference patch or current patch is too large (e.g. over 80°).}
        The remaining visual map points will be {used} to align the current image.
		\subsection{Sparse-Direct Visual Measurement Model} \label{sec:sparse_direct_visual_model}
		
  
        {The visual map points $\{^{G}{\mathbf{p}}_{i}\}$ extracted above are used to construct the visual measurement model. The underlying principle is that, when transforming the map point ${^{G}{\mathbf{p}}}_{i}$ to the current image $\mathbf I_k(\cdot)$ with the ground-truth state (i.e., pose) $\mathbf x_k$, the photometric error between the reference patch and the current patch should be zero:
\begin{align} \label{eq:meas_model_original}
\begin{split}~ 
    \mathbf{0} &=  \tau_k \mathbf I_k^{gt}(\underbrace{\boldsymbol{\pi}({^C\mathbf T_{I}}{(^G\mathbf T_{I})^{\scriptstyle -1}}{{}^{G}\mathbf{p}_i})}_{\mathbf{u}_i} + \Delta \mathbf{u}) \\ 
    &\ \ \ \ \ \ \ \ - \tau_r \mathbf I_r^{gt}(\underbrace{\boldsymbol{\pi}({^{C_r}\mathbf T_{G}}{^{G}\mathbf{p}_i})}_{\mathbf{u}'_{i}} + \mathbf{A}^r_i \Delta \mathbf{u}) 
\end{split}
\end{align}
where $\boldsymbol{\pi}(\cdot)$ is the common camera projection model (i.e., Pinhole, MEI, ATAN, Scaramuzza, Equidistant), $^{C_r}\mathbf T_{G}$ is the pose of the global frame $G$ w.r.t. reference frame $C_r$, which has been estimated when receiving and fusing the reference frame, {$\mathbf{A}^r_i$ is the affine warping matrix that transforms pixels from the $i$-th current patch to the reference patch}, $\Delta \mathbf{u}$ is the relative pixel position to the center $\mathbf{u}_i$ within the current patch, $\mathbf I_k^{gt}, \mathbf I_r^{gt}$ denote the ground-true pixel values of the reference and current frames, respectively. They are measured as the actual image pixel values $\mathbf I_k, \mathbf I_r$ with measurement noise $\mathbf{v}_c\!=\!(\bm{\delta}{\mathbf{I}}_k, \bm{\delta}{\mathbf{I}}_r)$, which originate from various sources (e.g., shot noise and the Analog-to-Digital Converter (ADC) noise of the camera CMOS). Hence, }
\begin{align}
\label{e:visual_meas_model}
    \underbrace{\mathbf{0}}_{\mathbf{y}_c} = \underbrace{\tau_k (\mathbf I_k({\mathbf{u}_i}\!+\!\Delta \mathbf{u})\!-\!\bm{\delta}{\mathbf{I}_k})\!-\!\tau_r (\mathbf I_r(\mathbf{u}'_{i}\!+\!\mathbf{A}^r_i \Delta \mathbf{u})\!-\!\bm{\delta}{\mathbf{I}_r})}_{\mathbf h_c(\mathbf x, \mathbf v_c)}   
\end{align}
{To enhance computational efficiency, we employ an inverse compositional formulation \cite{baker2004lucas, forster2016svo}, where the pose incremental $\bm{\delta}\mathbf{T} \in \mathbb{R}^{6}$ parameterizing $^G\mathbf T_{I} = ^G\!\!\widehat{\mathbf T}^{\kappa}_{I} \text{Exp}(\bm{\delta}\mathbf{T}) $ in $\mathbf u_i$ (see (\ref{eq:meas_model_original})), is moved from $\mathbf u_i$ to $\mathbf u'_i$ as follows:
\begin{align} 
    \begin{split}~ 
    \label{e:visual_meas_replace}
    \mathbf u_i &= \boldsymbol{\pi}({^C\mathbf T_{I}}{(^G\widehat{\mathbf T}^{\kappa}_{I})^{\scriptstyle -1}}{{}^{G}\mathbf{p}_i}) \\
    \mathbf u'_i &= \boldsymbol{\pi}({^{C_r}\mathbf T_{G}}\,\text{Exp}(\bm{\delta}\mathbf T)^{G}\mathbf{p}_i)
    \end{split}
\end{align}
Given that $\mathbf{u}'_i$ in the reference frame remains unchanged during each iteration, we only require a one-time computation of the Jacobian matrices w.r.t. $\bm{\delta}\mathbf{T}$, rather than re-calculating them for every iteration.}

To estimate the inverse exposure time $\tau_k$ from the measurement equation (\ref{e:visual_meas_model}), we fix the initial inverse exposure time $\tau_0 = 1$ to eliminate the degeneration of equation (\ref{e:visual_meas_model}) when all inverse exposure time are zeros. The estimated inverse exposure times of subsequent frames are therefore the exposure time relative to the first frame.
		 
   The equation (\ref{e:visual_meas_model}) is used in the visual update step across three levels (see {\bf Algorithm \ref{alg:state_estimation}}); the visual update starts from the coarsest level, after the convergence of a level, it proceeds to the next finer level. The estimated state is then used to generate visual map points (Section \ref{subsubsection:generation}) and update reference patch (Section \ref{subsubsec:Reference Patch}).
		\section{Datasets for Evaluation}\label{sec:dataset}
		In this section, we introduce datasets for performance evaluation, including public datasets NTU-VIRAL \cite{nguyen2022ntu}, Hilti'22 \cite{helmberger2022hilti}, Hilti'23 \cite{zhang2022hilti}, and MARS-LVIG \cite{li2024mars}, as well as our self-collected FAST-LIVO2 private dataset. Specifically, the NTU-VIRAL and Hilti datasets are {used} to conduct a quantitative benchmark comparison of our system against other state-of-the-art (SOTA) SLAM systems (Section \ref{subsec:benchmark}). The FAST-LIVO2 private dataset is primarily {used} to evaluate our system across various extremely challenging scenarios (Section \ref{subsec:challenge}), {to demonstrate its capability for high-precision mapping (Section \ref{subsec:high-precision}),} and to validate the functionality of the individual modules within our system {(Sections I-A through I-D in the Supplementary Material\cite{zheng2024supplementary})}. MARS-LVIG dataset is employed for application demonstrations (Section \ref{sec:application}) {and ablation study (Section I-E in the Supplementary Material\cite{zheng2024supplementary})}. 
		\subsection{NTU-VIRAL, Hilti and MARS-LVIG Dataset}\label{subsec:public}
 
		The NTU-VIRAL dataset, collected at the Nanyang Technological University campus using an aerial platform, presents diverse scenarios embodying unique aerial operational challenges. Specifically, the ``sbs'' sequences can only provide noisy visual features from distant objects. The ``nya'' sequences present challenges to LiDAR SLAM due to semi-transparent surfaces and to visual SLAM owing to intricate flight dynamics and low lighting conditions. The dataset is equipped with a 16-channel OS1 gen1\footnote[3]{\url{https://ouster.com/products/os1-lidar-sensor}} LiDAR sampled at 10 Hz and with a built-in IMU at 100 Hz, and two synchronized pinhole cameras triggered {at 10 Hz. The left camera is used for evaluation.}
		
		 The Hilti'22 and Hilti'23 datasets, collected by handheld and robot devices, encompass indoor and outdoor sequences from environments like construction sites, offices, labs and parking areas. These sequences introduce numerous challenges from long corridors, basements, and stairs, with textureless features, varying illumination conditions, and insufficient LiDAR plane constraints.
        Handheld sequences {use} a Hesai PandarXT-32\footnote[4]{\url{https://www.hesaitech.com/product/xt32/}} LiDAR at 10 Hz, five wide-angle cameras at 40 Hz, which is downsampled into 10 Hz, and an external Bosch BMI085 IMU at 400 Hz. Meanwhile, the robot-mounted sequences feature a Robosense BPearl\footnote[5]{\url{https://www.robosense.ai/en/rslidar/RS-Bpearl}} LiDAR at 10 Hz, eight omnidirectional cameras at 10 Hz, and an Xsens MTi-670 IMU at 200 Hz. {In both cases, the front-facing camera is used for all systems under evaluation.} Millimeter-accurate ground truth, obtained through a motion capture system (MoCap) or a Total Station \cite{klug2018measurement}, is provided for each sequence. Note that the ground truth of the Hilti datasets is not open-source; therefore, algorithmic results on these datasets are evaluated via the Hilti official website. Since ``Site 3'' in Hilti'23 does not provide in-depth analysis plots (e.g., RMSE), we exclude these four sequences, but our scoring results for these sequences can still be found on {their official website\footnote[6]{\url{https://hilti-challenge.com/leader-board-2022.html}, \url{https://hilti-challenge.com/leader-board-2023.html}}.} NTU-VIRAL and Hilti contribute a total of 25 sequences.
		
		The MARS-LVIG dataset provides high-altitude, ground-facing mapping data that encompasses diverse unstructured terrains such as jungles, mountains, and islands. The dataset was collected via a DJI M300 RTK quadrotor, which is equipped with a Livox Avia\footnote[7]{\url{https://www.livoxtech.com/avia}} LiDAR (with built-in BMI088 IMU) and a high-resolution global-shutter camera, both triggered at 10 Hz. This is notably distinct from the aforementioned NTU-VIRAL and Hilti datasets, which {use} $752\times480$ grayscale images, while the MARS dataset employs $2448\times2048$ RGB images, thereby facilitating the generation of clear, dense colored point clouds. Therefore, we leverage this public dataset to validate our capabilities in high-altitude aerial mapping applications.
    \begin{figure}[htp]
        \centering
        \vspace{-0.3cm}
        \includegraphics[width=1.0\columnwidth]{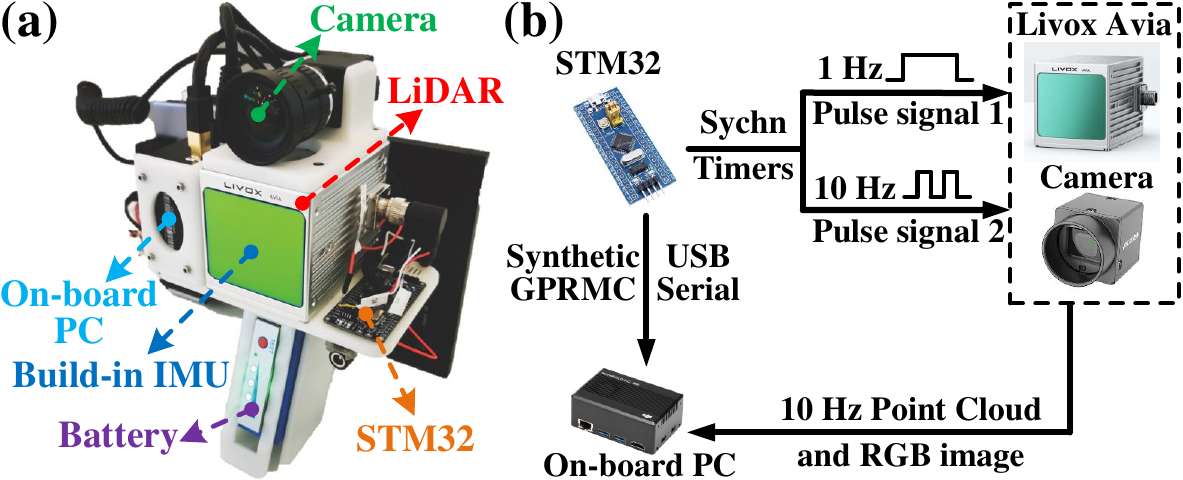}
        \caption{Our platform with hardware synchronization for data acquisition. (a) our handheld platform, (b) the hardware synchronization scheme.}
        \vspace{-0.5cm}
        \label{fig_handheld_device}
    \end{figure}
  \subsection{FAST-LIVO2 Private Dataset}\label{sec:private}
          {To validate the system's performance under more extreme conditions (e.g., LiDAR degeneration, low illumination, drastic exposure changes, and cases of no LiDAR measurements), we make a new dataset named FAST-LIVO2 private dataset. The dataset, hardware device, and hardware synchronization scheme are released with the codes of this work to facilitate the reproduction of our work.}
    \subsubsection{Platform}\label{sec:platform}
    Our data collection platform, illustrated in Fig. \ref{fig_handheld_device}, is equipped with an industrial camera (MV-CA013-21UC), a Livox Avia LiDAR, and a DJI manifold-2c (Intel i7-8550u CPU and 8 GB RAM) as onboard computers. The camera FoV is $70.6^{\circ}\times68.5^{\circ}$ and the LiDAR FoV is $70.4^{\circ}\times77.2^{\circ}$. All sensors are hard synchronized with a 10 Hz trigger signal, generated by STM32 synchronized timers. 
    \subsubsection{Sequence Description}
    As summarized in Table \ref{tab:dataset_setup} in the Supplementary Material\cite{zheng2024supplementary}, FAST-LIVO2 private dataset comprises 20 sequences across various scenes (e.g., campus buildings, corridors, basements, mining tunnel, etc.) characterized by structure-less, cluttered, dim, {variable-lighting}, and weakly textured environments, with a total duration of {66.9 min}. Most sequences exhibit visual and/or LiDAR degeneration, such as facing a single and/or texture-less plane, traversing an extremely narrow and/or dark tunnel, and experiencing {varying light conditions} from indoor to outdoor {(see Fig. \ref{sequence_discribe} in the Supplementary Material\cite{zheng2024supplementary})}. {To guarantee enhanced synchronous data collection between the camera and LiDAR, we configure the camera with fixed exposure time but auto-gain mode in most scenarios. For the remaining sequences with auto-exposure, we record their ground truth exposure times. In all sequences, the platform returns to the starting point, which enables the drift evaluation.}
		\section{Experiment Results}\label{sec:exp}
		{In this section, we conduct extensive experiments to evaluate our proposed system.}
\subsection{Implementation and System Configurations}\label{subsec:config}
    We implemented the proposed FAST-LIVO2 system in C++ and Robots Operating System (ROS). 
    In the default configuration, the exposure time estimation is enabled, while normal vector refinement is turned off. LiDAR points in a scan are downsampled temporally at a 1:3 ratio. The root voxel size for the voxel map is set at 0.5 m, and the maximum layer of the internal octree is 3. The image patch size is $8\times8$ for image alignment and $11\times11$ for normal refinement. Within the sequential ESIKF settings, for all the experiments, the camera photometric noise is set to a constant value of 100. The LiDAR depth error and bearing angle error are adjusted to 0.02 m and 0.05° for Livox Avia LiDAR and OS1-16, 0.001 m and 0.001° for PandarXT-32, 0.008 m and 0.01° for Robosense BPearl LiDAR. The laser beam divergence angle is set at 0.15° for Livox Avia LiDAR and OS1-16, and at 0.001° for the PandarXT-32 and Robosense BPearl LiDAR. Our system uses the same parameters in all sequences of all datasets with the same sensor setup. The computation platform for all experiments is a desktop PC equipped with an Intel i7-10700K CPU and 32 GB RAM. For FAST-LIVO2, we also test it on an ARM processor that is commonly used in embedded systems with reduced power and cost. The ARM platform is RB5\footnote[8]{\url{https://www.adlinktech.com/Products/Computer_on_Modules/SMARC/LEC-RB5?lang=en}}. with a Qualcomm Kryo585 CPU and 8 GB RAM. We refer to the implementation of FAST-LIVO2 on the ARM-based platform as ``FAST-LIVO2 (ARM)''.
\begin{table*}[htp]
    \caption{Absolute translational errors (RMSE, meters) in sequences}
    \centering
    \renewcommand\arraystretch{1.1}
    \setlength{\tabcolsep}{5.5pt}
    \begin{tabular}{clcccccccccc}
        \toprule
        Dataset & Sequence & \makecell{SDV-\\LOAM} & \makecell{Our\\LIO} & \makecell{FAST-\\LIO2} & R3LIVE & \makecell{LVI-\\SAM}  & \makecell{FAST-\\LIVO} & \makecell{Ours\\(w/o expo)} & \makecell{Ours\\(w normal)} & \makecell{Ours\\(w/o update)} & \makecell{Ours} \\ [1.0 ex]
        \hline
        \multirow{8}*{\textbf{Hilti'22}} 
        & Construction Ground & 25.121 & 0.011 & 0.013 & 0.021 & $\times$ & 0.022  & 0.011 & \textbf{0.008} & 0.015 & 0.010 \\
        & Construction Multilevel & 12.561 & 0.031 & 0.044 & 0.024 & $\times$ & 0.052  & 0.021 & \textbf{0.018} & 0.025 & 0.020 \\
        & Construction Stairs & 9.212 & 0.221 & 0.320 & 0.784 & 9.142 & 0.241 & 0.049 & 0.027 & 0.151 & \textbf{0.016} \\
        & Long Corridor & 19.531 & 0.061 & 0.064 & 0.061 & 6.312 & 0.065 & 0.069 & \textbf{0.059} & 0.071 & 0.067 \\
        & Cupola & 9.321 & 0.221 & 0.250 & 2.142 & $\times$ & 0.182 & 0.161 & 0.122 & 0.179 & \textbf{0.121} \\
        & Lower Gallery & 11.232 & 0.014 & 0.024 & 0.008 & 2.281 & 0.022 & 0.010 & 0.008 & 0.010 & \textbf{0.007} \\
        & Attic to Upper Gallery & 4.551 & 0.223 & 0.720 & 2.412 & $\times$ & 0.621 & 0.101 & 0.077 & 0.221 & \textbf{0.069} \\
        & Outside Building & 2.622 & 0.030 & \textbf{0.028} & 0.029 & 0.952 & 0.052 & 0.042 & 0.033 & 0.050  & 0.035 \\
        \hline
        \multirow{8}*{\textbf{Hilti'23}} 
        & Floor 0 & 4.621 & 0.028 & 0.031 & 0.018 & $\times$ & 0.021 & 0.025 & 0.023 & 0.023 & \textbf{0.022} \\
        & Floor 1 & 7.951 & 0.025 & 0.031 & 0.024 & 8.682 & 0.022  & 0.024 & \textbf{0.022} & 0.031 & 0.023 \\
        & Floor 2 & 7.912 & 0.041 & 0.083 & 0.046 & $\times$ & 0.048 & 0.023 & \textbf{0.021} & 0.051 & 0.022 \\
        & Basement & 6.151 & 0.021 & 0.038 & 0.024 & $\times$ & 0.035 & 0.020 & 0.018 & 0.018 & \textbf{0.016} \\
        & Stairs & 9.032 & 0.110 & 0.170 & 0.110 & 3.584 & 0.152  & 0.025 & 0.020 & 0.132 & \textbf{0.018} \\
        & Parking 3x floors down & 19.952 & 0.162 & 0.320 & 0.462 & $\times$ & 0.356 & 0.035 & \textbf{0.022} & 0.112 & 0.032 \\
        & Large room & 16.781 & 0.121 & 0.028 & 0.035 & 0.563 & 0.031 & 0.033 & 0.027 & 0.118 & \textbf{0.026} \\
        & Large room (dark) & 15.012 & 0.051 & \textbf{0.040} & 0.059 & $\times$ & 0.053 & 0.049 & 0.051 & 0.058 & 0.046 \\
        \hline
        \multirow{8}*{\textbf{NTU VIRAL}} 
        & eee\_01 & 0.301 & 0.122 & 0.212 & 0.072 & 3.901 & 0.191 & 0.069 & \textbf{0.066} & 0.109 & 0.068 \\
        & eee\_02 & 1.842 & 0.131 & 0.172 & 0.059 & 0.182 & 0.132 & 0.051 & 0.055 & 0.112 & \textbf{0.051} \\
        & eee\_03 & 0.301 & 0.124 & 0.213 & 0.078 & 0.287 & 0.192 & 0.068 & 0.070 & 0.099 & \textbf{0.068} \\
        & nya\_01 & 0.202 & 0.084 & 0.141 & 0.080 & 0.205 & 0.121 & 0.075 & 0.078 & 0.106 & \textbf{0.073} \\
        & nya\_02 & 0.214 & 0.153 & 0.212 & 0.084 & 1.296 & 0.182 & 0.076 & 0.081 & 0.118 & \textbf{0.075} \\
        & nya\_03 & 0.251 & 0.082 & 0.133 & 0.079 & 0.176 & 0.112 & 0.060 & 0.060 & 0.092 & \textbf{0.059} \\
        & sbs\_01 & 0.212 & 0.112 & 0.184 & 0.075 & 0.254 & 0.253 & 0.064 & 0.063 & 0.098 & \textbf{0.062} \\
        & sbs\_02 & 0.233 & 0.123 & 0.161 & 0.076 & 0.221 & 0.134 & 0.062 & \textbf{0.048} & 0.116 & 0.061 \\
        & sbs\_03 & 0.281 & 0.122 & 0.142 & 0.070 & 0.309 & 0.132 & 0.061 & \textbf{0.047} & 0.119  & 0.060 \\
        \hline
        \multirow{1}*{\textbf{Average}} 
        &  & 7.416 & 0.097 & 0.151 & 0.278 & 1.928 & 0.137 & 0.051 & \textbf{0.044} & 0.089 & 0.045 \\
        \toprule
    \end{tabular}
    \vspace{-0.1cm}
    \begin{tablenotes}
        \item[1] $\times$ denotes the system totally failed.
    \end{tablenotes}    
    \label{tab:ate_dataset}
\end{table*}
	\subsection{Benchmark Experiments}\label{subsec:benchmark}
		In this experiment, we conduct quantitative evaluations on 25 sequences from the NTU-VIRAL, Hilti’22, and 23 open datasets. Our approach is benchmarked against several state-of-the-art open-source odometry systems, including R3LIVE \cite{lin2022r}, a dense direct LiDAR-inertial-visual odometry system; FAST-LIO2 \cite{9697912}, a direct LiDAR-inertial odometry system; SDV-LOAM \cite{yuan2023sdv}, a semi-direct LiDAR-visual odometry system; LVI-SAM \cite{shan2021lvi}, a feature-based LiDAR-inertial-visual SLAM system; and our previous work FAST-LIVO \cite{zheng2022fast}.
		
		These systems are downloaded from their respective GitHub repositories. For FAST-LIO2, FAST-LIVO, and LVI-SAM, we {use} the recommended settings for indoor and outdoor scenes equipped with multi-line LiDAR sensors. For R3LIVE, we adapt the system to work with fisheye camera models and multi-line LiDARs equipped with external IMUs (the default configuration only supports internal IMUs). We disable the {real-time} optimization of the camera intrinsic and the extrinsic $^C\mathbf{T}_I$ due to adverse optimization caused by insufficient IMU excitation in the datasets. Other parameters, including the window size and pyramid level for optical flow tracking, the resolution for downsampling the point cloud of the current scan and the global map, are fine-tuned to achieve optimal performance. Since only the vision module of SDV-LOAM is open-sourced, we integrate it with LeGO-LOAM\cite{shan2018lego} in a loosely coupled manner, following the methodology described in the original paper\cite{yuan2023sdv}. This enhanced system continues to refine poses obtained from the vision module and we also open this implementation on GitHub\footnote[9]{\url{https://github.com/xuankuzcr/SDV-LOAM\_reimplementation}}. Given that all compared systems are odometry without loop closure, except for LVI-SAM, we remove the loop-closure module of LVI-SAM to ensure a fair comparison.
		Additionally, we conduct an ablation study on the exposure time estimation module, the normal refine module, {and the reference patch update strategy.} The default FAST-LIVO2 has {real-time} exposure estimation and reference patch update, but no normal refinement.
		
		The results of all methods are shown in Table \ref{tab:ate_dataset}. It is seen that our method achieves the highest overall accuracy across all sequences with an average RMSE of 0.044 m, which is three times more accurate than the second-place FAST-LIVO at 0.137 m. Our system delivers the best results in most sequences, except for "Outside Building" and "Large Room (dark)", where our system exhibits a slightly (millimeter-level) higher error compared to the LiDAR-inertial only odometry FAST-LIO2. This discrepancy can be attributed to the rich structural features but poor lighting conditions of these sequences, resulting in dim and blurred images. Consequently, fusing these low-quality images does not enhance odometry accuracy. 
		{Excluding these two sequences, our method, which leverages tightly-coupled LiDAR, inertial, and visual information, outperforms FAST-LIO2, our LIO subsystem, and the LiDAR-visual only odometry, SDV-LOAM, significantly.} Notably, SDV-LOAM performs particularly poorly on the Hilti datasets due to its lack of tight integration with IMU measurements, leading to drift in the LO subsystem. Additionally, the loose coupling between LiDAR and visual observations, along with poor initial values for VO, often results in local optima or even negative optimization.
		Our LIO subsystem generally surpasses FAST-LIO2 due to our more accurate noise modeling for each LiDAR point. In a few sequences where FAST-LIO2 outperforms slightly, the differences are minimal, at the millimeter level, and negligible.
		Moreover, our system's accuracy significantly exceeds that of other tightly coupled LiDAR-inertial-visual systems across all sequences. Among them, LVI-SAM fails in nine sequences primarily due to its feature-based LIO and VIO subsystems not fully utilizing raw measurements, which degrades its robustness in environments with subtle geometric or texture features. R3LIVE generally performs well, but struggles in ``Construction Stairs", ``Cupola", and ``Attic to Upper Gallery" sequences, where its performance is even worse than FAST-LIO2. This is because intense rotations at structure-less staircases result in inadequate pose priors, causing local optima when aligning colored map points with the current frame, and ultimately leading to negative optimization. FAST-LIVO and FAST-LIVO2 overcome such challenges in these sequences by the patch-based image alignments. Additionally, situations where the sensors are close to walls in these sequences highlight the effectiveness of raycasting in FAST-LIVO2, with the mapping results in these large-scale scenes shown {in Fig.~\ref{fig:hilti} in the Supplementary Material \cite{zheng2024supplementary}.} On the other hand, FAST-LIVO is outperformed by R3LIVE and FAST-LIVO2 on the NTU-VIRAL dataset, especially in unstructured scenes like ``nya" sequences, where the effects of affine warping based on constant depth assumptions are inaccurate. In contrast, the pixel-level alignment of R3LIVE and the plane prior (or refinement) of FAST-LIVO2 do not encounter such issues. 
  
    Comparing the different variants of FAST-LIVO2, we observe that the average accuracy without {real-time} exposure time estimation decreases by 6 mm compared to the default, as the exposure time estimation can actively compensate illumination changes in the environment. {On the other hand,} the average accuracy without the reference patch update decreases by 44 mm compared to the default, as the reference patch update strategy effectively selects patches with higher resolution and avoids selecting outlier patches. Finally, the normal refinement increases the average accuracy by 1 mm, and the accuracy improvement is not consistent in all sequences. The limited improvement is mainly because normal vector refinement yields positive optimization only in simple structured scenes with nice image observations. In the NTU-VIRAL dataset, images from the ``eee" and ``nya" sequences are extremely dim and blurry, where negative optimization is particularly severe. To further study the effectiveness of the different modules, including exposure time estimation, affine warping, reference patch update, normal convergence, on-demand raycasting{, and ESIKF sequential update}, we conducted a thorough study on our private dataset {and MARS-LVIG dataset}. The results are presented {in Section I (System Module Validation) in the Supplementary Material\cite{zheng2024supplementary} due to the space limit.} As confirmed in the results, {our system can achieve robust and accurate pose estimation in both structured and unstructured environments, under severe light variations, {in remarkably large-scale scenarios with long-term, high-speed data collection,} and even in extremely narrow spaces with few LiDAR measurements.}
\begin{figure*}[htp]
\begin{center}
    {\includegraphics[width=2.04\columnwidth]{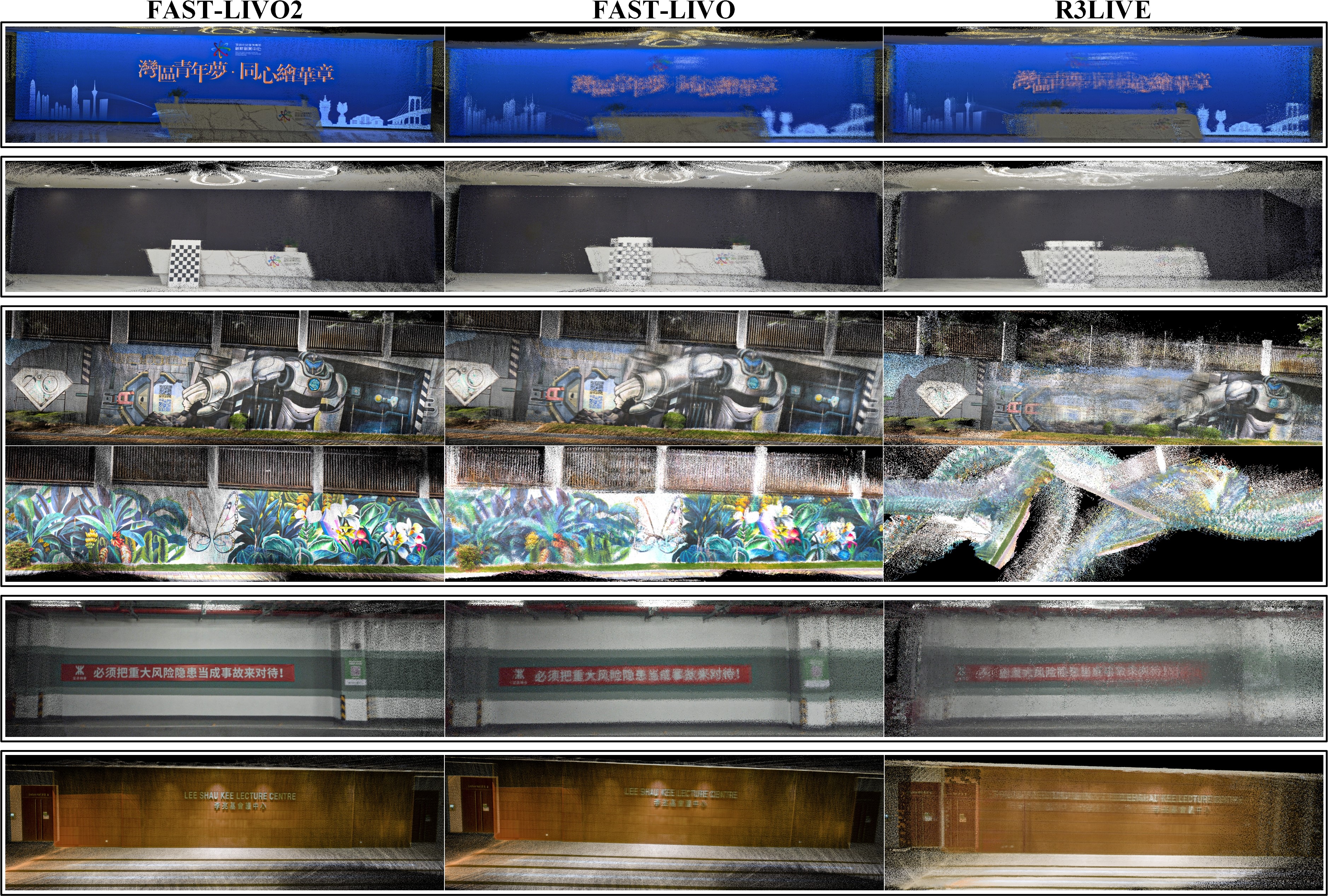}}
\end{center}
\vspace{-0.2cm}
\caption{\label{fig:degrade1}{The mapping results generated {in real time} in LiDAR degenerated scenes.
The point clouds from top to bottom correspond to ``Bright Screen Wall", ``Black Screen Wall", ``HIT Graffiti Wall" {(the third and fourth rows)}, ``Banner Wall", ``HKU Lecture Center", respectively, showing the comparison of colored point cloud constructed FAST-LIVO2, FAST-LIVO or R3LIVE (see more details on YouTube: \href{https://youtu.be/aSAwVqR22mo}{\tt{youtu.be/aSAwVqR22mo}}).}}
\vspace{-0.2cm}
\end{figure*}
\subsection{LiDAR Degenerated and Visually Challenging Environments}\label{subsec:challenge}
\begin{figure*}[htp]
    \begin{center}
            {\includegraphics[width=2\columnwidth]{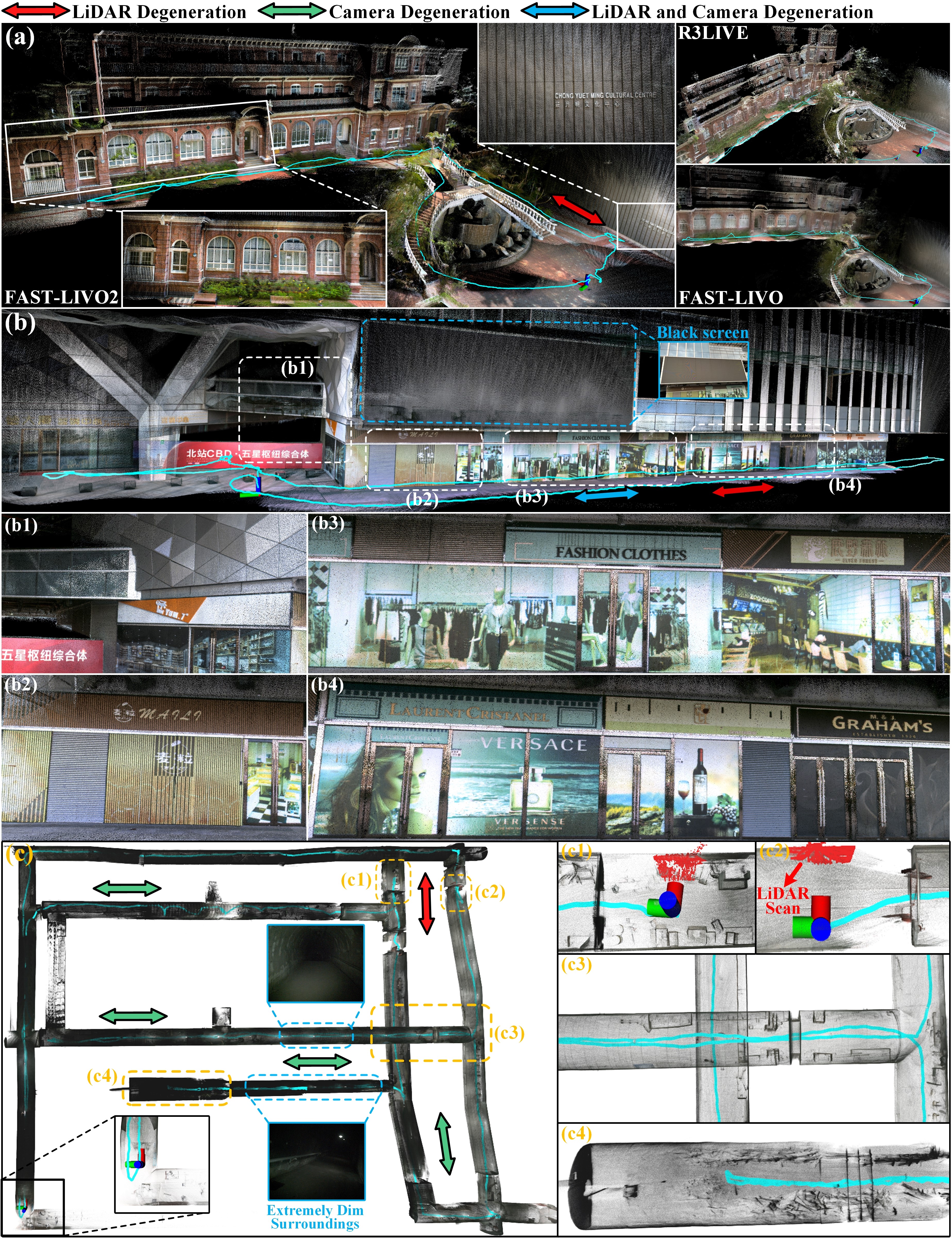}}
        \end{center}
        \vspace{-0.38cm}
        \caption{\label{fig:degrade2}{The mapping results generated {in real time} in complex LiDAR degenerated and visually challenging scenes. (a), (b), and (c) correspond to ``HKU Cultural Center", ``CBD Building 03", and ``Mining Tunnel", respectively. Different colored arrows indicate the directions of degeneration caused by different sensors (see more details on YouTube: \href{https://youtu.be/aSAwVqR22mo}{\tt{youtu.be/aSAwVqR22mo}}).} 
        }
\end{figure*}
{In this experiment, we evaluate the robustness of our system under environments experiencing LiDAR degeneration and/or visual challenges, comparing it with the qualitative mapping results of FAST-LIVO and R3LIVE in 8 sequences as shown in Fig. \ref{fig:degrade1} and \ref{fig:degrade2}.}
Fig. \ref{fig:degrade1} showcases LiDAR degeneration sequences where the LiDAR is facing a big wall while moving along the wall from one side to the other. Due to the absence of geometrical constraints since only one wall plane is being observed by the LiDAR, LIO methods would fail. 
{It's worth mentioning that the ``HIT Graffiti Wall" sequence spans nearly 800 meters with LiDAR continuously facing the wall, leading to considerable degeneration. In all sequences, FAST-LIVO2 distinctly showcases its robustness against even long-term degeneration and its capability to deliver high-precision colored point maps. In contrast, FAST-LIVO managed to obtain the geometric structure but with  completely blurred texture. R3LIVE struggles with both geometric structure and texture clarity. }
Fig. \ref{fig:degrade2} showcases tests in more complicated scenarios where LiDAR and/or camera both degenerate occasionally. The degeneration directions are indicated by respective arrows. ``HKU Cultural Center" (Fig. \ref{fig:degrade2} (a)) showcases the mapping results of FAST-LIVO2, R3LIVE, and FAST-LIVO. As can be seen, R3LIVE and FAST-LIVO have distorted point maps, blurred textures, and drifts exceeding 1 m. In contrast, FAST-LIVO2 successfully returns to the starting point, achieving an impressive end-to-end error of less than 0.01 m, while achieving a consistent point map with clear textures.
``CBD Building 03" (Fig. \ref{fig:degrade2} (b)) and ``Mining Tunnel" (Fig. \ref{fig:degrade2} (c)) display only FAST-LIVO2 results, as R3LIVE and FAST-LIVO failed. In Fig. \ref{fig:degrade2} (b), the blue arrow represents movement towards a pure black screen, indicating concurrent LiDAR and camera degeneration. In Fig. \ref{fig:degrade2} (c1) and (c2), the red points represent the LiDAR scan at that location, illustrating the areas of LiDAR degeneration due to the single plane being observed. Furthermore, the ``Mining Tunnel" exhibits very dim lighting throughout the whole sequence, coupled with frequent visual and LiDAR degeneration. {Despite of these challenges, FAST-LIVO2 still returns to the starting point with an end-to-end error of less than 0.01 m in both sequences.}
		\subsection{High-precision Mapping }\label{subsec:high-precision}
		In this experiment, we validate the high-precision mapping capabilities of our system. To explore the mapping accuracy across different algorithms and ensure fairness, we compare our system with FAST-LIO2, R3LIVE, and FAST-LIVO in scenes characterized by rich texture and structured environments. We take the sequences ``SYSU 01", ``HKU Landmark" and ``CBD Building 01" as examples. Fig. \ref{fig:degrade-non} in the Supplementary Material\cite{zheng2024supplementary} shows the colored point maps of these sequences reconstructed in real time. We can clearly observe that the point maps generated by FAST-LIVO2 retain the finest details among all the systems, the enlarged views of the colored point maps are akin to those in the actual RGB image. {In the ``SYSU 01" sequence}, our algorithm produces fewer white noise dots on the signboard because we normalize the image colors to a reasonable exposure time using the recovered exposure time before coloring, resulting in rarely overexposed colored point maps.  {The reconstruction of the human and motorcycle in ``CBD Building 01"} also exemplifies our ability to rebuild the details of unstructured objects. In all sequences, the estimated final position returns to the starting point with an end-to-end error of less than 0.01 m. We also tested FAST-LIVO2  in the remaining sequences of the prviate dataset, with mapping results shown in Fig. \ref{fig:hku_main_building}-\ref{fig:hkust_red} in the Supplementary Material\cite{zheng2024supplementary}.
		\subsection{Run Time Analysis}\label{subsec:runtime}
		In this section, we evaluate the average computational time per LiDAR scan and image frame of our proposed system, tested on a desktop PC equipped with an Intel i7-10700K CPU and 32 GB RAM. {Our evaluations span public datasets including Hilti'22, Hilti'23, and NTU-VIRAL, and our private dataset. As shown in Table \ref{tab:time_dataset}, our system exhibits the lowest processing time across all sequences. The average computation time consumption on an Intel i7 processor is only 30.03 ms {(17.13 ms per LiDAR scan and 12.90 ms per image frame)}, fulfilling real-time operation at 10 Hz. Besides, our system can even operate in real time on ARM processors with an average processing time per frame of just 78.44 ms.} 
       {LVI-SAM's LiDAR and visual feature extraction modules in LIO and VIO are time-consuming. In addition to the time consumed by LIO and VIO, LVI-SAM integrates IMU pre-integration constraints, visual odometry constraints, and LiDAR odometry constraints within a factor graph, further increasing the overall processing time.}
        For R3LIVE, although also employing a direct method, its pixel-wise image alignment necessitates the use of a large number of visual map points. In contrast, our approach {uses} sparse points with reference patches, enabling efficient alignment. Additionally, R3LIVE maintains a colored map that undergoes Bayesian updating, significantly increasing the computational load as the map resolution increases. {For FAST-LIO2, the average processing time per frame (Table \ref{tab:fast_livo_times} in the Supplementary Material\cite{zheng2024supplementary} due to space constraints) is approximately 10.35 ms less than FAST-LIVO2 due to not processing additional image measurements.}
  
		FAST-LIVO2 also shows noticeable improvements over the predecessor FAST-LIVO. The primary enhancement stems from our application of inverse compositional formulation in the sparse image alignment. Employing affine warping based on the plane prior from LiDAR points further enhances the convergence efficiency of our method. Consequently, FAST-LIVO2 reduces the number of iterations per pyramid level from $10$ to $3$, while still achieving superior accuracy. 
		\begin{table}[t]
			\caption{Processing time (ms) per LiDAR and image frame}
			\centering
                \scriptsize
			\renewcommand\arraystretch{1.1}
			\setlength{\tabcolsep}{2.3pt}
			\begin{tabular}{lllll@{\hspace{-6pt}}c}
				\toprule
				Dataset  & \makecell{R3LIVE}  & \makecell{LVI-\\SAM}  & \makecell{FAST-\\LIVO} & \makecell{FAST-LIVO2\\{(LiDAR / Image)}}& \makecell{FAST-LIVO2\\(ARM)} \\ [1.5 ex]
				\hline
				\specialrule{0em}{1pt}{1pt}
				\textbf{Hilti'22} & & & & &\\ 
				Construction Ground & 105.03 & $\times$ & 52.33 & \textbf{36.52 (20.44 / 15.05)} & 96.12\\
				Construction Multilevel & 112.13 & $\times$ & 56.12 & \textbf{38.77 (21.38 / 17.39)} & 95.38\\
				Construction Stairs & 125.41 & 138.47 & 51.34 & \textbf{39.33 (24.32 / 15.01)} & 98.43\\
				Long Corridor & 120.42 & 109.97 & 48.68 & \textbf{41.42 (26.21 / 15.21) }& 94.33\\
				Cupola & 151.52 & $\times$ & 59.42 & \textbf{43.54 (26.53 / 17.01)} & 98.12\\
				Lower Gallery & 119.74 & 131.37 & 51.15 & \textbf{41.11 (25.69 / 15.42)} & 92.13\\
				Attic to Upper Gallery & 144.39 & $\times$ & 58.61 & \textbf{44.21 (27.12 / 17.09)} & 91.15\\
				Outside Building & 105.17 & 107.92 & 44.25 & \textbf{33.82 (18.91 / 14.91)} & 85.43\\
				\specialrule{0em}{1pt}{1pt}
				\hline
				\specialrule{0em}{1pt}{1pt}
				\textbf{Hilti'23} & & & & &\\
				Floor 0 & 117.01& $\times$ & 52.23 & \textbf{42.22 (25.13 / 17.09)} & 92.24\\
				Floor 1 & 106.11 & 106.98 & 50.14 & \textbf{43.12 (27.43 / 15.69)} & 93.52\\
				Floor 2 & 154.65 & $\times$ & 53.24 & \textbf{41.78 (26.12 / 15.66)} & 94.43\\
				Basement & 118.21 & $\times$ & 48.23 & \textbf{39.65 (24.53 / 15.12)}& 93.42\\
				Stairs & 122.94 & 114.29 & 48.55 & \textbf{38.42 (22.64 / 15.78)}& 95.53\\
				Parking 3x floors down & 142.43 & $\times$ & 51.89 & \textbf{43.62 (26.99 / 16.63)} & 94.22\\
				Large room & 125.78 & 182.18 & 55.23 & \textbf{43.23 (26.43 / 16.80)} & 93.28\\
				Large room (dark) & 131.31 & $\times$ & 51.46 & \textbf{44.21 (29.12 / 15.09)} & 92.29\\
				\specialrule{0em}{1pt}{1pt}
				\hline
				\specialrule{0em}{1pt}{1pt}
				\textbf{NTU VIRAL} & & & & &\\
				eee\_01 & 105.89 & 113.61 & 38.22 & \textbf{31.45 (17.24 / 14.21)} & 78.03\\
				eee\_02 & 112.27 & 119.19 & 39.43 & \textbf{30.24 (16.23 / 14.01)} & 79.22\\
				eee\_03 & 108.11 & 108.77 & 37.53 & \textbf{29.44 (16.12 / 13.32)} & 77.43\\
				nya\_01 & 122.73 & 124.54 & 39.66 & \textbf{34.52 (18.14 / 16.38)} & 79.44\\
				nya\_02 & 111.96 & 117.18 & 35.11 & \textbf{32.23 (17.33 / 14.90)} & 80.52\\
				nya\_03 & 115.96 & 111.24 & 38.42 & \textbf{33.42 (18.12 / 15.30)} & 78.95\\
				sbs\_01 & 112.77 & 115.67 & 39.23 & \textbf{31.92 (14.68 / 17.24)} & 78.32\\
				sbs\_02 & 113.01 & 110.37 & 35.39 & \textbf{32.56 (17.62 / 14.94)} & 72.43\\
				sbs\_03 & 119.91 & 120.73 & 37.41 & \textbf{33.62 (18.02 / 15.60)} & 75.56\\
				\hline
				\specialrule{0em}{1pt}{1pt}
				\textbf{Private Dataset} & & & & &\\
				Retail Street & 85.32 & 70.23 & 31.22 & \textbf{19.44 (10.42 / 9.02)} & 64.12\\
				CBD Building 01 & 79.22 & 65.32 & 29.14 & \textbf{18.22 (10.32 / 7.90)} & 62.33\\
				CBD Building 02 & $\times$ & $\times$ & $\times$ & \textbf{21.53 (11.50 / 10.03)} & 69.98\\
				CBD Building 03 & $\times$ & $\times$ & $\times$ & \textbf{21.89 (11.98 / 9.91)} & 69.92\\
				HKU Landmark & 75.43 & 68.77 & 28.33 & \textbf{17.42 (10.12 / 7.30)} & 63.21\\
				HKU Lecture Center & 70.42 & $\times$  & 31.45 & \textbf{18.12 (10.99 / 7.13)} & 62.88\\
				HKU Centennial Garden & 74.33 & 70.12 & 33.46 & \textbf{19.22 (10.80 / 8.42)} & 64.52\\
				HKU Cultural Center & 102.34 & $\times$  & 35.62 & \textbf{21.43 (11.42 / 10.01)} & 72.43\\
				HKU Main Building & 75.62 & $\times$  & 30.21 & \textbf{19.43 (10.33 / 9.10)} & 65.31\\
				HKUST Red Sculpture & 84.22 & 90.13 & 32.43 & \textbf{20.14 (11.12 / 9.02)} & 65.42\\
				HIT Graffiti Wall & 112.56 & $\times$  & 39.98 & \textbf{21.71 (10.55 / 11.16)} & 68.99\\
				Banner Wall & 74.13 & $\times$  & 30.22 & \textbf{19.32 (10.01 / 9.31)} & 63.43\\
				Bright Screen Wall & 73.23 & $\times$  & 28.43 & \textbf{18.22 (9.92 / 8.30)} & 61.21\\
				Black Screen Wall & 71.42 & $\times$  & 27.66 & \textbf{17.99 (9.10 / 8.89)} & 60.91\\
				Office Building Wall & $\times$ & $\times$ & $\times$ & \textbf{19.42 (10.11 / 9.31)} & 62.33\\
                Narrow Corridor & $\times$ & $\times$ & $\times$ & \textbf{18.44 (8.32 / 10.12)} & 61.42 \\
				Long Corridor &  $\times$ & $\times$ & $\times$ & \textbf{19.53 (10.43 / 9.10)} & 62.55\\
				Mining Tunnel & $\times$ & $\times$ & $\times$ & \textbf{29.43 (15.42 / 14.01)} & 79.32\\
                SYSU 01 & 112.66 & 88.95 & 32.23 & \textbf{23.65 (12.51 / 11.14)} & 77.43\\
                SYSU 02 & 110.12 & $\times$ & 32.12 & \textbf{22.63 (11.82 / 10.81)} & 72.31\\
				\specialrule{0em}{1pt}{1pt}
                    \hline
                    \textbf{Average} & 108.36 & 108.45 & 41.43 & \textbf{30.03 (17.13 / 12.90)} & 78.44\\
				\toprule
			\end{tabular}
                \vspace{-0.1cm}
                \begin{tablenotes}
                \item[1] $\times$ denotes the system totally failed.
                \end{tablenotes}    
                \vspace{-0.15cm}
			\label{tab:time_dataset}
		\end{table}
		\section{Applications}\label{sec:application}
		{To showcase the superior performance and versatility of FAST-LIVO2 in real-world applications, we develop multiple solutions, including fully onboard autonomous UAV navigation, airborne mapping, textured mesh generation, and 3D Gaussian splatting reconstruction for 3D scene representation.}
		\begin{figure}[htp]
			\centering
			\includegraphics[width=1.0\columnwidth]{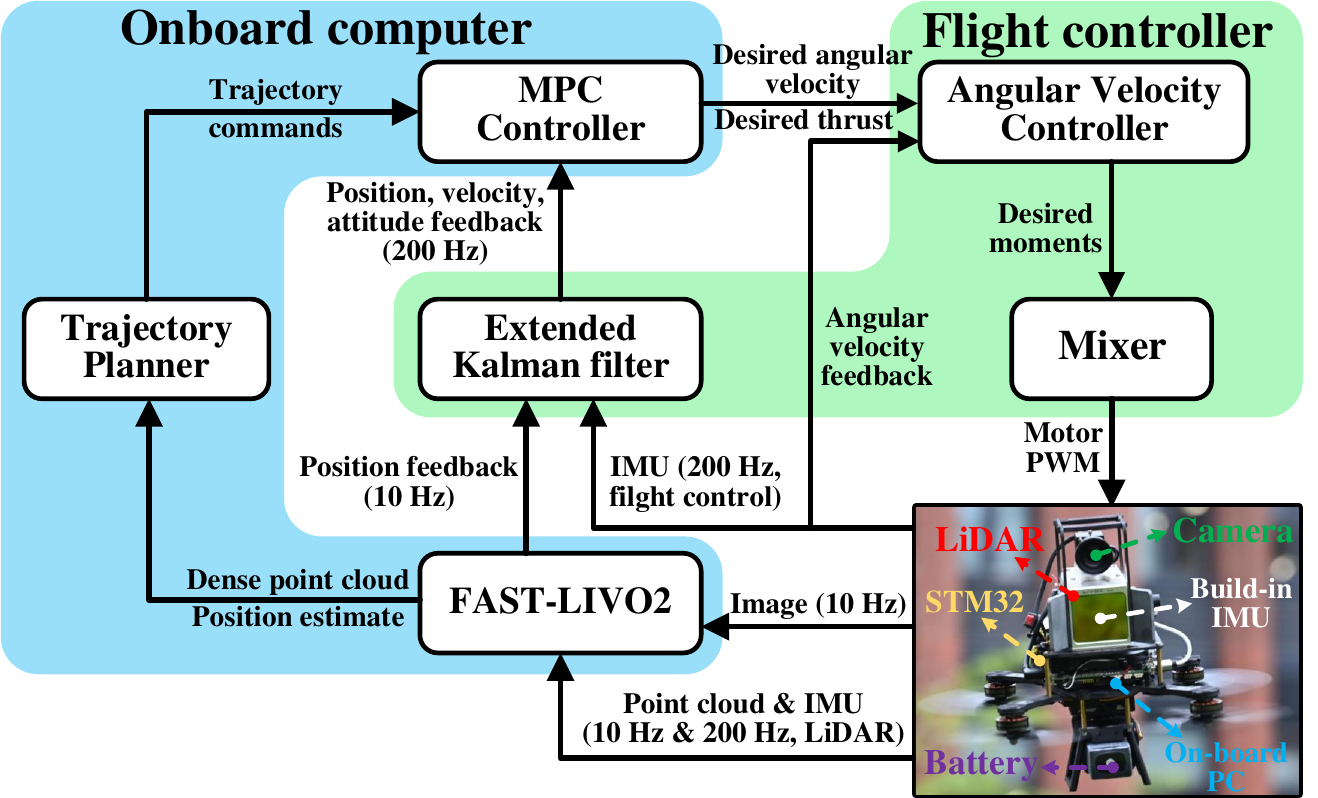}
               \vspace{-0.5cm}
			\caption{Fully onboard UAV navigation algorithm flowchart.}
			\label{fig:uav}
            \vspace{-0.25cm}
		\end{figure}
  	    \begin{figure}[htp]
			\centering
			\includegraphics[width=1.0\columnwidth]{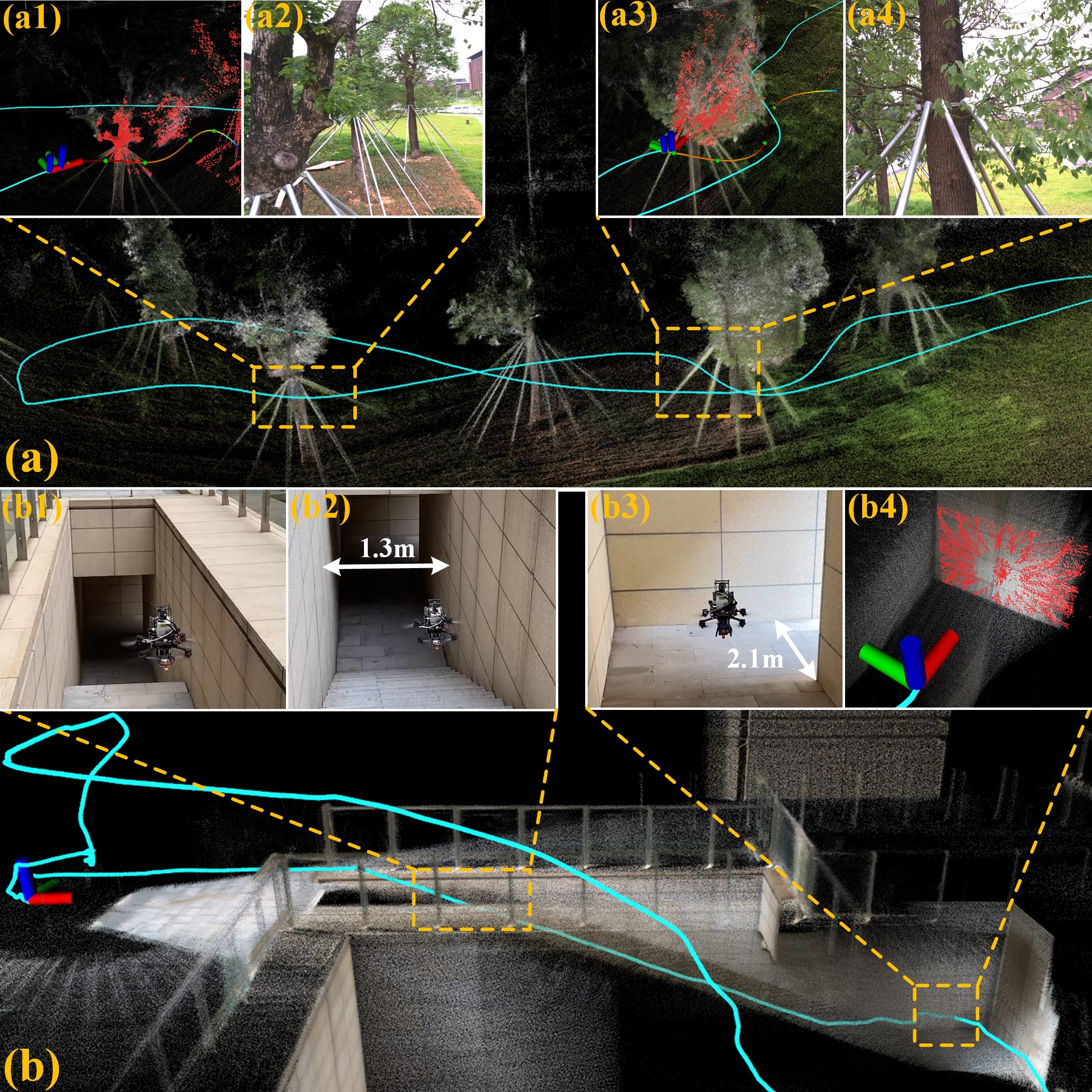}
              \vspace{-0.5cm}
			\caption{(a) and (b) are the enlarged point maps of the ``Woods" and ``Narrow Opening" experiments, respectively. The red points in (a1), (a3), and (b4) represent the current scan. (a2) and (a4) represent the first-person view at the corresponding locations. (b1), (b2), and (b3) depict the third-person view (see more details on YouTube: \href{https://youtu.be/aSAwVqR22mo}{\tt{youtu.be/aSAwVqR22mo}}).}	
			\label{fig:uav_nav}
           \vspace{-0.35cm}
		\end{figure}
  \vspace{-0.85cm}
		\subsection{Fully Onboard Autonomous UAV Navigation}\label{subsec:uav}
		Given the high precision and robust localization performance of FAST-LIVO2, along with its real-time capabilities, we conduct closed-loop autonomous UAV flights. 
	\subsubsection{System Configurations}\label{subsubsec:uavconfig}
		The hardware and software setup are illustrated in Fig. \ref{fig:uav}. For hardware, we {use} a NUC (Intel i7-1360P CPU and 32 GB RAM) as the onboard computer. In terms of software, the localization component is powered by FAST-LIVO2, which provides position feedback at 10Hz. The localizationr result is fed to the flight controller to achieve 200Hz feedback on position, velocity, and attitude. Besides localization, {FAST-LIVO2 supplies a dense registered point cloud to the planning module, the Bubble planner \cite{ren2022bubble}, which plans a smooth trajectory that is then tracked by an on-manifold Model Predictive Control (MPC) \cite{lu2022manifold}.} The MPC calculates the desired angular rates and thrust, which are tracked by respective low-level angular rate controllers running on the flight controller. Importantly, the MPC, Planner, and FAST-LIVO2 all operate on the onboard computer in real time.
		\subsubsection{UAV Autonomous Navigation}
		We conduct 4 fully onboard autonomous UAV navigation experiments, ``Basement", ``Woods", ``Narrow Opening", and  ``SYSU Campus" (Table~\ref{tab:uav_dataset_setup} in the Supplementary Material\cite{zheng2024supplementary}). ``Basement" and ``Woods" experiments are fully autonomous flights incorporating all planning, MPC, and FAST-LIVO2 modules, while ``Narrow Opening” and ``SYSU Campus" are manual flights with only MPC and FAST-LIVO2 (without the planning component).        
        As can be seen, ``Basement" and ``Woods" showcase the UAV's successful autonomous navigation and obstacle avoidance. In  ``Narrow Opening", the UAV is commanded to fly close proximity to a wall leading to few LiDAR points measurements. Nevertheless, the raycasting module recalls a greater number of visual map points, providing abundant constraints for localization, which allows for stable localization. Moreover,
        ``Basement" and ``Narrow Opening" experience LiDAR degeneration, observing only a single wall (see Fig. \ref{fig_cover} (e1) and (e4), Fig. \ref{fig:uav_nav} (b1-b4)), along with significant exposure variations (see Fig. \ref{fig_cover} (e5-e6)). Despite these challenges, our UAV system performed exceptionally well. ``Woods" involves the UAV moving at high speeds up to 3 m/s, demanding rapid response from the entire UAV system (see Fig. \ref{fig:uav_nav} (a1-a4)). ``SYSU Campus", a non-degenerated scene, primarily demonstrates the onboard high-precision mapping capabilities (see Fig. \ref{fig:sysu} in the Supplementary Material\cite{zheng2024supplementary}). {Finally, it is worth mentioning that in all these four UAV flights, severe lighting variation occurred. FAST-LIVO2 is able to estimate exposure time that closely follows the ground-truth values (see Fig. \ref{fig:test_expo2} in the Supplementary Material\cite{zheng2024supplementary}).}
		
\begin{figure}[htp]
        \vspace{-0.2cm}
			\centering
			\includegraphics[width=1.0\columnwidth]{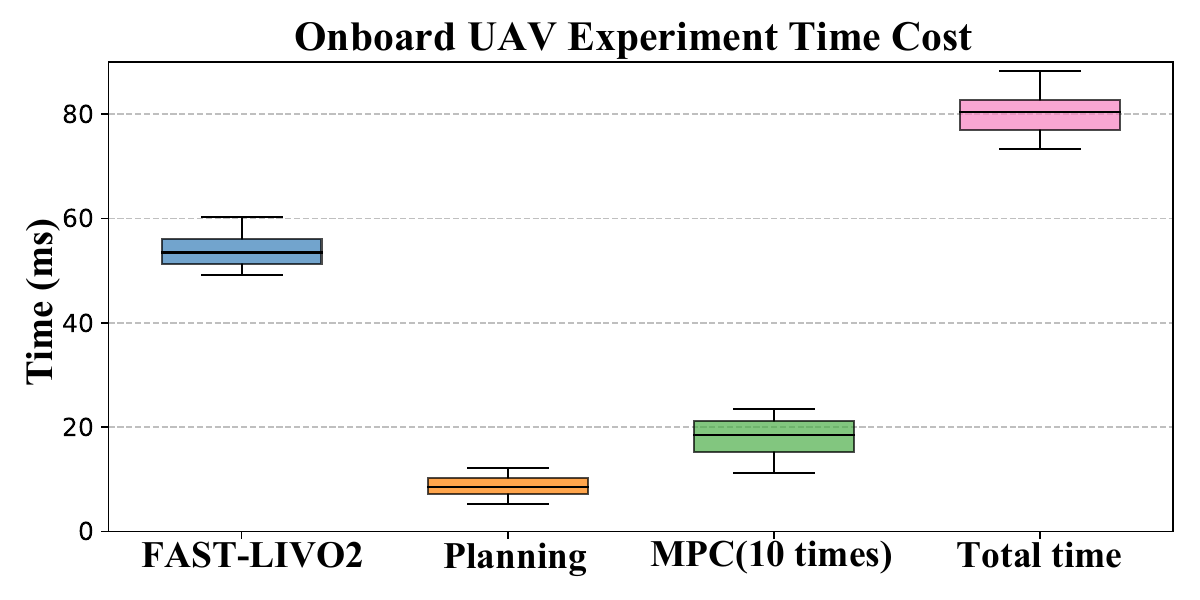}
            \vspace{-0.6cm}
			\caption{{The processing time of each module and in total in UAV autonomous navigation experiments across ``Basement", ``Woods", ``Narrow Opening", and ``SYSU Campus".} The MPC executes at 100 Hz while planning and FAST-LIVO2 execute at 10 Hz, so its computation time is counted for 10 times.}
			\label{fig:uav_time}
                \vspace{-0.5cm}
	\end{figure}

		Regarding onboard computational time, the need to run MPC (at 100 Hz) and Planning (at 10 Hz) on the onboard computer consumes computational resources and memory, limiting the computation resources available to FAST-LIVO2. Despite of the concurrent execution of control and planning, {as illustrated in Fig. \ref{fig:uav_time}, the average onboard processing time per LiDAR scan and image frame for FAST-LIVO2, approximately 53.47 ms, is still well below the frame period 100 ms. The average processing times for planning and MPC are 8.43 ms and 18.5 ms, respectively. The total average processing time of 80.4 ms meets very well the real-time requirements for onboard operations.}
  	\begin{figure}[htp]
			\centering
			\includegraphics[width=1.0\columnwidth]{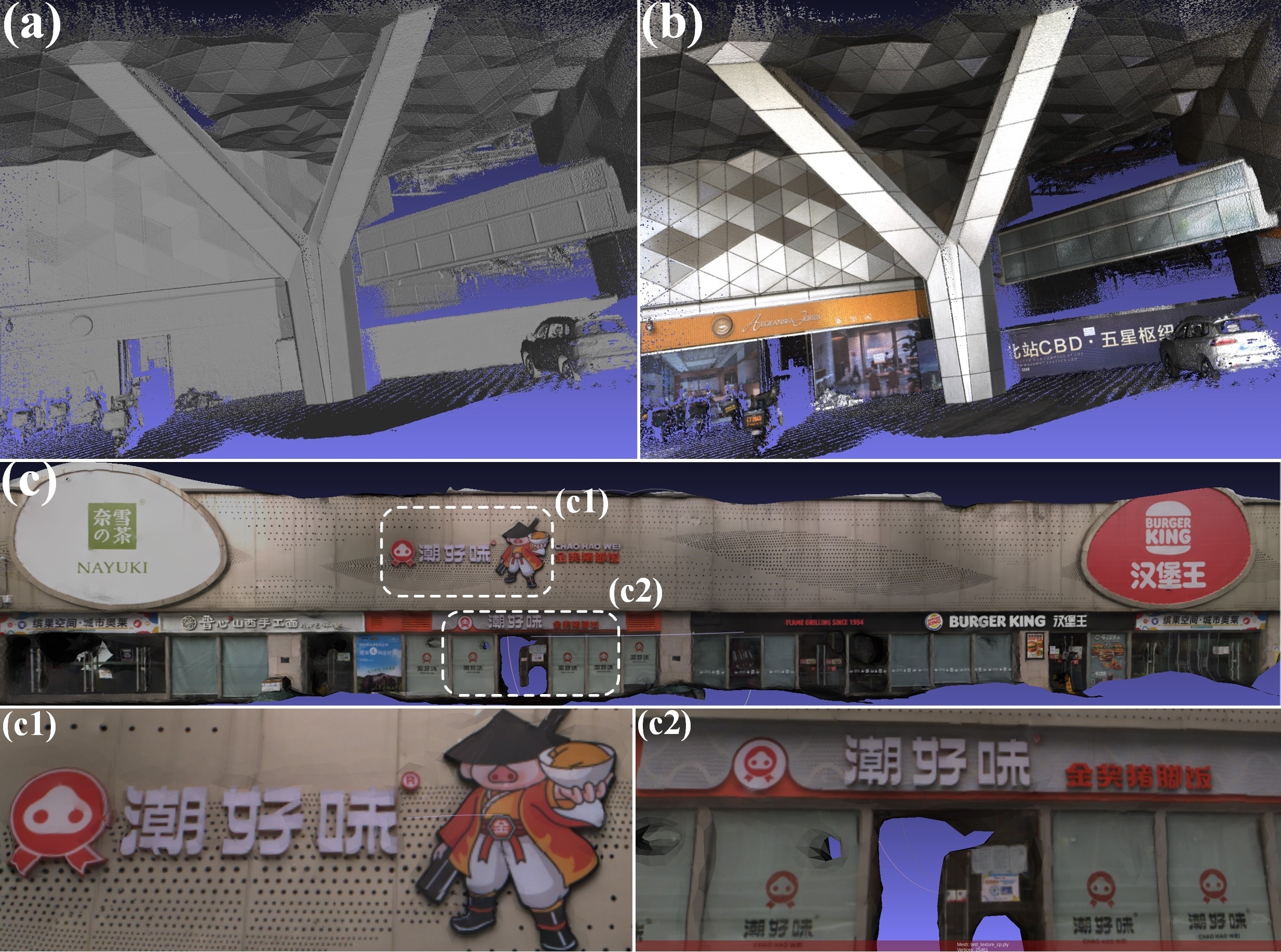}
                \vspace{-0.6cm}
			\caption{(a) and (b) are the mesh and texture mapping of ``CBD Building 01", respectively. (c) is the texture mapping of ``Retail Street", with (c1) and (c2) showing local details.}
			\label{fig:mesh}
            \vspace{-0.4cm}
        \end{figure}
		\subsection{Airborne Mapping}
		Airborne mapping represents a crucial task in surveying and mapping applications. To evaluate the suitability of FAST-LIVO2 for this application, {we conduct an aerial mapping experiment using the public dataset MARS-LVIG \cite{li2024mars} whose hardware configuration is detailed in Section \ref{subsec:public}. We evaluate the two sequences ``HKairport01" and ``HKisland01", whose real-time mapping results are illustrated in Fig. \ref{fig_cover} (a-c), with (a) and (c) corresponding to ``HKisland01", and (b) depicting ``HKairport01".} The results demonstrate the effectiveness of FAST-LIVO2 in unstructured environments such as forests and islands. The system successfully captures many fine structures and sharp coloring effects, including buildings, lane marks on roads, road curbs, tree crowns, and rocks, all of which are clearly visible.
	The APE (RMSE) for these sequences are 0.64 m and 0.27 m for FAST-LIVO2,   respectively, compared to 2.76 m and 0.52 m for R3LIVE. The average processing times on the desktop PC (Section \ref{subsec:config}), are approximately 25.2 ms and 21.8 ms, respectively, compared to 110.5 ms and 100.2 ms for R3LIVE.
	\begin{figure*}[htp]
	\centering
	\includegraphics[width=2.0\columnwidth]{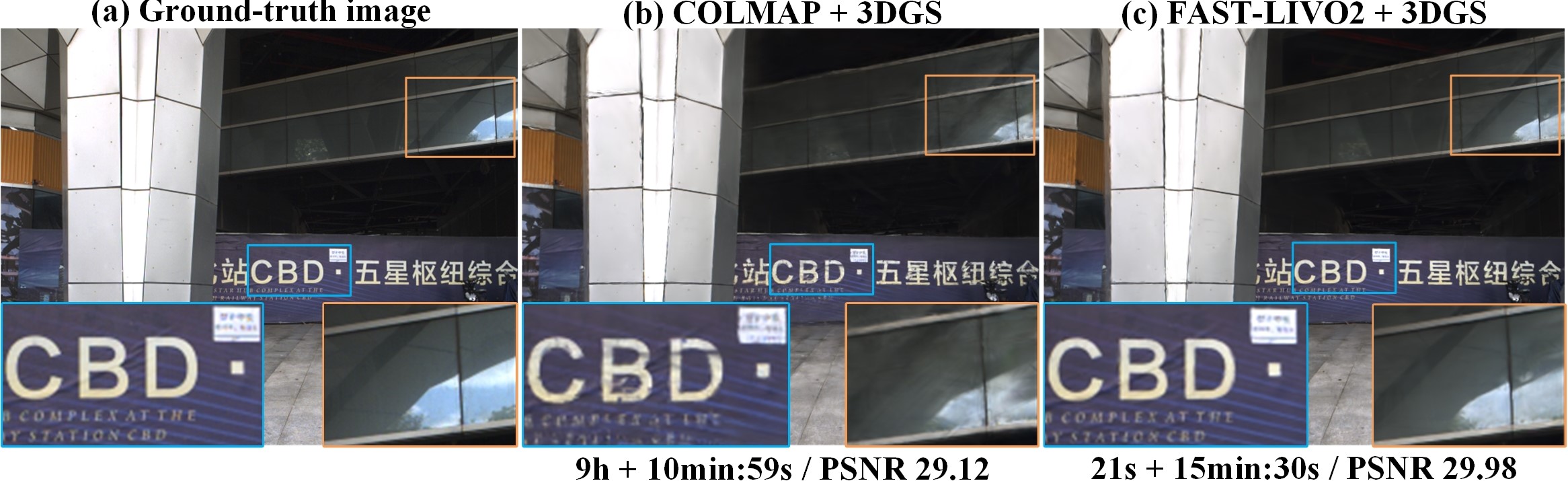}
        \vspace{-0.2cm}
	\caption{Comparison of ground-truth image, COLMAP+3DGS, and FAST-LIVO2+3DGS in terms of render details, {computational time (time for generating point clouds and estimating poses + training time)}, and PSNR for a random frame in ``CBD Building 01" (see more details on YouTube: \href{https://youtu.be/aSAwVqR22mo}{\tt{youtu.be/aSAwVqR22mo}}).}
	\label{fig:nerf}
        \vspace{-0.3cm}
        \end{figure*}
		\subsection{Supporting 3D Scene Applications: Mesh Generation, Texture, and Gaussian Splatting}
			
            Leveraging the high-precision sensor localization and dense 3D colored point map obtained from FAST-LIVO2, we develop software applications for rendering pipelines including meshing and texturing, as well as emerging NeRF-like rendering pipeline such as 3D Gaussian Splatting (3DGS). {For meshing, we employ VDBFusion \cite{vizzo2022sensors} based on the Truncated Signed Distance Function (TSDF) in ``CBD Building 01", shown in Fig. \ref{fig:mesh} (a). The sharp edges on the columns and the distinct structure of the roof are clearly visible, demonstrating the high quality of the mesh. This level of detail is achieved due to the high density of FAST-LIVO2's point clouds and the exceptional accuracy of structural reconstruction. After mesh construction, we {use} OpenMVS \cite{cernea2020openmvs} to perform texture mapping using the estimated camera poses in ``CBD Building 01" and ``Retail Street", shown in Fig. \ref{fig:mesh} (b-c). In Fig. \ref{fig:mesh} (c1-c2), the texture images applied on the triangular facets are seamless and accurately aligned, resulting in a highly clear and precise texture mapping. This is attributed to pixel-level image alignment achieved by FAST-LIVO2.}
   
            {The dense color point clouds from FAST-LIVO2 can also directly serve as the input of 3DGS.} We conduct tests on the sequence ``CBD Building 01" utilizing 300 frames out of a total of 1,180 images. The results are shown in Fig. \ref{fig:nerf}. Compared to COLMAP\cite{7780814}, our method significantly reduces the time required to obtain dense point clouds and poses from 9 hours to 21 s. However, the training time increases from 10 min and 59 s to 15 min and 30 s. This increase is attributed to the denser point clouds (downsampled to 5 cm), which introduce more parameters to optimize. Nonetheless, the increased density and precision of our point clouds result in a slightly higher Peak Signal-to-Noise Ratio (PSNR) compared to the PSNR obtained from COLMAP inputs.
		\section{Conclusion and Future Work}\label{sec:conclusion}
		{This paper proposed FAST-LIVO2, a direct LIVO framework achieving fast, accurate, and robust state estimation while reconstructing the map on the fly. FAST-LIVO2 can achieve
        high localization accuracy while being robust to severe LiDAR and/or visual degeneration.}
        
         {The gain in speed is attributed to the use of raw LiDAR, inertial, and camera measurements within an efficient ESIKF framework with sequential update. In the image update, an inverse compositional formulation along with a sparse patch-based image alignment is further adopted to boost the efficiency.
         The gain in accuracy is attributed to the use (and even refine) of plane priors from LiDAR points to enhance accuracy of image alignment. Besides, a single unified voxel map is {used} to manage simultaneously the map points and the observed high-resolution image measurements. The voxel map structure, which supports geometry construction and update, visual map point generation and update, and reference patch update, is developed and validated. 
         The gain in robustness is due to {real-time} estimation of exposure time, which effectively handles environment illumination variation, and on-demand voxel raycasting to cope with LiDARs' close proximity blind zones.} 
         {The efficiency and accuracy of FAST-LIVO2 were evaluated on extensive public datasets, while the robustness and effectiveness of each system module were evaluated on private dataset. The applications of FAST-LIVO2 in real-world robotics applications, such as UAV navigation, 3D mapping, and model rendering, were also demonstrated.}
    
        
	   {As an odometry, FAST-LIVO2 may have drifts over long distances. In the future, we could integrate loop closure and the sliding window optimization into FAST-LIVO2 to mitigate this long-term drift. Moreover, the accurate and dense colored point maps could be {used} to extract semantic information for object-level semantic mapping.}
		\bibliography{paper}

\clearpage
\setcounter{equation}{0}
\setcounter{figure}{0}
\setcounter{table}{0}
\setcounter{page}{1}
\setcounter{section}{0}%
\setcounter{subsection}{0}%
\setcounter{subsubsection}{0}%
\setcounter{paragraph}{0}%
\onecolumn

\begin{center}
	\textbf{\large Supplementary Material for FAST-LIVO2: Fast, Direct LiDAR-Inertial-Visual Odometry}
\end{center}
\renewcommand{\theequation}{S\arabic{equation}}
\renewcommand{\thefigure}{S\arabic{figure}}
\renewcommand{\thetable}{S\arabic{table}}
\section*{I. System Module Validation}\label{subsec:module_valid}
In this section, we validate the key modules of our system, including the affine warping, normal refinement, {reference patch update}, on-demanding raycasting query, exposure time estimation, {and ESIKF sequential update utilizing the FAST-LIVO2 private dataset and MARS-LVIG dataset.}
\subsection*{A. Evaluation of Affine Warping}\label{subsubsec:affine_warp}
\begin{figure}[htp]
\begin{center}
{\includegraphics[width=1\columnwidth]{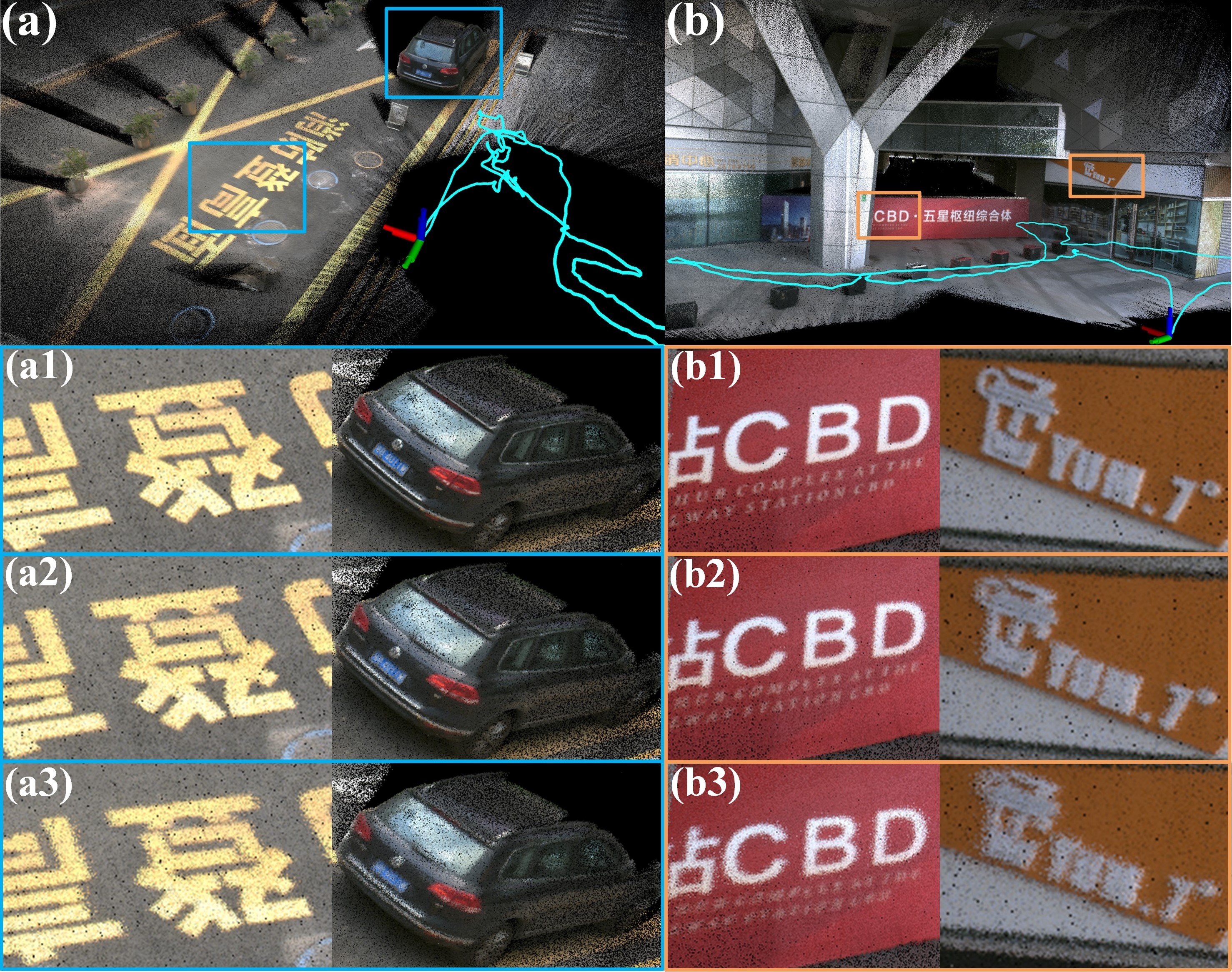}}
\end{center}
\vspace{-0.3cm}
\caption{\label{fig:WarpCloud}{(a) and (b) are FAST-LIVO2 default mapping results in ``CBD Building 02" and ``Office Building Wall", respectively. (a1, b1), (a2, b2), (a3, b3) are enlarged views of the point clouds using ``Plane normal refined'', ``Plane prior'', and ``Constant depth'', respectively.}}
\end{figure}

{In this experiment, we aim to comprehensively evaluate the various affine warping effects based on the {constant depth assumption} (a common technique utilized in semi-dense methods), the plane prior from point clouds, and the refined plane normal of our proposed system (denoted as {``Constant depth"}, ``Plane prior", and ``Plane normal refined“, respectively). To achieve this, we compare the mapping results and drift metrics of the three methods on ``CBD Building 02" and ``Office Building Wall".}
 As depicted in Fig. \ref{fig:WarpCloud}, ``Plane normal refined'' delivers the most clear and accurate mapping results and the next best is ``Plane prior''. Notably, ``Plane normal refined'' renders text and patterns on the ground and walls, as well as lane markings, with remarkable clarity. Moreover, the drift associated with ``Plane prior'' and ``Plane normal refined'' remains below \SI{0.01}{\meter}, whereas ``Constant depth'' does not return to the starting position, experiencing a drift of \SI{0.22}{\meter}. Such results confirm the enhanced performance of affine warping based on the plane prior and enhancements by the plane normal refinement.
 
\begin{figure}[htp]
    \begin{center}
    \includegraphics[width=1.0\columnwidth]{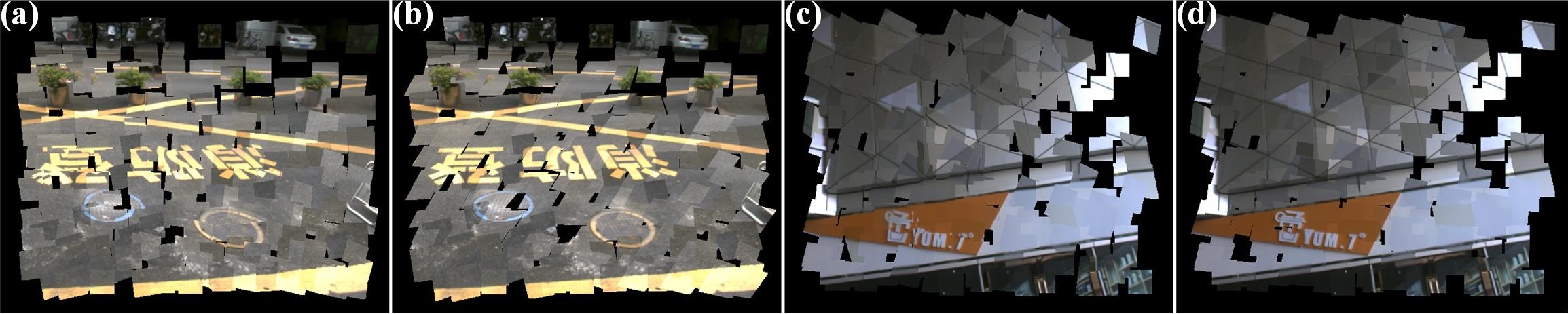}
    \end{center}
    \vspace{-0.4cm}
        \caption{\label{fig:WarpProjection} (a) and (c) depict the warped patches derived from the {constant depth assumption}, while (b) and (d) represent those based on the plane prior. }
    \vspace{-0.4cm}
\end{figure}

Besides, we compare warped projection effects based on ``Constant depth'' and ``Plane prior'' on the sequence ``CBD Building 02" and ``Office Building Wall". We randomly select several image frames from these two sequences for the qualitative analysis. For each frame, we project the reference patches attached to the visual map points visible in the frame onto a blank image of the current frame. This process yields a novel RGB image. If the affine warping and pose estimation are both performed well, areas with patch projections will produce a seamless and minimally distorted appearance, closely resembling the raw RGB image. The comparison of warped patches is presented in Fig. \ref{fig:WarpProjection}. The results indicate that the pose accuracy and warped performance under ``Plane prior'' significantly outperform those under ``Constant depth''
. 
\begin{figure}[htp]
    \begin{center}
        {\includegraphics[width=1.0\columnwidth]{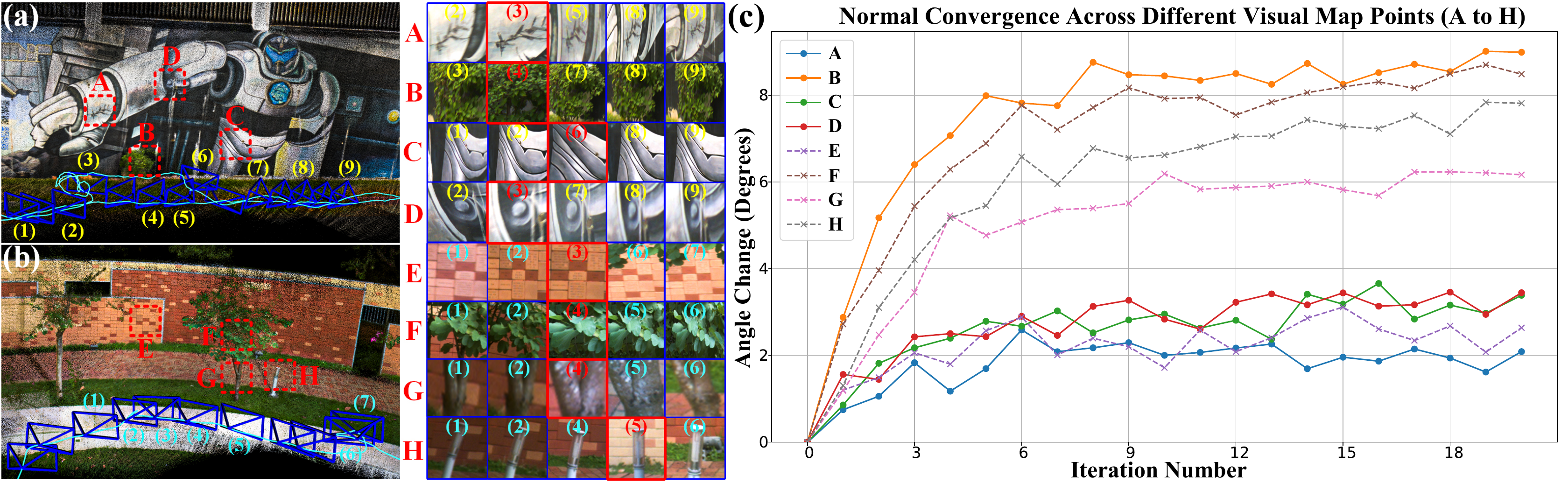}}
    \end{center}
    \vspace{-0.2cm}
    \caption{\label{fig:reference_update}The illustration of reference patch update. (a) and (b) are the reconstructed point clouds of sequences ``HIT Graffiti Wall" and ``HKU Centennial Garden", respectively. Regions A to H encompass both structured and unstructured areas in the scene. On the right, the $40\times40$ image patches illustrate various observations for the same region, with numbers indicating the corresponding camera frame on the left. The reference patch for each region is highlighted by a red box. (c) shows the convergence of patch normal across regions A to H.}
\end{figure}
\begin{figure}[htp]
        \begin{center}
        {\includegraphics[width=1.0\columnwidth]{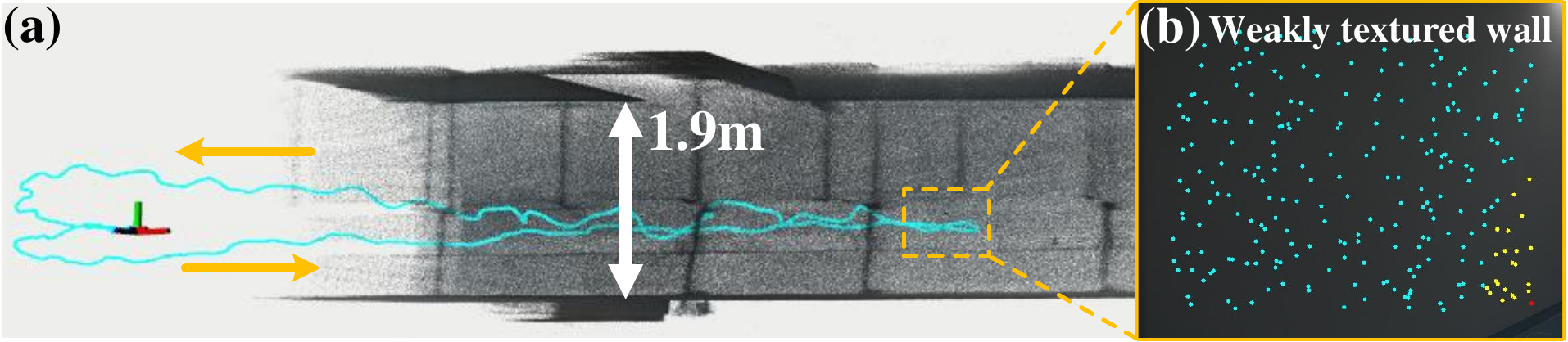}}
        \end{center}      
        \vspace{-0.2cm}
        \caption{\label{fig:raycasting_exp}The illustration of on-demand raycasting and voxel query. (a) displays a top-down view of the reconstructed point cloud for ``Narrow Corridor", with a corridor width of \SI{1.9}{\meter}. In (a), the blue line indicates the trajectory and the yellow arrows represent the direction of movement. The LiDAR camera sensor suite is turned to face a wall on the left side at the location contained in the dashed box. (b) shows the camera image during the turn, having dim lighting conditions and very few LiDAR points. In (b), blue dots represent points acquired from raycasting, yellow dots indicate points from voxel query, and red dots are points in the LiDAR scan.}
        \vspace{-0.2cm}
\end{figure}
        \begin{figure}[t]
            \begin{center}
                {\includegraphics[width=1.0\columnwidth]{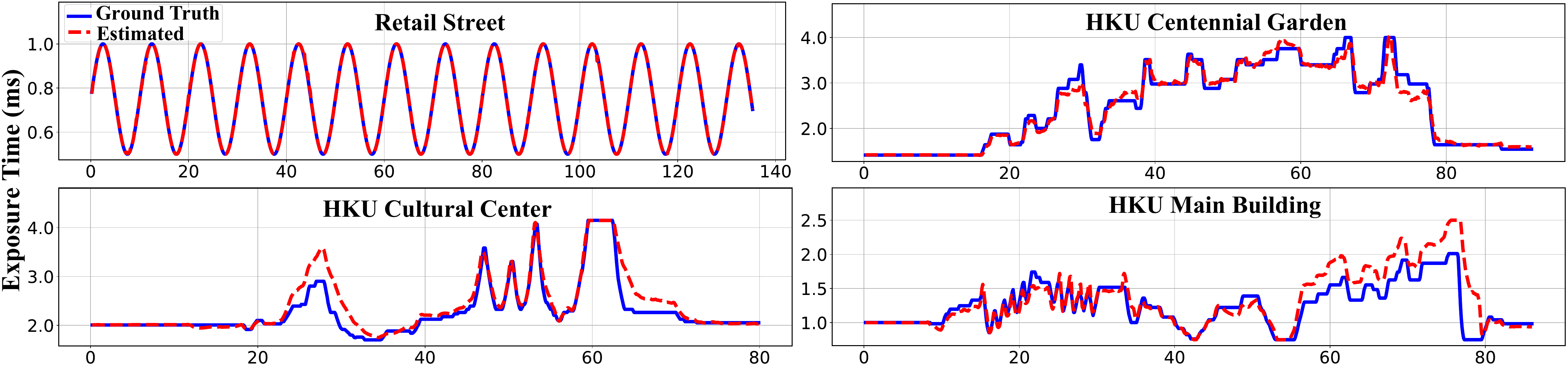}}
            \end{center}
            \vspace{-0.2cm}
            \caption{\label{fig:test_expo}Comparison of estimated exposure time versus ground truth exposure time {in ``Retail Street", ``HKU Centennial Garden", ``HKU Cultural Center", and ``HKU Main Building".}}
        \vspace{-0.2cm}
        \end{figure}
        \subsection*{B. Evaluation of Reference Patch Update and Normal Convergence}\label{subsubsec:update_refine} {In this experiment, we validate the effects of the reference patch update strategy and normal convergence on ``HIT Graffiti Wall" and ``HKU Centennial Garden". As shown in Fig. \ref{fig:reference_update}, (a) and (b) are the reconstructed point clouds of these two sequences. On the right, for each region from A to H, we respectively present five patch observations captured at different poses, each with a patch size of $40\times40$ pixels for visualization. These patches are observed in camera frames located at corresponding numbers on the left. As can be seen, our reference path update strategy tends to choose a high-resolution reference patch that faces the plane along its normal. It is also noticed that, despite of visual map points and patches generated at non-planar locations (e.g., tree leaves, trunk, and a lamp stand), the overall mapping quality is still high. }
      
        We also evaluate the convergence of our proposed normal estimation across patches in regions A to H. Each patch is in the size of $11\times 11$ pixels. The initial normal vector is estimated from the LiDAR points. The convergence curves, which represent the angle change between the initial and optimized normal vectors at each iteration number, are shown in Fig. \ref{fig:reference_update}. Regions A, C, D, and E are structured areas, whereas regions B, F, G, and H are unstructured ones. It can be observed that normal vectors for structured areas converge faster (within 6 iterations) with small normal refinement (2 to 4 degrees) because the initial normal provided by point clouds is relatively accurate. In unstructured areas, such as shrubs and tree leaves (i.e., B and F), the normal refinement is significant (up to 9 degrees) and requires 9 iterations to converge. Overall, normal refinement across these 8 regions demonstrates good convergence properties.
	\subsection*{C. Evaluation of On-demand Raycasting}\label{subsubsec:raycast}
		In this experiment, we assess the performance of the on-demand raycasting module under extreme conditions where the current and recent LiDAR scans have few or even no points due to LiDARs' close proximity blind zone\footnote{\href{https://www.livoxtech.com/avia/specs}{https://www.livoxtech.com/avia/specs}}. 
        We use the sequence ``Narrow Corridor" for an in-depth analysis illustrated in Fig. \ref{fig:raycasting_exp}. In this sequence, we traverse an extremely narrow tunnel, approximately \SI{1.9}{\meter} in width, and turn to face the weakly textured wall on one side. {Due to the limited points in the LiDAR scans when facing the wall, we can only acquire few visual map points through voxel query (yellow dots in Fig. \ref{fig:raycasting_exp} (b)). In this case, raycasting offers sufficient visual constraints to mitigate degeneration (blue dots in Fig. \ref{fig:raycasting_exp} (b)).}  
        The visualization results demonstrate that the on-demand raycasting module works well under challenging conditions with few points in LiDAR scans.
	\subsection*{D. Evaluation of Exposure Time Estimation}\label{subsubsec:exposure}
		In this experiment, we validate the exposure time estimation module in two parts:
		1) For the sequence with fixed exposure and gain, we multiply each pixel of the received raw image by an exposure factor that changes sinusoidally over time. We verify the effectiveness of our estimation by comparing the estimated exposure times against the sinusoidal function applied.		
		2) For sequences with auto-exposure and either fixed or auto-gain settings, we evaluate the accuracy of our estimated exposure times by comparing them with the ground truth values retrieved from the camera's API.
		
	In part one, we test on the sequence ``Retail Street", applying an exposure factor to images with fixed exposure and gain. As shown in Fig. \ref{fig:test_expo}, the estimated relative inverse exposure time matches the true values very well, evidencing the convergence of our exposure estimation in synthetic conditions. 
    In part two, we use the sequences ``HKU Centennial Garden", ``HKU Cultural Garden" and ``HKU Main Building", which have significant exposure time changes, for testing.
    We scale the estimated relative exposure times by the first frame to recover the actual exposure time (ms) of each frame. As shown in Fig. \ref{fig:test_expo}, the estimated exposure time follows closely the ground-truth values, which validates the effectiveness of our exposure time estimation module. The occasional mismatches are possibly due to the unmodeled response function and vignetting factor \cite{engel2016photometrically}. 
\subsection*{E. Evaluation of ESIKF Sequential Update}\label{subsubsec:sequential}
In this experiment, we evaluate different ESIKF update strategies for LiDAR and camera states. We compare asynchronous versus synchronous updates, as well as standard versus sequential updates. Specifically, we assess three strategies: ``asynchronous (standard update)'', where the camera and LiDAR states are updated at their respective sampling times without scan recombination; ``synchronous (standard update)'', where LiDAR scans are recombined to sync with camera images and the state is updated with both LiDAR and camera measurements within a standard ESIKF; and ``synchronous (sequential update)'', where the LiDAR and camera are synced but the state is first updated by LiDAR measurements and then updated by camera measurements. These strategies are evaluated in terms of accuracy, robustness, and efficiency using the ``AMvalley03'' sequence of the MARS-LVIG dataset. We select this sequence for several key reasons:
\begin{itemize}
\item[(1)] This sequence includes slopes that lead to both LiDAR and visual degenerations, making it a challenging test case.
\item[(2)] This sequence represents an extremely large-scale scenario (approximately \SI{901}{\meter}$\times$\SI{500}{\meter}$\times$\SI{130}{\meter}) with long-term and
high-speed data collection (covering \SI{600}{\second} at a speed of \SI{12}{\meter/\second}), where pose deviations are prone to occur (due to the
long-term and high-speed conditions), and even slight drifts can cause significant blurring in the colored point clouds
(due to the large scale), leading to more pronounced comparative results.
\item[(3)] This sequence provides the RTK ground truth data, allowing for more accurate quantitative comparisons.
\end{itemize}

We compare the qualitative mapping results, quantitative APE, and the average processing time of the three update strategies. The experimental configuration is as follows: LiDAR updates involve up to 5 iterations, visual updates use a three-level pyramid with up to 5 iterations per level and no more than 3 iterations per level when the camera and LiDAR are updated simultaneously in a standard ESIKF, and the scale normalization factor (from visual photometric error to LiDAR point-to-plane distance) for the ``synchronous (standard update)'' is set to 0.0032, which has been meticulously tuned for optimal performance.

\begin{figure}[htp]
        \begin{center}
            {\includegraphics[width=1.0\columnwidth]{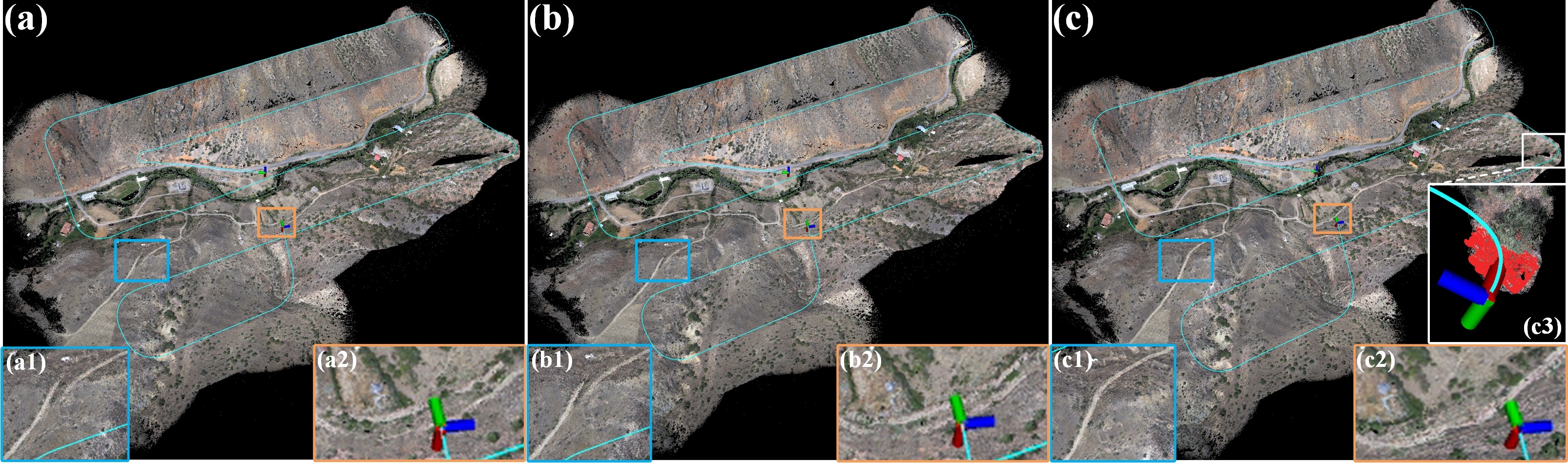}}
        \end{center}
        \vspace{-0.2cm}
    \caption{\label{fig:test_sequential}{(a), (b), and (c) are the mapping results in ``AMvalley03'' for the ``asynchronous (standard update)'', ``synchronous (standard update)'', and ``synchronous (sequential update)'' strategies, respectively. (a1, b1, c1) and (a2, b2, c2) are enlarged views of the point clouds for different update strategies. The blue line represents the UAV's flight path, and the red points in (c3) indicate the LiDAR scan at that moment.}}
    \vspace{-0.3cm}
    \end{figure}

Fig. \ref{fig:test_sequential}, (a-c) present the reconstructed colored point clouds for this sequence. It is evident that the ``synchronous (sequential update)'' strategy produces accurate mapping results, particularly in the areas highlighted by the blue and orange boxes, where the mountain roads are reconstructed without any layering. In contrast, the other two strategies exhibit misalignments in these areas, although the ``synchronous (standard update)'' performs slightly better than the ``asynchronous (standard update)''.
The superior performance of the ``synchronous (sequential update)'' strategy is mainly attributed to its robustness in handling significant LiDAR and visual degenerations, as seen in the white box (c3). This area features a large, textureless slope, and the UAV passes over it at a high speed, heavily relying on a strong prior. The other two methods, which rely solely on the IMU prior, struggle to compute a relatively accurate image gradient descent direction, leading to significant linearization errors.

The APE (RMSE) metrics for the ``AMvalley03'' sequence are \SI{3.12}{\meter}, \SI{2.45}{\meter}, and \SI{0.68}{\meter} for the ``asynchronous (standard update)'', ``synchronous (standard update)'', and ``synchronous (sequential update)'', respectively. The average processing times on a desktop PC (Section IX-A) are approximately \SI{27.6}{\milli\second}, \SI{49.9}{\milli\second}, and \SI{23.1}{\milli\second}. 
Our proposed ``synchronous (sequential update)'' achieves the highest efficiency and accuracy, while the ``asynchronous (standard update)'' has the lowest accuracy. The ``synchronous (standard update)'' is the most time-consuming, primarily because it requires fusing all LiDAR measurements at each level of the image pyramid.

Overall, our proposed ``synchronous (sequential update)'' offers superior accuracy and efficiency, while the ``asynchronous (standard update)'' has the lowest accuracy, and the ``synchronous (standard update)'' is the most time-consuming.
\clearpage
\section*{II. Additional Information}\label{sec:appendix}
\begin{table*}[htp]
    \caption{Overview of FAST-LIVO2 Private Dataset}
    \vspace{-0.2cm}
    \centering
    \begin{threeparttable}[b]
        \renewcommand\arraystretch{1.2}
        \setlength{\tabcolsep}{3.5pt}
        \begin{tabular}{lccccccc}
            \toprule
            Sequence & \makecell{Duration \\ (minute:second)} & \makecell{Sensor \\ Degeneration} & \makecell{Return \\  Origin\tnote{1}} & \makecell{Exposure / \\ Gain Mode} & \makecell{Exposure \\ Time\tnote{2}} & \makecell{Close Surface \\ Distance\tnote{3}} & \makecell{Scene \\ Characteristics} \\ [1.5 ex]
            \hline
            Retail Street & 2 min : 15 sec & --- & \checkmark& Fixed / Fixed &  &  & Outdoor\\
            CBD Building 01 & 1 min : 58 sec & --- & \checkmark& Fixed / Auto &  &  & Outdoor\\
            CBD Building 02 & 3 min : 54 sec & LiDAR & \checkmark & Auto / Fixed & \checkmark &  & Outdoor\\
            CBD Building 03 & 5 min : 01 sec & Camera, LiDAR & \checkmark & Auto / Fixed & \checkmark &  & Outdoor, Textureless\\
            HKU Landmark & 1 min : 31 sec & --- & \checkmark& Auto / Fixed & \checkmark &  & Outdoor\\
            HKU Lecture Center & 1 min : 17 sec & LiDAR & \checkmark& Auto / Fixed & \checkmark &  & Indoor\\
            HKU Centennial Garden & 1 min : 32 sec & LiDAR & \checkmark& Auto / Fixed & \checkmark &  & Outdoor\\
            HKU Cultural Center & 4 min : 01 sec & LiDAR & \checkmark& Auto / Fixed & \checkmark &  & Outdoor\\
            HKU Main Building & 1 min : 36 sec & Camera, LiDAR & \checkmark& Auto / Auto & \checkmark & & Out(In)door\tnote{4}, Textureless\\
            HKUST Red Sculpture & 2 min : 10 sec & --- & \checkmark & Fixed / Auto  &  &  & Outdoor, Cluttered\\
            HIT Graffiti Wall & 8 min : 57 sec & LiDAR & \checkmark& Fixed / Auto &  &  & Outdoor\\
            Banner Wall  & 1 min : 40 sec & LiDAR & \checkmark& Fixed / Auto &  &  & Indoor, Dim\\
            Bright Screen Wall & 1 min : 18 sec & LiDAR & \checkmark& Fixed / Auto &  &  & Indoor\\
            Black Screen Wall & 0 min : 56 sec & Camera, LiDAR & \checkmark& Fixed / Auto &  &  & Indoor, Textureless\\ 
            Office Building Wall & 1 min : 9 sec & LiDAR & \checkmark& Fixed / Auto &  & \checkmark & Outdoor\\
            Narrow Corridor & 1 min : 1 sec & Camera, LiDAR & \checkmark& Fixed / Auto &  & \checkmark & Indoor, Dim, Textureless\\
            Long Corridor & 1 min : 35 sec & LiDAR & \checkmark& Fixed / Auto &  & \checkmark & Indoor, Dim\\
            Mining Tunnel & 15 min : 49 sec & Camera, LiDAR & \checkmark& Fixed / Auto &  &  & Indoor, Dim, Textureless\\
            SYSU 01 & 4 min : 40 sec &   ---  & \checkmark & Auto / Auto &  &  & Outdoor, Light\tnote{5}\\
            SYSU 02 & 4 min : 30 sec & Camera, LiDAR & \checkmark & Auto / Auto &  &  & Out(In)door, Light\\
            \hline
            \textbf{Total} & 66 min : 50 sec & Camera, LiDAR & \checkmark & Auto (Fixed) / Auto (Fixed) & \checkmark & \checkmark & \makecell{Out(In)door, Light, \\Dim, Textureless}\\
            \toprule
        \end{tabular}
        \begin{tablenotes}
            \item[1] Sequences are collected by traveling in a loop, starting and ending at the same position.
            \item[2] Sequences with ground truth camera exposure time read from camera’s API.
            \item[3] Sequences exist where LiDAR captures no/limited point clouds due to close proximity blind zones.
            \item[4] Sequences include the process of moving from indoor to outdoor environments.
            \item[5] Sequences characterized by significant lighting variations.
        \end{tablenotes}
    \end{threeparttable}
    \vspace{-0.4cm}
    \label{tab:dataset_setup}
\end{table*}
\begin{table*}[htp]
    \caption{Overview of UAV Autonomous Navigation Experiments}
    \vspace{-0.2cm}
    \centering
    \begin{threeparttable}[b]
        \renewcommand\arraystretch{1.2}
        \setlength{\tabcolsep}{5.9pt}
        \begin{tabular}{lcccccccc}
            \toprule
            Experiment & \makecell{Flight Duration \\ (minute:second)} & \makecell{Flight \\ Mode\tnote{6}} & \makecell{Sensor \\ Degeneration} & \makecell{Return \\  origin} & \makecell{Exposure / \\ Gain Mode} & \makecell{Exposure \\ Time} & \makecell{Close Proximity \\ to Obstacles} & \makecell{Scene \\ Characteristics} \\ [1.5 ex]
            \hline
            Basement  & 5 min & Autonomous & Camera, LiDAR & \checkmark& Auto / Auto & \checkmark  &  & Out(In)door, Light\\
		Narrow Opening  & 2 min:32 sec & Manual & Camera, LiDAR & \checkmark& Auto / Auto & 
            \checkmark & \checkmark & Out(In)door, Light\\
		SYSU Campus & 3 min:6 sec & Manual & --- & \checkmark& Auto / Auto &  &  & 
            Outdoor, Light\\
		Woods  & 2 min:49 sec & Autonomous & Camera & \checkmark& Auto / Fixed & \checkmark &  & 
            Outdoor, Cluttered\\
            \toprule
        \end{tabular}
        \begin{tablenotes}
            \item[6] {In autonomous mode, FAST-LIVO2, planning, and MPC modules are all running. In manual mode, FAST-LIVO2 and MPC are running while planning is disabled.}
        \end{tablenotes}
    \end{threeparttable}
    \vspace{-0.6cm}
    \label{tab:uav_dataset_setup}
\end{table*}

\begin{table*}[h!]
    \caption{{Average processing time per frame for FAST-LIO2 across different datasets}}
    \vspace{-0.2cm}
    \centering
    {
    \begin{threeparttable}[b]
        \renewcommand\arraystretch{0.93}
        \setlength{\tabcolsep}{7pt}
        \begin{tabular}{lccclc}
        \toprule
        Hilti'22 \& Hilti'23        & Processing Time (ms)  & NTU VIRAL & Processing Time (ms) & Private Dataset         & Processing Time (ms) \\ 
        \midrule
        Construction Ground      & 24.32               & eee\_01           & 17.63              & Retail Street             & 10.23         \\
        Construction Multilevel  & 23.21               & eee\_02           & 17.33              & CBD Building 01           & 10.40         \\
        Construction Stairs      & $\times$            & eee\_03           & 15.32              & CBD Building 02           & $\times$      \\
        Long Corridor            & 28.43               & nya\_01           & 19.99              & CBD Building 03           & $\times$      \\
        Cupola                   & $\times$            & nya\_02           & 18.62              & HKU Landmark              & 11.62         \\
        Lower Gallery            & 24.49               & nya\_03           & 19.32              & HKU Lecture Center        & $\times$      \\
        Attic to Upper Gallery   & $\times$            & nya\_04           & 15.71              & HKU Centennial Garden     & $\times$      \\
        Outside Building         & 18.67               & sbs\_02           & 16.10              & HKU Cultural Center       & $\times$      \\
        Floor 0                  & 27.34               & sbs\_03           & 17.71              & HKU Main Building         & $\times$      \\
        Floor 1                  & 27.12               &                   &                    & HKUST Red Sculpture       & 10.98         \\
        Floor 2                  & 25.22               &                   &                    & HIT Graffiti Wall         & $\times$      \\
        Basement                 & 22.31               &                   &                    & Banner Wall               & $\times$      \\
        Stairs                   & 20.91               &                   &                    & Bright Screen Wall        & $\times$      \\
        Parking 3x floors down   & $\times$            &                   &                    & Black Screen Wall         & $\times$      \\
        Large room               & 29.26               &                   &                    & Office Building Wall      & $\times$      \\
        Large room (dark)        & 28.20               &                   &                    & Narrow Corridor           & $\times$      \\
                                 &                     &                   &                    & Long Corridor             & $\times$      \\
                                 &                     &                   &                    & Mining Tunnel             & $\times$      \\
                                 &                     &                   &                    & SYSU 01                   & 11.21         \\
                                 &                     &                   &                    & SYSU 02                   & $\times$      \\
        \midrule
        Overall Average & 19.68 & & & & \\ 
        \bottomrule
        \end{tabular}
        \begin{tablenotes}
        \item[7] $\times$ denotes the system totally failed.
        \end{tablenotes}    
    \end{threeparttable}
\label{tab:fast_livo_times}}
\end{table*}

\begin{figure}[htp]
    \centering
    \includegraphics[width=1.0\columnwidth]{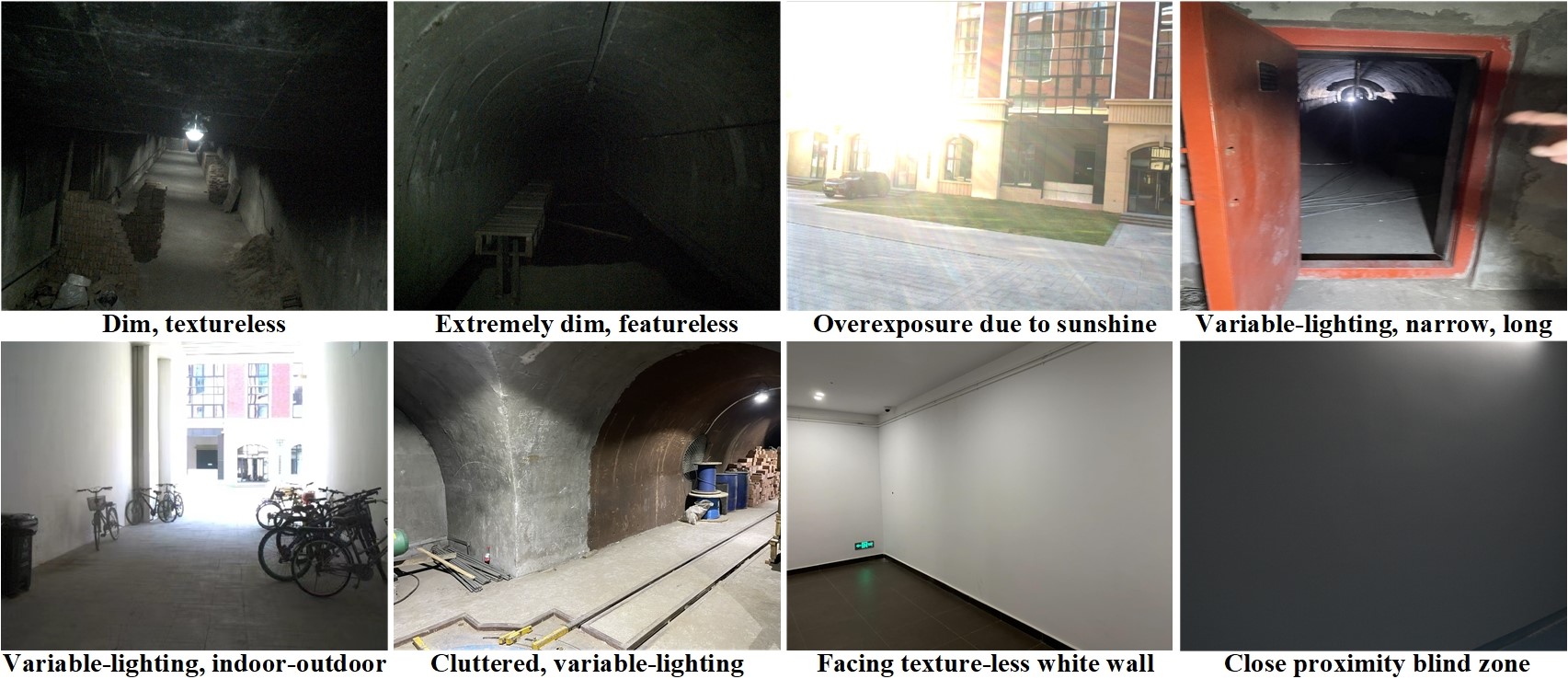}
    \caption{{Challenging environments captured in the FAST-LIVO2 private dataset.}}
    \label{sequence_discribe}
\end{figure}
\begin{figure}[htp]
    \begin{center}
    {\includegraphics[width=1.0\columnwidth]{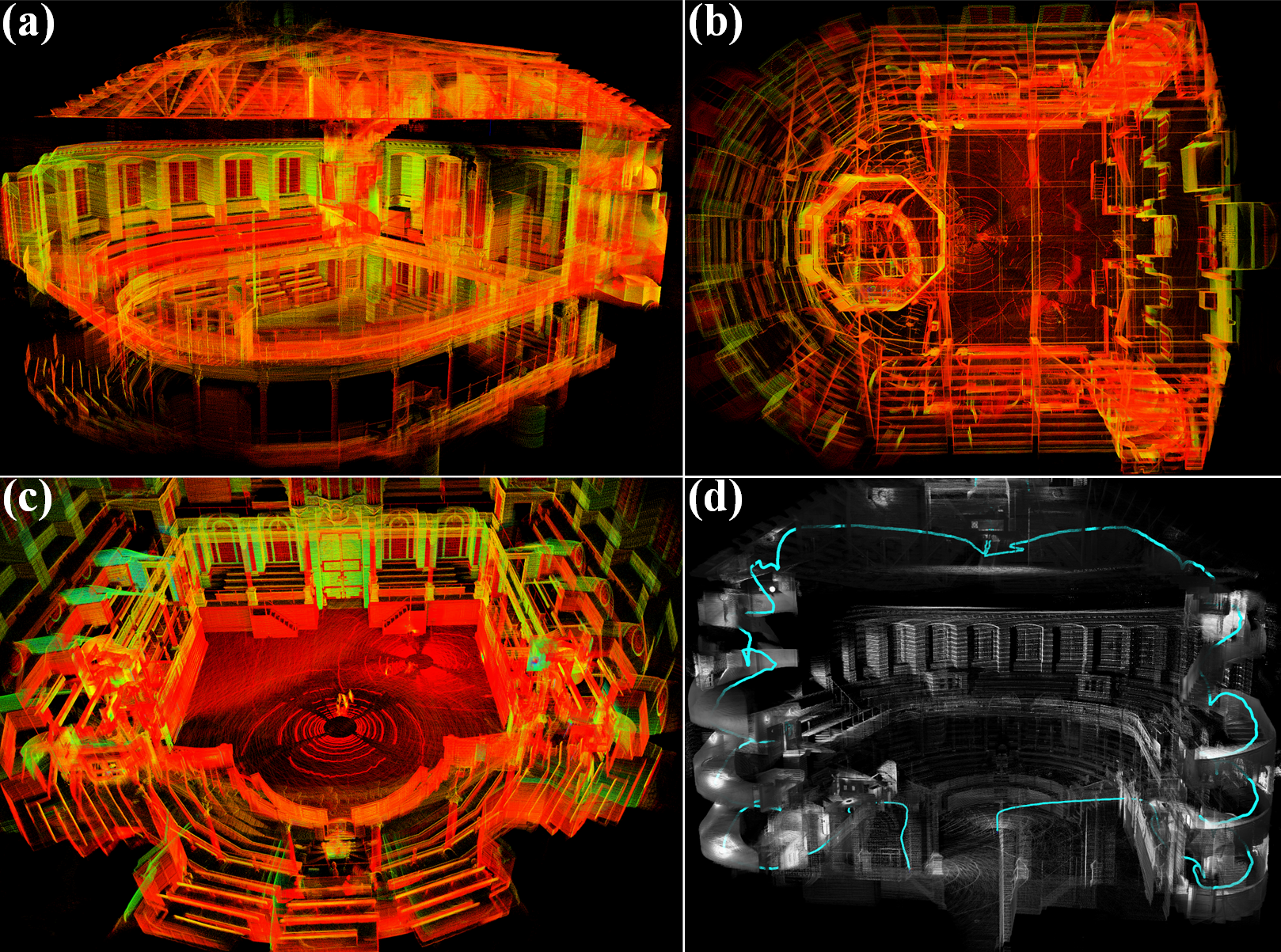}}
    \end{center}
    \caption{\label{fig:hilti}The real-time mapping results of FAST-LIVO2 in the Hilti'22 dataset. (a) ``Attic to Upper Gallery", (b) ``Cupola", (c) ``Lower Gallery", and (d) ``Construction Stairs". The point clouds in (a-c) are colored by intensity, while (d) is colored using grayscale images.}
\end{figure}
\begin{figure}[htp]
    \begin{center}
    {\includegraphics[width=1.0\columnwidth]{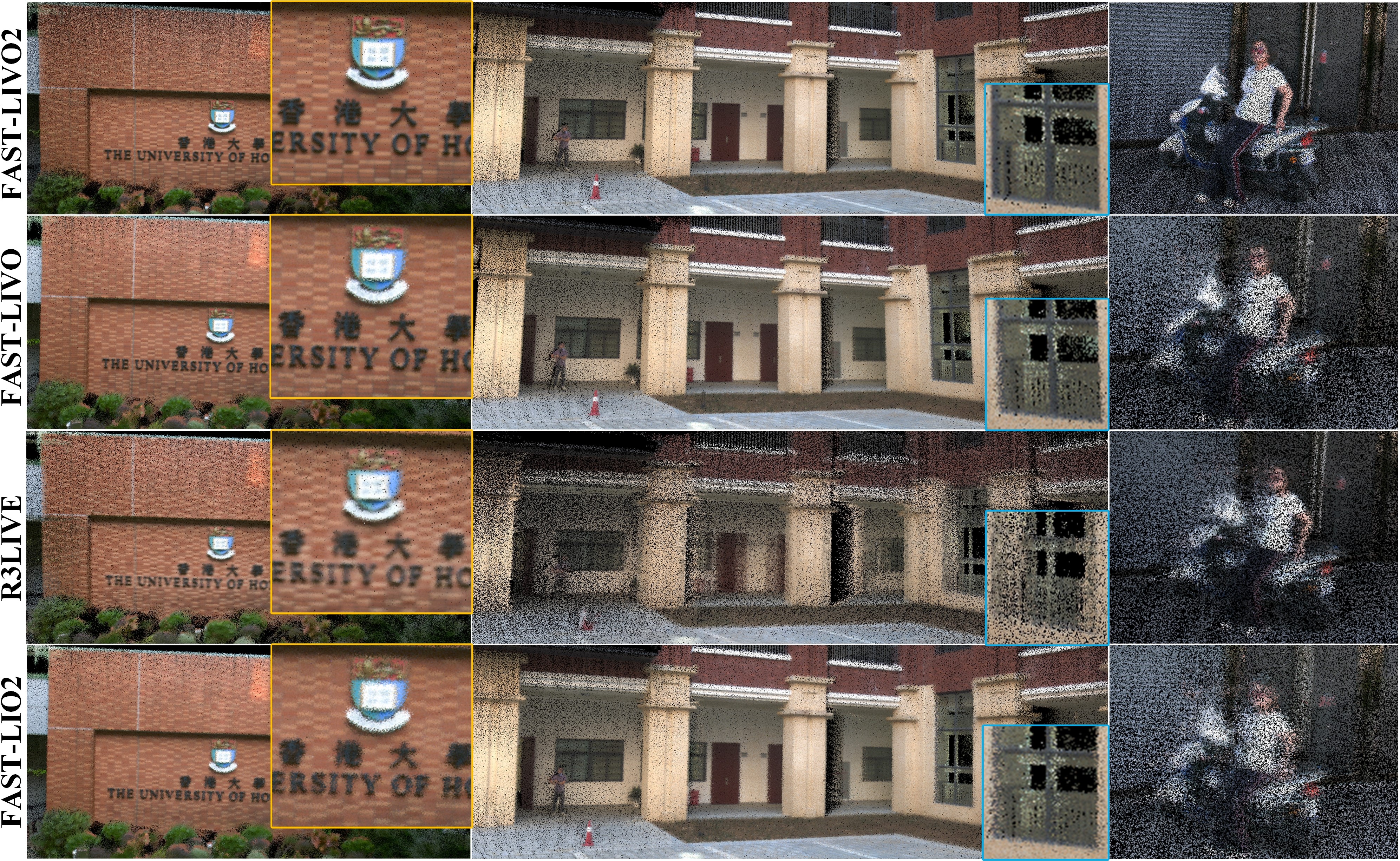}}
    \end{center}
    \vspace{-0.3cm}
    \caption{\label{fig:degrade-non} {The real-time mapping results generated online in rich texture and structured scenes. The point clouds from left to right correspond to ``HKU Landmark", ``SYSU 01" and "CBD Building 01", respectively, showing the comparison of colored point cloud accuracy among FAST-LIVO2, FAST-LIVO, R3LIVE, and FAST-LIO2.}}
    \vspace{-0.3cm}
\end{figure}
\begin{figure}[htp]
    \begin{center}
    \includegraphics[width=1.0\columnwidth]{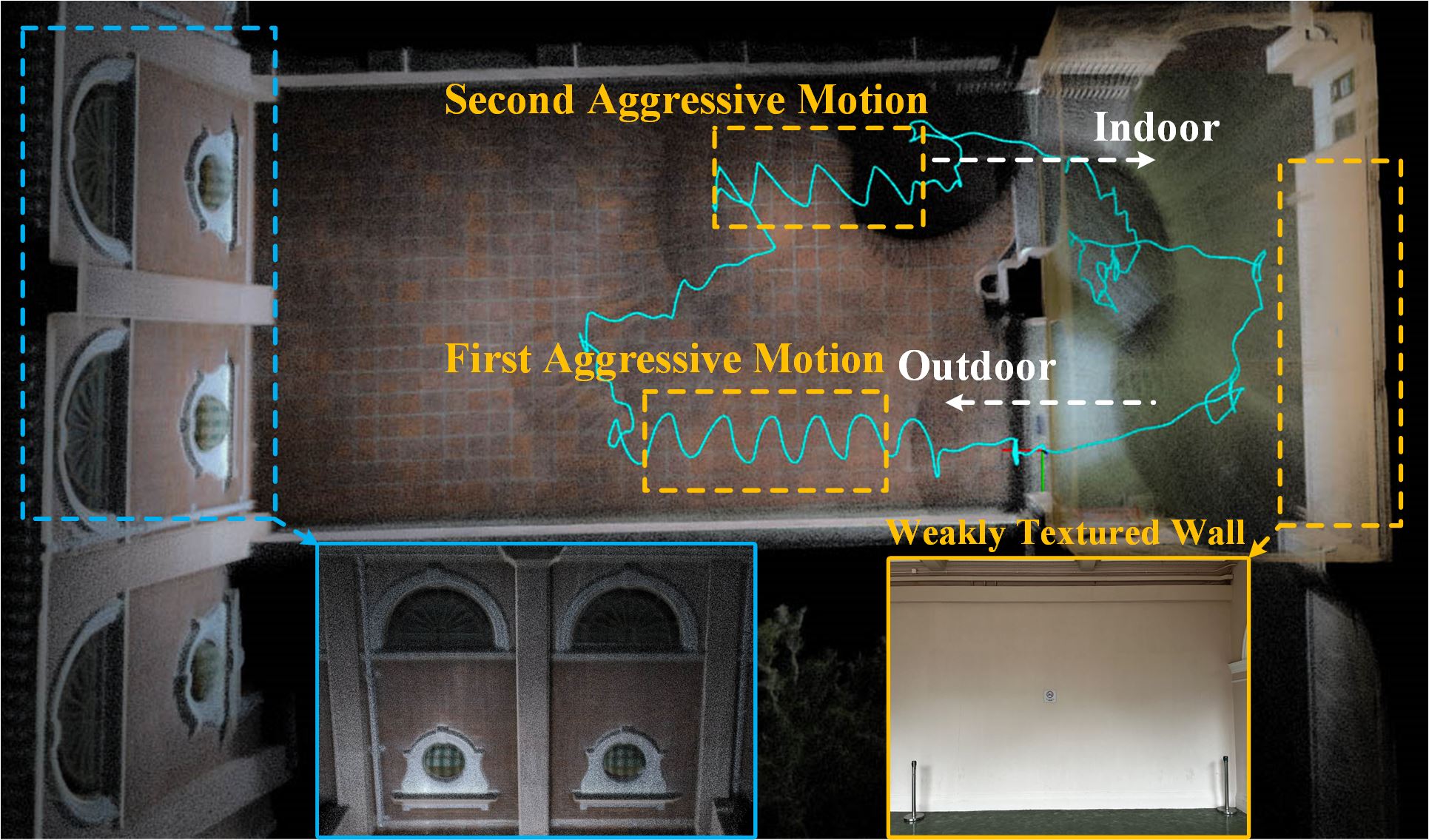}
    \end{center}
    \vspace{-0.3cm}
    \caption{\label{fig:hku_main_building}The real-time mapping results of FAST-LIVO2 in ``HKU Main Building", containing aggressive motions, indoor-to-outdoor, outdoor-to-indoor, and a weakly textured wall.}
\end{figure}
\begin{figure}[htp]
    \begin{center}
    \includegraphics[width=1.0\columnwidth]{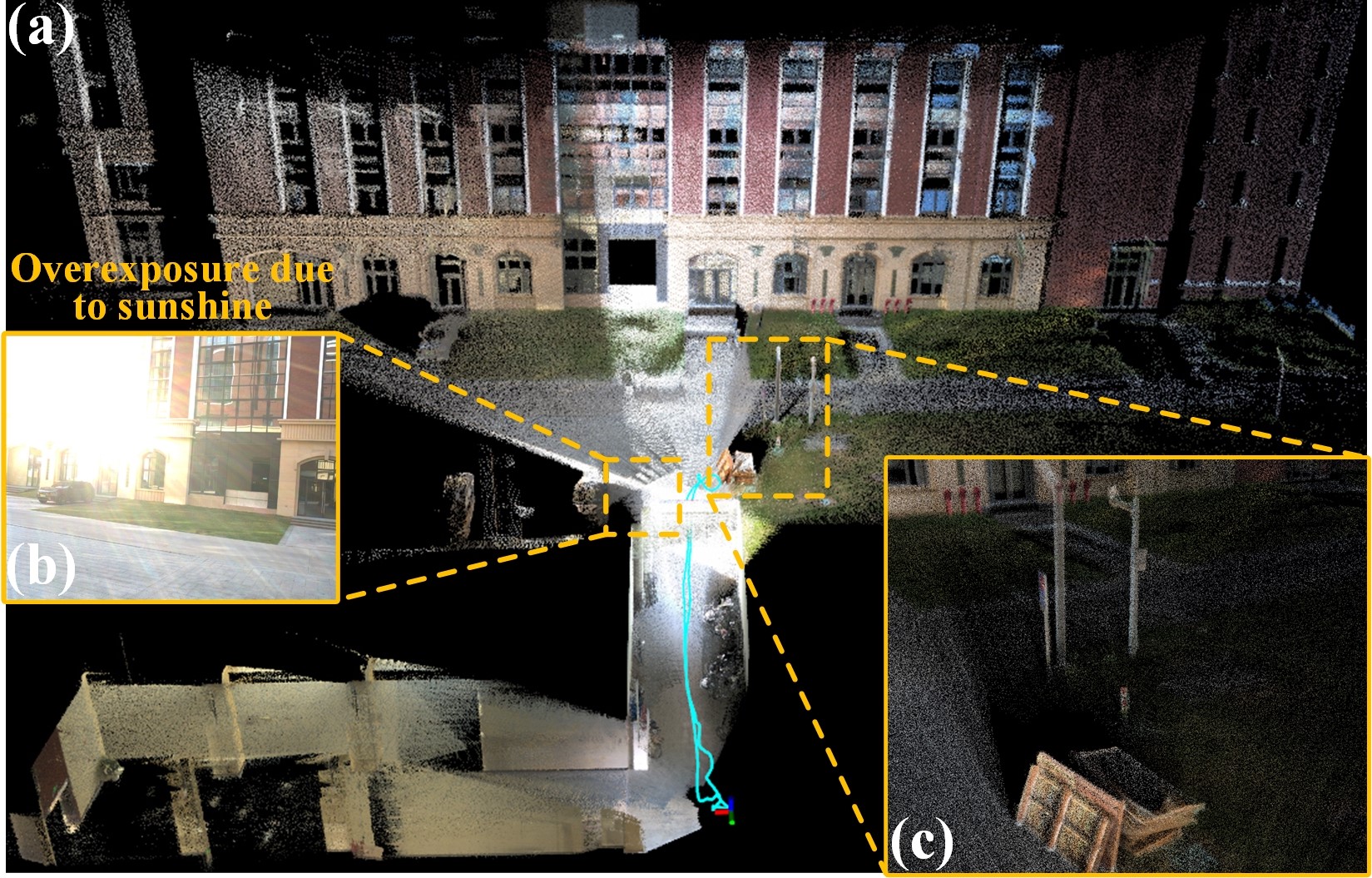}
    \end{center}
    \caption{\label{fig:sysu02}
The real-time mapping results of FAST-LIVO2 in ``SYSU 02". (a) birdview of the colored point map, (b) drastic lighting changes from indoor to outdoor facing the sun, (c) closeup view of details.}
\end{figure}
\begin{figure}[htp]
    \begin{center}
    \includegraphics[width=1.0\columnwidth]{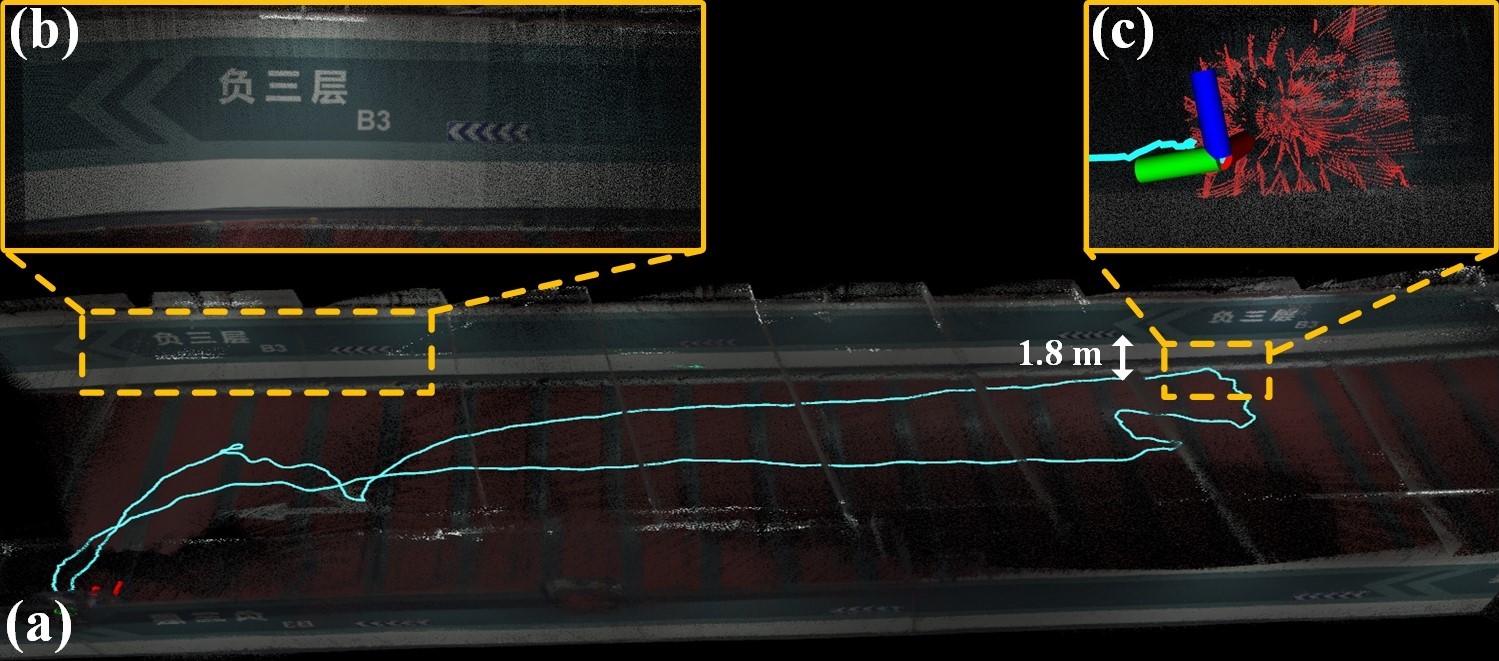}
    \end{center}
    \caption{\label{fig:long_corridor}The real-time mapping results of FAST-LIVO2 in ``Long Corridor". (a) birdview of the colored point map, (b) closeup view of details, (c) LiDAR facing a single wall causing LiDAR degeneration.}
\end{figure}
\begin{figure}[htp]
    \begin{center}
    \includegraphics[width=1\columnwidth]{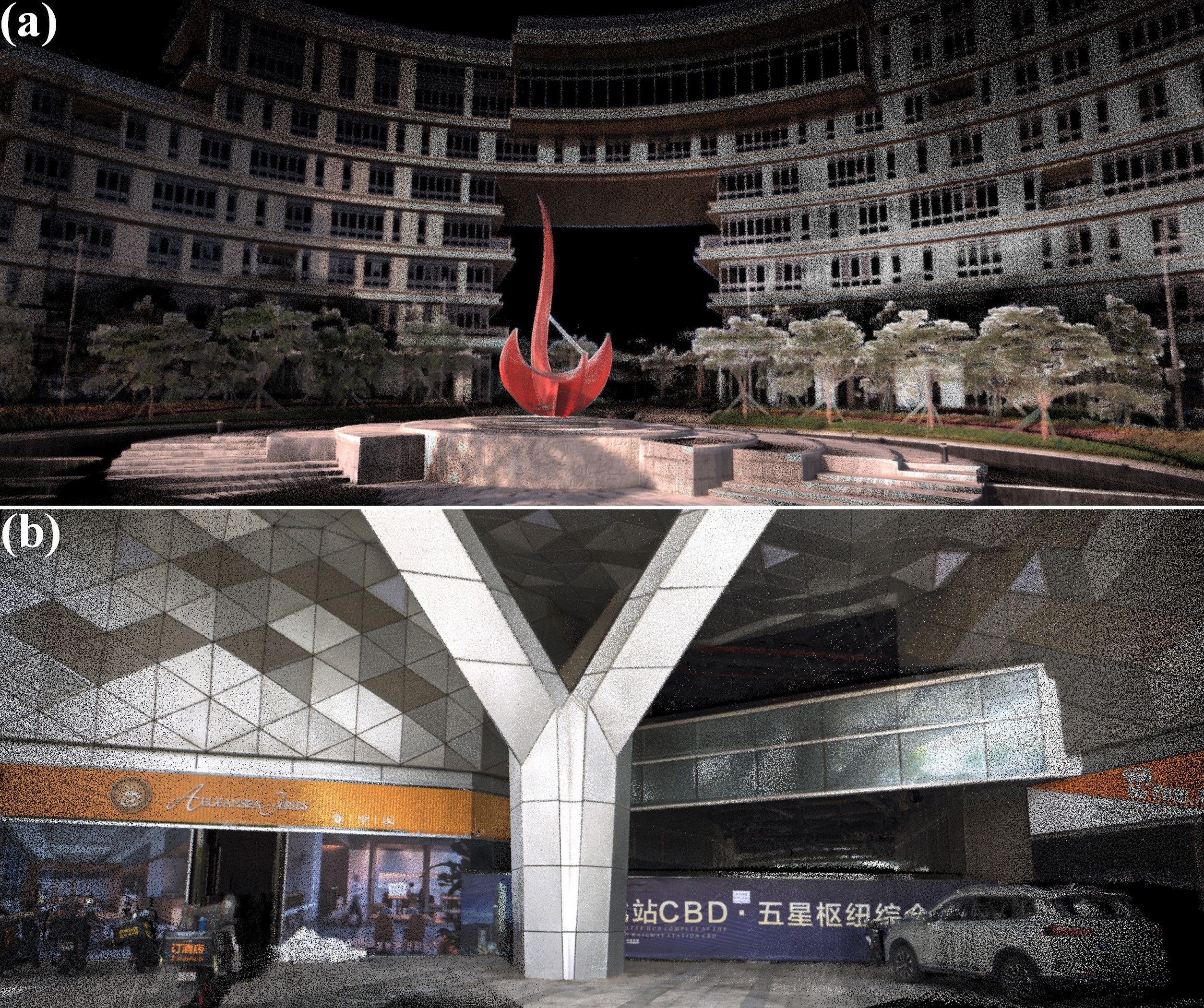}
    \end{center}
    \caption{\label{fig:hkust_red}The real-time mapping results of FAST-LIVO2 in 
    rich texture and structured scenes. (a) ``HKUST Red Sculpture", (b) ``CBD Building 01".}
\end{figure}
\begin{figure*}[htp]
    \begin{center}
    {\includegraphics[width=1.0\columnwidth]{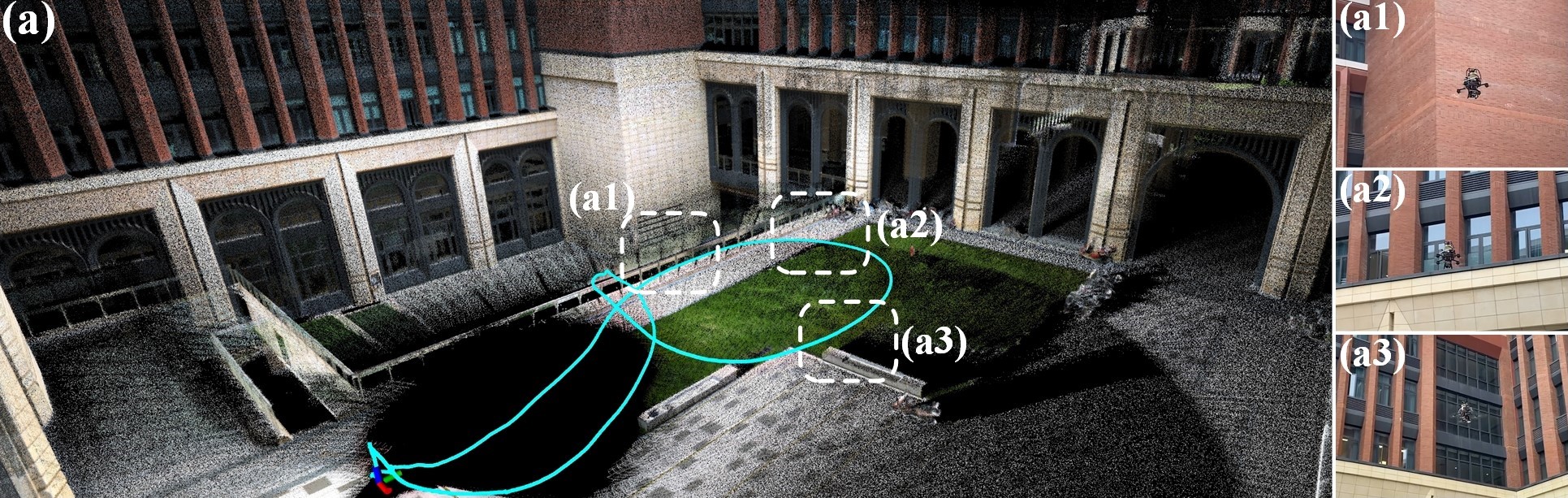}}
    \end{center}
    \caption{\label{fig:sysu}(a) is the enlarged point cloud image of the ``SYSU Campus" experiment. (a1), (a2), and (a3) represent the third-person view at the corresponding locations.}
\end{figure*}
\clearpage
\begin{figure*}[htp]
    \begin{center}
    {\includegraphics[width=1.0\columnwidth]{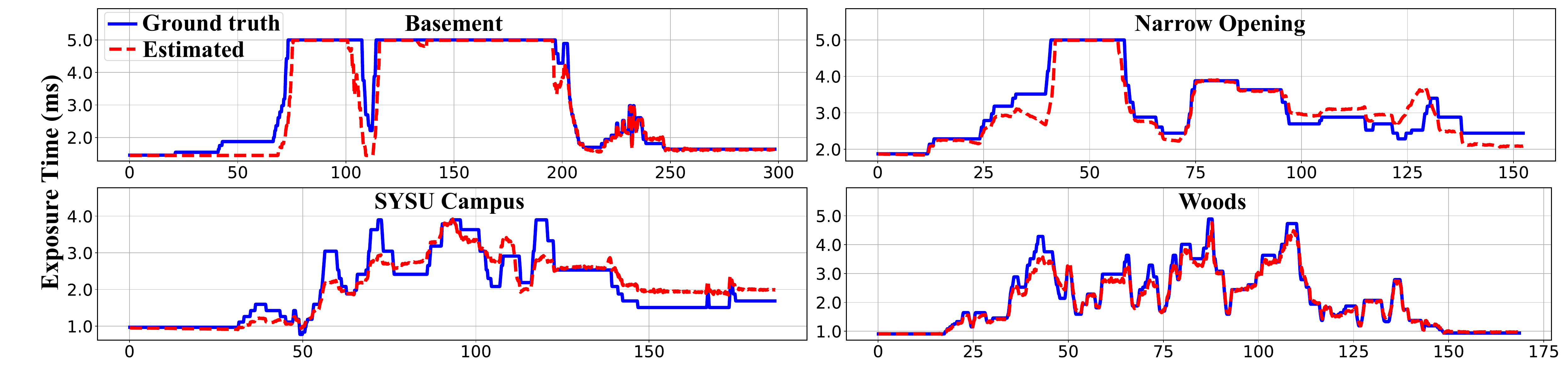}}
    \end{center}
    \caption{\label{fig:test_expo2}Comparison of estimated exposure time versus the ground truth exposure time in ``Basement", ``Narrow Opening", ``SYSU Campus", and ``Woods".}
\end{figure*}	
	\end{document}